\documentclass[iicol]{sn-jnl}

\usepackage{subcaption}
\usepackage{graphicx}
\usepackage{algorithm}
\usepackage{algpseudocode}
\usepackage{amsmath}
\usepackage{amsfonts}
\usepackage{xcolor}
\usepackage{blindtext}
\usepackage{upgreek}
\usepackage{multirow}
\usepackage{pifont}
\newcommand{\xmark}{\ding{55}}
\usepackage[para,online,flushleft]{threeparttable}

\usepackage{natbib}
\setcitestyle{authoryear,open={(},close={)}}

\graphicspath{ {./Images/} }

\jyear{2021}%

\theoremstyle{thmstyleone}%
\theoremstyle{thmstyletwo}%

\theoremstyle{thmstylethree}%

\begin{document}

\title[\hspace{7.5cm}A Review of Digital Twin - Part 2: Roles of UQ and Optimization, a Battery Digital Twin, and Perspectives]{A Comprehensive Review of Digital Twin - Part 2: Roles of Uncertainty Quantification and Optimization, a Battery Digital Twin, and Perspectives}


\author[1]{\fnm{Adam} \sur{Thelen}}\email{acthelen@iastate.edu}

\author[2]{\fnm{Xiaoge} \sur{Zhang}}\email{xiaoge.zhang@polyu.edu.hk}

\author[3]{\fnm{Olga} \sur{Fink}}\email{olga.fink@epfl.ch}

\author[4]{\fnm{Yan} \sur{Lu}}\email{yan.lu@nist.gov}

\author[5]{\fnm{Sayan} \sur{Ghosh}}\email{sayan.ghosh1@ge.com}

\author[6]{\fnm{Byeng D.} \sur{Youn}}\email{bdyoun@snu.ac.kr}

\author[7]{\fnm{Michael D.} \sur{Todd}}\email{mdtodd@eng.ucsd.edu}

\author[8]{\fnm{Sankaran} \sur{Mahadevan}}\email{sankaran.mahadevan@vanderbilt.edu}

\author*[1]{\fnm{Chao} \sur{Hu}}\email{chaohu@iastate.edu}

\author*[9]{\fnm{Zhen} \sur{Hu}}\email{zhennhu@umich.edu}

\affil[1]{\orgdiv{Department of Mechanical Engineering}, \orgname{Iowa State University}, \orgaddress{\city{Ames}, \postcode{50011}, \state{IA}, \country{USA}}}

\affil[2]{\orgdiv{Department of Industrial and Systems Engineering}, \orgname{The Hong Kong Polytechnic University}, \orgaddress{\city{Kowloon}, \country{Hong Kong}}}

\affil[3]{\orgdiv{Intelligent Maintenance and Operations Systems}, \orgname{Swiss Federal Institute of Technology Lausanne}, \orgaddress{\city{Lausanne}, \postcode{12309}, \state{NY}, \country{Switzerland}}}

\affil[4]{\orgdiv{Information Modeling and Testing Group}, \orgname{National Institute of Standards and Technology}, \orgaddress{\city{Gaithersburg}, \postcode{20877}, \state{MD}, \country{USA}}}

\affil[5]{\orgdiv{Probabilistic Design and Optimization group}, \orgname{GE Research}, \orgaddress{\city{Niskayuna}, \postcode{12309}, \state{NY}, \country{USA}}}

\affil[6]{\orgdiv{Department of Mechanical Engineering}, \orgname{Seoul National University}, \orgaddress{\city{Gwanak-gu}, \postcode{151-742}, \state{Seoul}, \country{Republic of Korea}}}

\affil[7]{\orgdiv{Department of Structural Engineering}, \orgname{University of California, San Diego}, \orgaddress{\city{La Jolla}, \postcode{92093}, \state{CA}, \country{USA}}}

\affil[8]{\orgdiv{Department of Civil and Environmental Engineering}, \orgname{Vanderbilt University}, \orgaddress{\city{Nashville}, \postcode{37235}, \state{TN}, \country{USA}}}

\affil[9]{\orgdiv{Department of Industrial and Manufacturing Systems Engineering}, \orgname{University of Michigan-Dearborn}, \orgaddress{\city{Dearborn}, \postcode{48128}, \state{MI}, \country{USA}}}

\abstract{As an emerging technology in the era of Industry 4.0, digital twin is gaining unprecedented attention because of its promise to further optimize process design, quality control, health monitoring, decision and policy making, and more, by comprehensively modeling the physical world as a group of interconnected digital models. In a two-part series of papers, we examine the fundamental role of different modeling techniques, twinning enabling technologies, and uncertainty quantification and optimization methods commonly used in digital twins. This second paper presents a literature review of key enabling technologies of digital twins, with an emphasis on uncertainty quantification, optimization methods, open source datasets and tools, major findings, challenges, and future directions. Discussions focus on current methods of uncertainty quantification and optimization and how they are applied in different dimensions of a digital twin. Additionally, this paper presents a case study  where a battery digital twin is constructed and tested to illustrate some of the modeling and twinning methods reviewed in this two-part review. Code and preprocessed data for generating all the results and figures presented in the case study are available on \cite{github_code}.
}

\keywords{Digital twin; Optimization; Machine learning; Enabling technology; Perspective; Industry 4.0, Review}



\maketitle

\section{Introduction}\label{sec1}
\label{Sec1}
This paper is the second in a series of two that analyze the role of modeling and twinning enabling technologies, uncertainty quantification (UQ), and optimization in digital twins. Modeling and twinning enabling technologies are fundamental methods used to bridge the information gap between a physical system and its digital counterpart. 

Part 1 of our two-part review provided an introduction to current state-of-the art methods used for digital twin modeling and proposed a five-dimensional digital twin model ~\citep{SMO1} based on the flow of data through the model. Additionally, Part 1 reviewed modeling and twinning technologies commonly used to model a physical system as a digital counterpart (P2V) and to model the return of decisions/actions determined by the digital twin back to the physical system which will carry them out (V2P).

\begin{figure*}[!ht]
    \centering
    \includegraphics[scale=0.56]{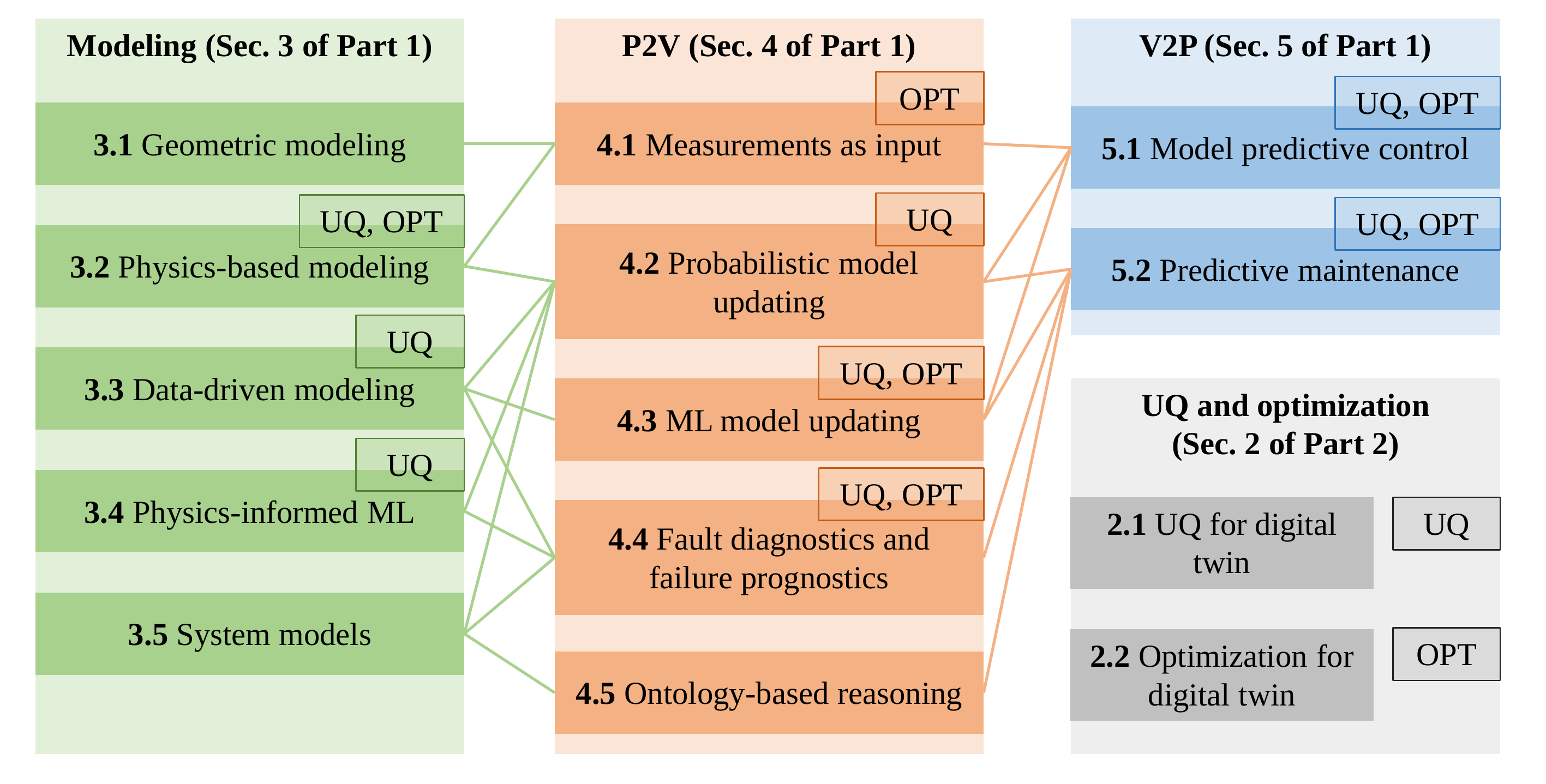}
    \caption{Roles of UQ and Optimization in different dimensions of digital twins}
    \label{fig:connection}
\end{figure*}

\begin{figure}[!h]
  \centering
    {\includegraphics[scale=0.65]{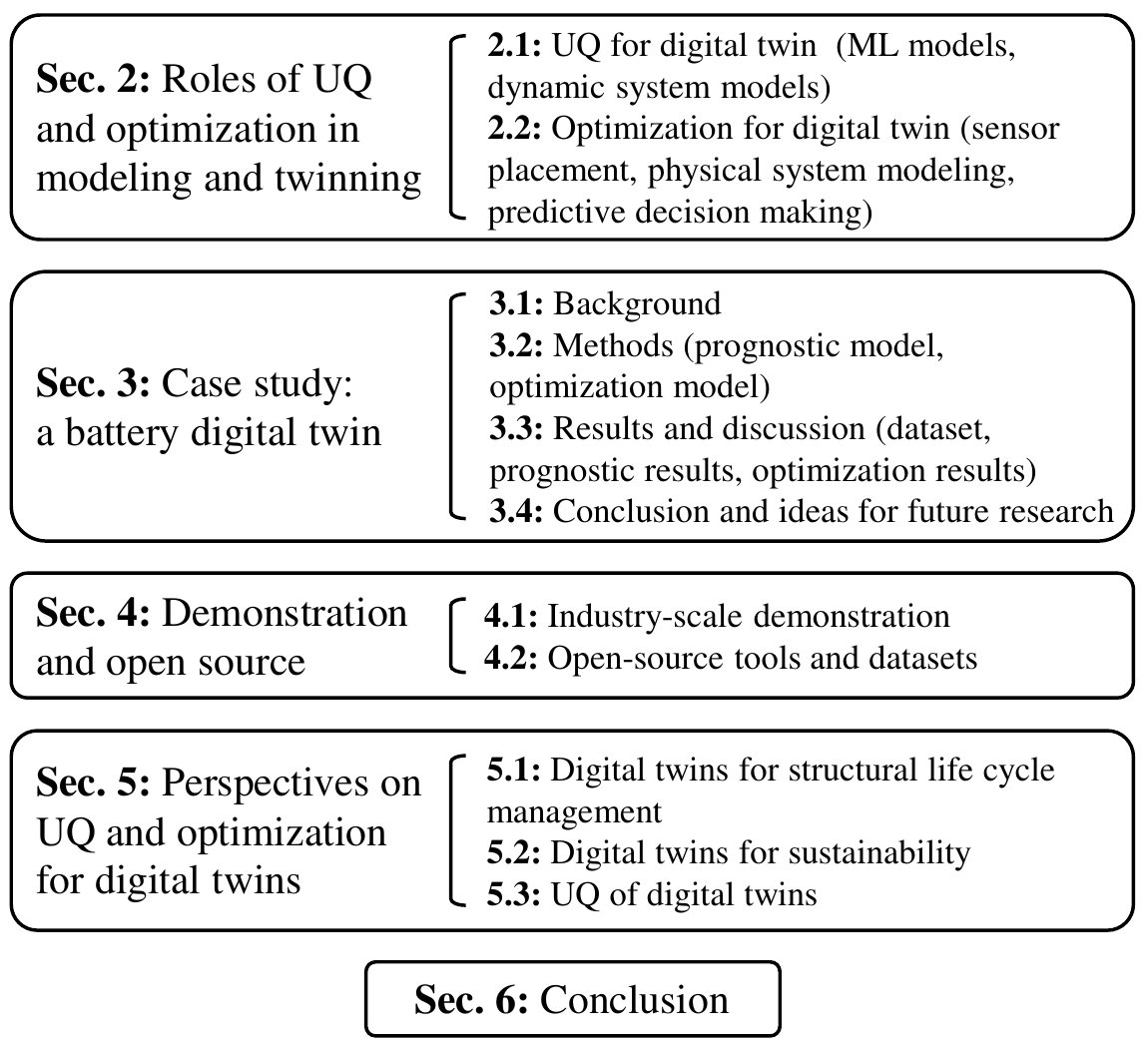}}
  \caption{Overview of topics covered in this paper}
  \label{fig:outline}
\end{figure}

In this paper, we review and analyze many methods and modeling techniques currently used to quantify uncertainty and support probabilistic inference and estimation in the presence of uncertainty in a digital twin. In addition, we also examine the crucial role of optimization in bridging the gap between a physical system and its digital counterpart through informative data collection and modeling. As indicated in Fig. \ref{fig:connection}, UQ and optimization play vital roles in all three dimensions (i.e., modeling, P2V, and V2P) of digital twins discussed in Part 1 of the two-part review paper. For instance, quantifying uncertainty that arises from various sources in modeling a physical system is essential for building an accurate digital twin and making informed decisions under uncertainty. Another example lies in ensuring effective P2V connection for model updating, fault diagnostics, failure prognostics, and other reasoning tasks. It is very important to optimize how data is collected from a physical system to maximize the value of information in the collected data. Moreover, optimization is indispensable for most tasks in the V2P dimension of digital twins, such as system reconfiguration, process control, production planning, maintenance scheduling, and path planning. The three dimensions reviewed in Part 1 are the fundamental pillars of digital twins, while UQ and optimization are essential to ensure the seamless synthesis of the three dimensions to allow digital twins to effectively perform their intended functions, such as design optimization, quality control, and maintenance planning, in uncertain environments. This part of the review paper is dedicated to the roles of UQ and optimization in digital twins. We also explicitly demonstrate the benefits of predictive decision making augmented by optimization in several applications. To demonstrate many of the concepts discussed in both Part 1 and Part 2 of this review, we construct a battery digital twin and use this digital twin to optimize the retirement of a battery cell from its first life application, which vividly showcases the application of digital twin in the context of predictive maintenance scheduling (Sec. \ref{sec:predictive_maintenance_scheduling}). Last, we close by discussing digital twin trends in industry, and present some open source software and datasets which may be of use to researchers and practitioners. Figure \ref{fig:outline} gives an overview on the topics covered in this paper.

We begin by analyzing the integration of UQ and optimization for use in digital twins in Sec. \ref{sec:UQ and optimization}. In what follows, Sec.~\ref{case_study} demonstrates some of the reviewed techniques with a case study of a battery digital twin. Sec. \ref{sec5} reviews the applications of digital twins at industrial scale and available open-source tools and datasets related to digital twins. Next, we discuss challenges and future research directions in Sec. \ref{sec8}. Finally, concluding remarks are given in Sec. \ref{sec9}.

\section{Roles of UQ and optimization in digital twins}
\label{sec:UQ and optimization}
In this section, we discuss the role of UQ in digital twins, cover UQ of ML models and UQ of dynamic system models. Following that, we review the role of optimization in digital twins, and discuss optimization methods for sensor placement, physical system modeling, and predictive decision making.

\subsection{UQ for digital twins}
\label{sec:UQ_digital_twin}
\begin{figure}[!ht]
    \centering
    \includegraphics[scale=0.75]{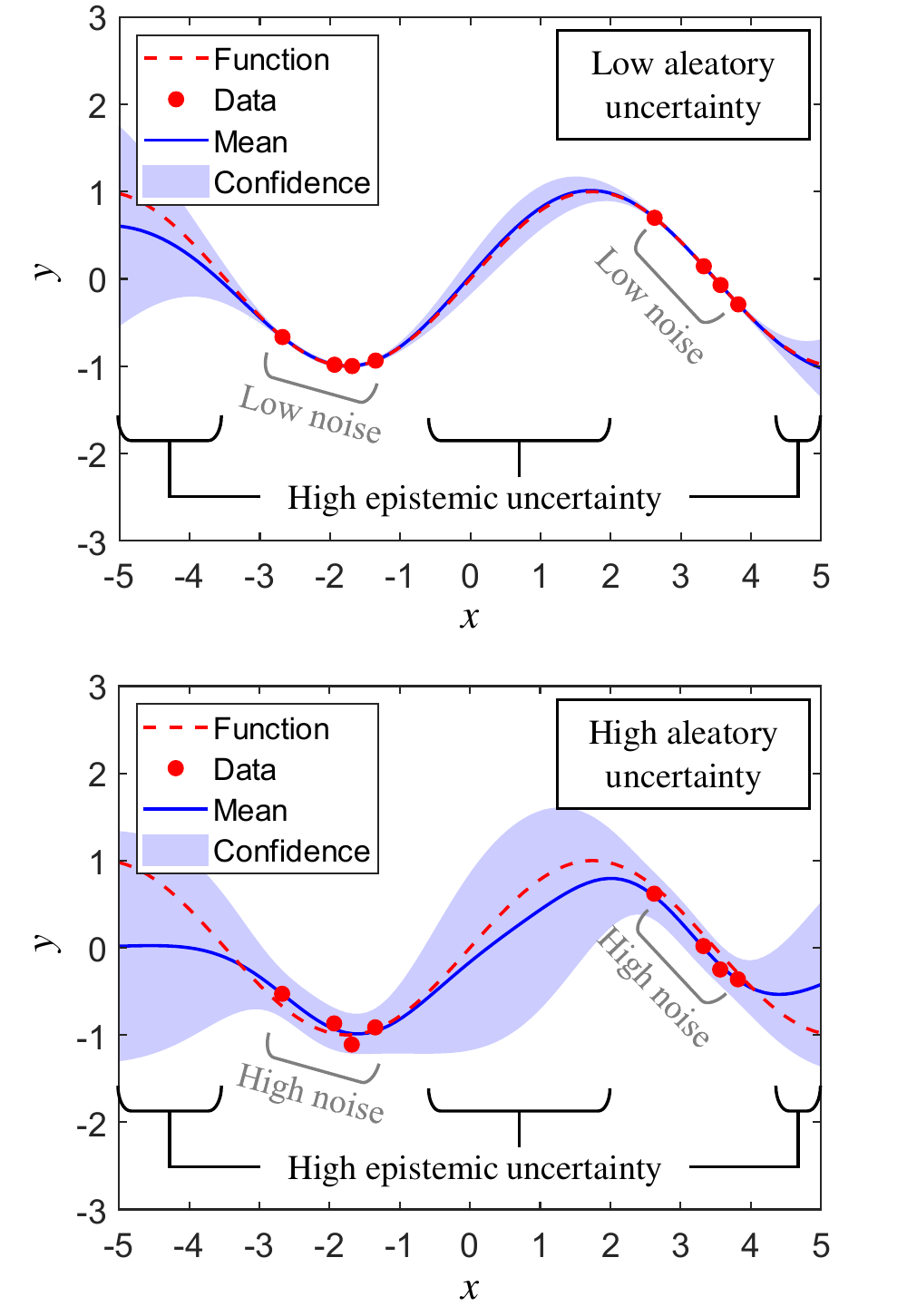}
    \caption{A graphical comparison of aleatory uncertainty and epistemic uncertainty.}
    \label{fig:aleatory_epistemic}
\end{figure}

First mentioned in the definition by Glaessgen  and Stargel in their conference paper \citep{glaessgen2012digital}, digital twin is \emph{\lq\lq an integrated multiphysics, multiscale, \emph{probabilistic} simulation of..."}. Probabilistic simulation plays an essential role in digital twins since variability is inherent and inevitable. A large and heterogeneous set of uncertainty sources is present in the five dimensions of the proposed digital twin model in Fig. 3 of \cite{SMO1}. The uncertainty sources in a digital twin can be classified into two categories \citep{der2009aleatory}:
\begin{itemize}
\item \emph{Aleatory uncertainty} refers to uncertainty due to natural variability, which is inevitable and irreducible. Examples include sensor measurement errors and variability in material properties and load conditions. This intrinsic uncertainty can often be captured by fitting a probability distribution to a limited amount of data. 
\item \emph{Epistemic uncertainty} refers to uncertainty caused by limited data, lack of knowledge, and/or model simplifications and assumptions. These sources of uncertainty are reducible when more information or data becomes available. For example, model uncertainty discussed in Sec. \ref{sec:UQ_dynamic_system_models} is one of the most important epistemic uncertainty sources. Another example is ML models will have high predictive uncertainty if trained on small volumes of data. This model uncertainty can be reduced by either gathering more data (Sec. 4.2 on probabilistic model updating in \cite{SMO1} or incorporating known physics (see Sec. 3.4 on physics-informed ML in \cite{SMO1}).
\end{itemize}
A simple graphical comparison of aleatory uncertainty and epistemic uncertainty is given in Fig. \ref{fig:aleatory_epistemic}. \cite{hu2017uncertainty} provides a detailed discussion on how to model various uncertainty sources in the context of additive manufacturing. In this paper, we mainly focus on the quantification of epistemic uncertainty that is particularly relevant to digital twins. In addition, it is worth mentioning that there are many different ways of modeling uncertainty, such as probabilistic versus non-probabilistic and frequentist versus Bayesian statistics. For example, various non-probabilistic methods, including interval theory \citep{gao2010probabilistic}, fuzzy method \citep{bing2000practical}, and evidence theory \citep{zhang2017reliability,soundappan2004comparison}, have been investigated in the past decades to model epistemic uncertainty in different engineering domains. According to the literature review in Part 1, probabilistic methods are more widely used than non-probabilistic methods in digital twins. Therefore, this paper mainly focuses on probabilistic methods for UQ in digital twins.

\subsubsection{UQ of ML models}
\label{sec:UQ_ML_models}
From the literature review, it is observed that ML models are extensively used in constructing digital models in the virtual space (see Sec. 3.3.2 of our Part 1 paper on ML models). As data-driven models, the performance of ML models is significantly affected by the quantity and quality of the data used for model training. As discussed in Sec. 4.3 of our Part 1 paper on ML model updating, ML models have difficulties generalizing to test data outside of a training distribution. When these trained ML models are deployed in digital twins, they may fail unexpectedly on out-of-distribution (OOD) samples. These unexpected failures reduce end users' trust and limit industry-scale, real-world adoptions of digital twin. This generalizability issue can be mitigated, to some degree, by incorporating physics (see Sec. 3.4 of our Part 1 paper on physics-informed ML) or by fine-tuning ML models based on newly labeled samples (see Sec. 4.3 of our Part 1 paper). However, predictions by physics-informed ML models and those with online updating still will not be perfect. It is highly desirable to quantify the predictive uncertainty of ML models, and in some safety-critical applications, such as autonomous driving and medical diagnostics, UQ of ML models becomes crucial. High quality estimation of an ML model's predictive uncertainty provides an accurate estimate of the model's confidence in a certain prediction, may allow for the detection of a data/concept shift, and most importantly, helps determine when the model is likely to fail. Over the past few years, UQ of ML models has become an established subdiscipline of ML, developed and promoted by the ML community. This subsection aims to provide an overview of this subdiscipline. More detailed and dedicated reviews can be found in two recent review papers \citep{abdar2021review, gawlikowski2021survey} and some benchmarking work has been presented in \cite{nado2021uncertainty}.

UQ of ML models mainly deals with two tasks. First, it measures the predictive uncertainty for every training/test sample. For example, for the direct mapping strategy in Fig. 22 of our Part 1 paper, ML models capable of UQ can predict a probability distribution of RUL for every vector/matrix of input features rather than a point estimate. The so-called ``calibration curve" or ``reliability curve" can be plotted to visualize the quality of uncertainty estimation by a probabilistic ML model \citep{niculescu2005predicting, kuleshov2018accurate}. In a calibration curve, the observed model confidence ($y$) is plotted against the expected model confidence ($x$) for equally spaced values between 0 and 1. An ML model with perfect uncertainty estimation should produce a reliability curve that follows the straight line $y = x$. For example, at a confidence level of 95\%, we expect the observation (ground truth) to fall inside the 95\% confidence interval produced by the perfect ML model 95\% of the time. This way to evaluate the accuracy of UQ is interestingly similar to the U-pooling method used in the statistical validation of computer simulation models \citep{ferson2008model,liu2011toward} in that they both look at the area difference between an observed curve and an ideal straight line $y = x$. However, the U-pooling method is used to measure the disagreement between the probability distributions of a model prediction and an experimental observation, not the difference between the expected and actual model confidence.

Approaches for UQ of ML models mostly focus on estimating epistemic uncertainty (incomplete knowledge due to lack of data), as aleatory uncertainty (inherent noise in data) can be learned directly from data. Table \ref{tab:uq_of_ml_comp} compares four popular approaches to quantify the uncertainty of ML models, and these four approaches are elaborated in what follows. We note that UQ of ML models is an active and quickly evolving field of research, and many new approaches (not discussed in this review) are emerging to estimate the predictive uncertainty of ML models.

\begin{itemize}
\item \textbf{Gaussian process regression} is probably one of the earliest probabilistic ML algorithms that can capture epistemic uncertainty \citep{williams1995gaussian}. The basic idea of Gaussian process regression is to assume the $y$s (i.e., output values of training data) at $x$ coordinates follow a multivariate Gaussian and derive the conditional Gaussian of the $y$ at a new $x$ coordinate (test point), given the $y$ values observed at some $x$ coordinates (training points). Gaussian process regression has a rigorous mathematical formulation and deduction. It allows one to estimate the model predictive uncertainty in a closed-form expression. A limitation of Gaussian process regression is its difficulty in scaling to high-dimensional input spaces. Some dimensionality reduction or feature extraction will be needed for high-dimensional problems, but this intermediate step may degrade the prediction accuracy. 
\item \textbf{Bayesian neural networks} represent a principled way to measure the predictive uncertainty of a neural network. In this approach, a probabilistic neural network is built by first assuming the network weights and biases follow some prescribed probability distributions (often referred to as a prior) and then inferring the posterior based on the prior and some training data \citep{kendall2017uncertainties}. Bayesian estimation of neural network parameters is similar to the standard Bayesian inference approach (Category 1: Parameter calibration) described in Sec. \ref{sec:UQ_dynamic_system_models} (a). As discussed in Sec. 3.3.2 of Part 1, when dealing with high input dimensions and given large volumes of training data, deep neural networks become an attractive alternative to traditional ML algorithms such as Gaussian process regression and random forest. However, training Bayesian deep neural networks involves approximate Bayesian inference such as Markov chain Monte Carlo ~\citep{andrieu2003introduction} or variational inference on many network parameters. It requires significant changes to the standard model training procedure and is more computationally costly than training non-Bayesian deep neural networks ~\citep{papamarkou2021challenges}. 
\item \textbf{Ensembles of neural networks}, also called deep ensembles in the case of deep neural networks, are widely accepted as a powerful approach for UQ of ML models \citep{lakshminarayanan2017simple}. An ensemble consists of independently trained neural networks with an identical architecture. For regression problems, a Gaussian layer is often added at the end of each network, allowing for predicting the mean and variance of a Gaussian output. Two central ideas of this ensemble-based approach are: (1) a measure of the difference between different predictors can be used as a proxy for epistemic uncertainty, and (2) the Gaussian layer of each network captures aleatory uncertainty. Deep ensembles are simple to train and test. Although more efficient to train and test than Bayesian neural networks, deep ensembles still require high computational costs (multiple forward passes) and a large memory footprint (use of multiple neural networks). These issues impede their adoption in real-world digital twin applications where computational power and resources are limited. 
\item \textbf{More efficient approaches} than deep ensembles include Monte Carlo dropout \citep{gal2016dropout} and approaches using deterministic neural networks \citep{van2020uncertainty, liu2020simple, mukhoti2021deterministic}. Monte Carlo dropout samples multiple sets of network weights to build multiple predictors from the same trained neural network and uses all predictors when predicting. It has been extensively studied on CNNs ~\citep{gal2015bayesian} and recurrent neural networks ~\citep{gal2016theoretically} and is sometimes viewed as an efficient approximation (via variational inference) of a Bayesian neural network. \cite{gal2016dropout} proved that Monte Carlo dropout minimizes the Kullback–Leibler divergence between the approximate posterior and true posterior of a Bayesian neural network. Monte Carlo dropout shares some similarities with deep ensembles, although deep ensembles use multiple trained networks at test time and showed better accuracy in uncertainty estimation in several studies \citep{lakshminarayanan2017simple, nemani2021ensembles}. Unlike deep ensembles and Monte Carlo dropout, which require multiple forward passes, deterministic neural network approaches require only a single forward pass to estimate uncertainty and have a shorter inference time \citep{van2020uncertainty, liu2020simple, mukhoti2021deterministic}. The basic idea of these deterministic approaches is to estimate the density of training points close to a test point in the embedded feature space learned by an ML model and use this density estimate as a proxy for epistemic uncertainty. The logic behind this idea is that adding new training points close to a test point in a high-level feature space is expected to reduce the epistemic uncertainty at the test point significantly. Aleatory uncertainty for in-distribution samples can be captured by including a softmax function in the final layer of a neural network classifier \citep{mukhoti2021deterministic} or adding a Gaussian output layer to a neural network regressor \citep{lakshminarayanan2017simple}. This way, these deterministic network networks not only quantify the overall predictive uncertainty attributable to aleatory uncertainty and epistemic uncertainty, but they may also separate the contributions from the two types of uncertainty. 
\end{itemize}

\begin{table*}[!ht]\small
    \centering
    \caption{A comparison of four different approaches for UQ of ML models}
    \begin{threeparttable}
    \begin{tabular}{p{2.2cm}|p{2.4cm}|p{2.4cm}|p{2.4cm}|p{2.2cm}|p{2cm}}
    \hline \hline
    Quantity of interest    &  Gaussian process regression & Bayesian neural networks & Ensembles of neural networks & Monte Carlo dropout & Deterministic approaches\\
    \hline
    Accuracy in UQ  &	High  &	High-medium$^{a}$  &	High   &	Medium &	High-medium \\
    \hline
    Computational efficiency (test time)  &	High$^{b}$  &	Low$^{c}$ &	Medium-low &	Medium-low &	High \\
    \hline
    Ability to detect OOD samples &	Strong &	Weak (may estimate low uncertainty) &	Weak (may estimate low uncertainty) &	Weak (may estimate low uncertainty) &	Strong \\
    \hline
    Scalability to high dimensions &	Low &	High &	High &	High &	High \\
    \hline \hline
    \end{tabular}
    \begin{tablenotes}
          \item[a] Accuracy is largely affected by the quality of the assumed prior. \\
          \item[b] Efficient only for problems of low dimensions (typically $<$ 10). \\
          \item[c] Could be efficient if variational Bayesian methods are employed to approximate the output posterior in in Bayesian neural networks.
    \end{tablenotes}
    \end{threeparttable}
    \label{tab:uq_of_ml_comp}
\end{table*}

The second task is to identify test samples where a trained ML model has low confidence in predicting. These test samples often differ significantly from the samples the ML model is trained on and can be called OOD test samples, which we have discussed many times (thus, this task is sometimes referred to as OOD detection). These low confidence predictions are not trustworthy and should be examined by domain experts if a time delay from a prediction to a decision is acceptable. The need for extra caution is because low-confidence predictions are likely largely incorrect, and decisions made based on them without consideration of uncertainty will be flawed. For example, an ML model for fault diagnostics used in the ML pipeline for predictive maintenance shown in Fig. 26 of Part 1 may produce a false alarm at an OOD sample due to large measurement noise, warning that maintenance is needed on a pump that has only degraded slightly and has plenty of useful life left. This false-positive scenario may lead to an unnecessary maintenance action that can erode the trust of the end-users. A better alternative would be to associate this prediction with low confidence. Reliability and maintenance engineers can then investigate the model input features and determine if the model prediction makes physical sense before taking any maintenance actions. 

Approaches to measuring model confidence include quantifying the degree of disagreement among an ensemble of ML models \citep{weigert2018content} and measuring distances between a test sample and its nearest training neighbors in the embedded space learned by an ML model \citep{mandelbaum2017distance, liu2020simple}. The ensemble disagreement approach is inspired by and based on deep ensembles that have been discussed. It computes the average difference between the predictive distributions of the ML models in an ensemble and the predictive distribution of the ensemble. Although the Kullback–Leibler divergence was used as the distribution difference measure \citep{weigert2018content}, other measures of how one probability distribution is different from another can also be used. The distance-based approach calculates a confidence score for each predicted class at a test point based on the local density of training points with the same class in the embedded space \citep{mandelbaum2017distance}. It was originally developed for classification problems but could be extended for regression problems. It is closely related to the deterministic approaches to estimating epistemic uncertainty, and the confidence score can be viewed as a side-product of epistemic uncertainty estimation.

\subsubsection{UQ of dynamic system models}
\label{sec:UQ_dynamic_system_models}
UQ of dynamic system models is a process of quantifying uncertainty in certain system outputs due to both aleatory and epistemic uncertainty sources (see the classification of uncertainty sources in Sec. \ref{sec:UQ_digital_twin}). It could be forward uncertainty propagation or inverse UQ \citep{smith2013uncertainty}. The former focuses on propagating various uncertainty sources to the uncertainty of outputs. The latter emphasizes quantifying model uncertainty of computer simulation models based on observations. We focus on the latter in this section, and UQ herein means quantification of model uncertainty, an important type of uncertainty in digital twins that needs to be properly quantified and managed. 

Model uncertainty arises from two main sources:
\begin{enumerate}
    \item \emph{Model parameter uncertainty}: This is the uncertainty in model parameters ${\boldsymbol{\uptheta}}$ due to lack of knowledge. It is worth mentioning that model parameter uncertainty could be either epistemic uncertainty due to lack of knowledge/data, or aleatory uncertainty due to natural variability (i.e., specimen to specimen variability), or both. In this section, we mainly focus on epistemic uncertainty. 
    
    \item \emph{Model form uncertainty}: It results from imperfect modeling due to model assumption, simplification, lack of good understanding of the physics, etc. It is also referred to as model structure errors, model discrepancy, model bias, and model form error in the literature~\citep{jiang2020sequential,arendt2012quantification,kennedy2001bayesian}.
\end{enumerate}

We first introduce three categories of methods for UQ of general system models, which we call generalized methods. Following that, we discuss methods for UQ of dynamic system models, focusing on UQ of measurement equations and UQ of state transition equations.

\paragraph{(a) Generalized methods}
Due to the difficulty in completely separating uncertainty in model parameters from uncertainty in model form, quantification of model uncertainty remains a challenging issue in the modeling and simulation of various engineered systems \citep{arendt2012quantification}. To improve the prediction accuracy of computer codes/simulation models, various approaches have been developed in the past decades and they can be roughly classified into three categories:
\begin{itemize}
\item {\bf{Category 1: Parameter calibration}} This group of methods captures model uncertainty using uncertain model parameters and a noise term \citep{astroza2019dual,song2019accounting}. A generalized model is formulated as
\begin{equation}
\label{eq:model_uncertainty1}
{{\vartheta_{e}}} = {{g}_m}({\bf{z}},\;{\boldsymbol{\uptheta }}) + {{\gamma }},
\end{equation}
where ${{\vartheta_{e}}} \in \mathbb{R}$ is an observation, ${\vartheta}={{g}_m}({\bf{z}},\;{\boldsymbol{\uptheta }})$ is a computer simulation model with output ${\vartheta}$ and inputs ${\bf{z}}$ and ${\boldsymbol{\uptheta }}$, ${\bf{z}} \in \mathbb{R}^{n_z}$ is a vector of measurable/controllable input variables which may change with observations, ${\boldsymbol{\uptheta }}\in \mathbb{R}^{n_t}$ is a vector of uncertain model parameters,
the true values of ${\boldsymbol{\uptheta }}$ are usually fixed but unknown to us, and ${{{\gamma }}}$ stands for model noise which is typically modeled as a Gaussian random variable with either unknown mean and variance \citep{song2019accounting} or zero mean and unknown variance \citep{astroza2019dual}. The noise term ${{{\gamma }}}$ accommodates observation noise and part of model form uncertainty that is not captured by ${\boldsymbol{\uptheta }}$. We note that the model output, observation, and noise could be vectors in dynamic system models (see Sec. \ref{sec:UQ_dynamic_system_models}(b)). Here, they are constrained to be scalars to simplify the explanation \citep{kennedy2001bayesian}.

\hspace{0.4cm} A benefit of formulating the quantification of model uncertainty problem in Eq. (\ref{eq:model_uncertainty1}) is that it casts the problem as a standard Bayesian model updating problem, allowing Bayesian inference methods to be used directly for model updating. Attributing model form uncertainty to ${\boldsymbol{\uptheta }}$ and the noise term ${{{\gamma }}}$, however, makes it independent from the inputs ${{\bf{z}}}$. As a result, it may overestimate or underestimate model form uncertainty for some values of ${{\bf{z}}}$.

\item {\bf{Category 2: Bias correction}} A point estimate ${\boldsymbol{\uptheta}^*}$ of the uncertain model parameters ${\boldsymbol{\uptheta }}$ is first obtained using the maximum likelihood estimation method or another offline calibration method discussed in Sec. \ref{sec:opt_physical_offline}. Afterwards, this category of methods fixes the uncertain model parameters ${\boldsymbol{\uptheta }}$ at ${\boldsymbol{\uptheta}^*}$ and  uses a model discrepancy function and a noise term to account for model uncertainty. The computer simulation model ${\vartheta}={{g}_m}({\bf{z}},\;{\boldsymbol{\uptheta }})$ after adding the model discrepancy/bias function is given by \citep{wang2009bayesian}
\begin{equation}
\label{eq:model_uncertainty2}
\begin{split}
{{\vartheta_{e}}} = {g_m}({\bf{z}},\;{\boldsymbol{\uptheta }^*}) + \delta ({\bf{z}}) + {{\gamma }},
\end{split}
\end{equation}

where $\delta ({\bf{z}})$ is a model discrepancy function that corrects the original computer simulation model. In the equation above, $ \delta ({\bf{z}})$ accommodates most of the model form uncertainty and ${{{\gamma }}}$ represents the residual model form uncertainty and observation noise. Similar to methods in Category 1, ${{{\gamma }}}$ is modeled as a Gaussian random variable with either unknown mean and variance or zero mean and unknown variance \citep{xiong2009better}. The rationale of the above formulation is that the additional model form uncertainty caused by the inaccurate estimation of ${\boldsymbol{\uptheta}^*}$ can be compensated by the model discrepancy function $\delta ({\bf{z}})$ and the noise term ${{{\gamma }}}$. 

\hspace{0.4cm} A data-driven model is usually constructed for $\delta ({\bf{z}})$ using methods discussed in Sec. 3.3 of our Part 1 paper on data-driven modeling. The predictive capability of data-driven models enables this line of methods to improve the prediction accuracy of the model under previously unseen conditions, as long as it is within the prediction capability of the data-driven model. The challenge for this category of methods is how to build an accurate model of ${{{\delta }}}({\bf{z}})$.

\item {\bf{Category 3: The KOH framework}} In order to simultaneously quantify various sources of model uncertainty, \cite{kennedy2001bayesian} developed a model calibration framework using Bayesian method and Gaussian process regression models. It is now the most widely used and commonly referred to as the KOH framework in the literature. The KOH framework constructs a Gaussian process regression model for the computer simulation model and another Gaussian process regression model as the model bias term. The two Gaussian process regression models are related to each other and other uncertainty sources as follows:
\begin{equation}
\label{eq:KOH_equation}
\begin{split}
{{\vartheta}_e} = \rho \hat {g}_m({\bf{z}},\;{\boldsymbol{\uptheta }} ; {{\boldsymbol{\upeta }}_g}) + \hat \delta ({\bf{z}}; {{\boldsymbol{\upeta }}_\delta }) + \gamma ,
\end{split}
\end{equation}
where  $\rho$ is an unknown regression coefficient within the range of $[0, 1]$ and it accounts for model uncertainty using a multiplication in addition to an additive bias term $\hat \delta(\cdot)$ and a noise term $\gamma$, $\hat {\vartheta}=\hat g_m({\bf{z}},\;{\boldsymbol{\uptheta }}; {{\boldsymbol{\upeta }}_g})$ and $\hat \delta ({\bf{z}};{{\boldsymbol{\upeta }}_\delta })$ are, respectively, the Gaussian process regression models of the computer simulation model ${\vartheta}={{g}_m}({\bf{z}},\;{\boldsymbol{\uptheta }})$ and the model bias term, ${{\boldsymbol{\upeta }}_g}$ and ${{\boldsymbol{\upeta }}_\delta}$ are hyperparameters of the two Gaussian process regression models respectively. For the noise term $\gamma$, it is modeled as a Gaussian random variable with zero mean and unknown standard deviation $\sigma_{\gamma}$ in the original KOH framework. In some variants of the KOH framework, however, $\gamma$ is modeled as a Gaussian random variable with unknown mean and unknown standard deviation \citep{xiong2009better}. To estimate the unknowns (i.e., $\rho$, ${\boldsymbol{\uptheta }}$,  ${{\boldsymbol{\upeta }}_g}$, ${{\boldsymbol{\upeta }}_\delta}$, and statistical parameters of $\gamma$), full or modular Bayesian approaches have been developed \citep{arendt2012quantification,kennedy2001bayesian}.

\hspace{0.4cm} It has been shown in various applications that the KOH framework is more effective in general than the other two categories of approaches when quantifying model uncertainty and improving prediction accuracy of computer simulation models  \citep{jiang2020sequential,arendt2012quantification,xiong2009better}. The implementation of the KOH framework, however, is much more complicated than its counterparts due to the higher number of unknowns to be estimated. Additionally, the accuracy of the KOH framework could be affected by the prior distributions of ${\boldsymbol{\uptheta }}$ as shown in \cite{jiang2020sequential,arendt2012quantification}, since the KOH framework is fundamentally a Bayesian method, of which prior distribution is a vital part.
\end{itemize}

The above reviewed three categories of methods have been applied to various computer simulation models, including static, quasi-static, and dynamic models. Since digital models of a digital twin usually are dynamic, next, we summarize variants of the above three categories of methods for dynamic system models. When the digital models are formulated in a state-space form as given in Eq. (9) of our Part 1 paper for model updating in the P2V connection (see Sec. 4.2 of Part 1 on probabilistic model updating) and control in the V2P connection (see Sec. 5.1 of Part 1 on model predictive control), the above reviewed three categories of methods could be applied to either the state transition equation or the measurement equation. According to which equation in a state-space model that the quantification of model uncertainty method is applied to, we classify the existing methods into two groups, namely (1) UQ of measurement equation, and (2) UQ of state transition equation (governing equations).

\paragraph{(b) UQ of measurement equations}

Let us now look at UQ of dynamic system models, in particular state-space models such as the ones in Eqs. (3), (8), and (9) of our Part 1 paper. A typical state transition equation, ${{\bf{x}}_k} = {\bf{f}}({{\bf{x}}_{k - 1}},\;{{\bf{u}}_{k - 1}}) + {{\boldsymbol{\upomega }}_k}$, is a vector of difference state transition functions plus a vector of noises. Given an  initial estimate of ${{\bf{x}}_0}$, the outputs of the state transition equation at time step $k$ (i.e., state variables ${{\bf{x}}_k}$) are essentially functions of exogenous inputs ${\bf{u}}_{0:k}$ or functions of ${\bf{u}}_{0:k}$ and  uncertain model parameters ${\boldsymbol{\uptheta }}$ if ${\boldsymbol{\uptheta }}$ is also considered in state transition \citep{beck1998updating}. Therefore, ${{\bf{x}}_k}$ can be represented as ${{\bf{x}}_k} = {\bf{F}}({{\bf{u}}_{0:k}},\;{{\boldsymbol{\uptheta }}}) + {{\boldsymbol{\upepsilon }}_k}$, where ${\bf{F}}(\cdot)$ are numerical solutions to the recursion using ${\bf{f}}(\cdot)$ and ${{\boldsymbol{\upepsilon }}_k}$ are the residual model form errors caused by this representation. ${\bf{F}}(\cdot)$ do not have close form expressions for most problems and need to be solved numerically. Moreover, the measurement function ${{\bf{g}}({{\bf{x}}_k})}$ are quasi-static models with state variables ${{\bf{x}}_k}$ as the input. If we embed the state transition function ${\bf{f}(\cdot)}$ in Eq. (9) of our Part 1 paper or ${\bf{F}}({{\bf{u}}_{0:k}},\;{{\boldsymbol{\uptheta }}})$ into the measurement function ${\bf{g}(\cdot)}$, the overall dynamic system model as a whole can be written as
\begin{equation}
\label{eq:measurement_equation_UQ1}
\begin{split}
  & {{\bf{y}}_k} = {\bf{g}}({\bf{F}}({{\bf{u}}_{0:k}},\;{\boldsymbol{\uptheta }})) + {{\boldsymbol{\upgamma }}_k},  \cr 
  & \;\;\;\; = {\bf{h}}({{\bf{u}}_{0:k}},\;{\boldsymbol{\uptheta }}) + {{\boldsymbol{\upgamma }}_k}, \cr
\end{split}
\end{equation}
in which ${\bf{h}(\cdot)}$ is a new function created by implicitly embedding ${\bf{f}(\cdot)}$ into ${\bf{g}(\cdot)}$, and ${{\boldsymbol{\upgamma}}_k}$ is a vector of Gaussian noise variables with either an unknown mean vector and covariance matrix or zero mean and an unknown covariance matrix. The initial conditions are omitted from the above equation to simplify the notations. Note that ${{\boldsymbol{\upgamma}}_k}$ is used here to account for observation noises and all residual model uncertainty that is not accounted for by the uncertain model parameters ${{\boldsymbol{\uptheta }}}$.

Based on the representation given in Eq.(\ref{eq:measurement_equation_UQ1}) and to facilitate the quantification of model uncertainty, the state-space model given in Eq. (9) of our Part 1 paper is re-formulated similar to Eq. (8) of Part 1 as \citep{astroza2019dual,song2019accounting,burns2018identification}
\begin{equation}
\label{eq:measurement_equation_UQ}
\begin{split}
  & {\rm{State}}\;{\rm{transition}}:{{\boldsymbol{\uptheta }}_k} = {{\boldsymbol{\uptheta }}_{k - 1}} + {{\bf{r}}_k},  \cr 
  & {\rm{Measurement}}:\;{{\bf{y}}_k} = {\bf{h}}({{\bf{u}}_{0:k}},\;{{\boldsymbol{\uptheta }}_k}) + {{\boldsymbol{\upgamma}}_k}, \cr
\end{split}
\end{equation}
where ${{\bf{y}}_k} \in \mathbb{R}^{n_y \times 1}$ is a vector of observations at $t_k$, ${\bf{h}(\cdot)}$ is the new measurement function as mentioned above. When the uncertain model parameters ${{\boldsymbol{\uptheta }}}$ do not change with time, Eq. (\ref{eq:measurement_equation_UQ}) will reduce to ${{\bf{y}}_k} = {\bf{h}}({{\bf{u}}_{0:k}},\;{{\boldsymbol{\uptheta }}}) + {{\boldsymbol{\upgamma}}_k}$.

One may notice that the measurement equation in Eq. (\ref{eq:measurement_equation_UQ}) or the reduced form of Eq. (\ref{eq:measurement_equation_UQ}) (i.e., ${{\bf{y}}_k} = {\bf{h}}({{\bf{u}}_{0:k}},\;{{\boldsymbol{\uptheta }}}) + {{\boldsymbol{\upgamma}}_k}$) is very similar to the equations given in the aforementioned three categories of UQ methods (i.e., Eqs. (\ref{eq:model_uncertainty1})-(\ref{eq:KOH_equation}) in Sec. \ref{sec:UQ_dynamic_system_models} (a)). This similarity is beneficial as it allows us to quantify the uncertainty of dynamic system models by directly using methods originally developed for static models. 
\begin{itemize}
    \item For instance, based on the formulation in Eq. (\ref{eq:measurement_equation_UQ}), \cite{astroza2019dual,song2019accounting,behmanesh2017probabilistic} suggested several approaches for the quantification of model uncertainty of structural dynamic system models. Since ${\boldsymbol{\uptheta }}$ and ${{\boldsymbol{\upgamma}}_k}$ are used to account for all possible model uncertainty in their methods, those approaches can be classified as the \emph{category 1} method (see Sec. \ref{sec:UQ_dynamic_system_models} (a)). Moreover, in dynamic system models, the uncertain model parameters ${\boldsymbol{\uptheta }}$ change very slowly or do not change with time, while statistical parameters of ${{\boldsymbol{\upgamma}}_k}$ change relatively faster due to the variability of model form uncertainty over time along with the exogenous inputs. The quantification of model uncertainty using the \emph{category 1} methods based on the formulation given in Eq. (\ref{eq:measurement_equation_UQ}) is, therefore, very similar to the state and parameter estimation discussed in Sec. 4.2.4 of our Part 1 paper. The difference in the time scales of ${\boldsymbol{\uptheta }}$ and statistical parameters of ${{\boldsymbol{\upgamma}}_k}$ needs to be considered in the UQ process. To this end, \cite{astroza2019dual} applied dual Kalman filter to simultaneously estimate ${\boldsymbol{\uptheta }}$ and the diagonal elements of the covariance matrix of ${{\boldsymbol{\upgamma}}_k}$ over time. \cite{song2019accounting,behmanesh2017probabilistic} employed hierarchical Bayesian updating methods to estimate ${\boldsymbol{\uptheta }}$ and parameters of ${{\boldsymbol{\upgamma}}_k}$. Since the \emph{category 1} methods convert the quantification of model uncertainty problem into a standard Bayesian updating problem, the formulation given in Eq. (\ref{eq:measurement_equation_UQ}) makes it possible to perform online model-uncertainty quantification using various Bayesian inference methods reviewed in Sec. 4.2 of Part 1.
    \item The \emph{category 3} methods (i.e., the KOH framework reviewed in Sec. \ref{sec:UQ_dynamic_system_models} (a) and its variants) have also been applied to quantify model uncertainty of the measurement equation based on the formulation given in Eq. (\ref{eq:measurement_equation_UQ}). For example, by following the KOH framework, \cite{burns2018identification,ramancha2022accounting,burns2014parameter,ward2021continuous} added a model discrepancy term to the measurement equation in addition to ${\boldsymbol{\uptheta }}$ and ${{\boldsymbol{\upgamma}}_k}$ to quantify model uncertainty. To address the computational challenge introduced by the KOH framework, \cite{burns2018identification,burns2014parameter} used a parametric function as model discrepancy term such that the problem becomes to be a standard Bayesian updating problem which is similar to the category 1 methods. \cite{ramancha2022accounting} assumed the dynamic system model to be linear, which, as a result, reduced the computational burden required in applying the KOH framework. \cite{ward2021continuous} compared a variant of the KOH framework based on particle filter against a sequential KOH approach in the context of digital twins. They concluded that the computational effort required by the sequential KOH framework to track time-varying model parameters is high, which makes it not suitable for online updating in digital twin applications. The particle filter-based variant is computationally cheaper than the sequential KOH framework for digital twins \citep{ward2021continuous}.
\end{itemize}

\begin{figure*}[!ht]
 \centering
    {\includegraphics[scale=0.61]{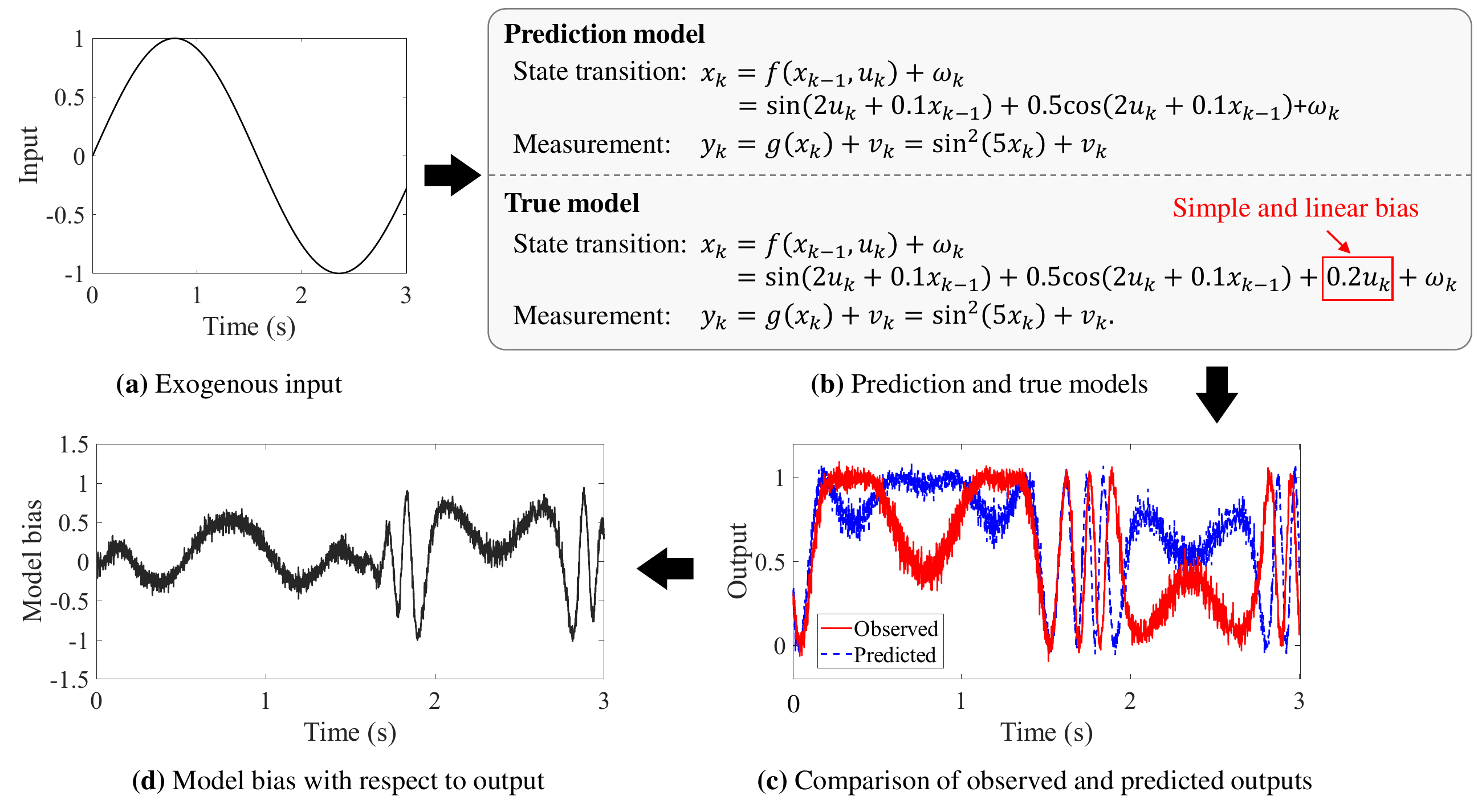}}
 \caption{Illustration of how a simple linear model bias in the state transition equation could be translated into a highly nonlinear model discrepancy of the dynamic system output}
 \label{fig:bias_dynamic}
\end{figure*}

In summary, the benefit of formulating the state-space model as Eq. (\ref{eq:measurement_equation_UQ}) is that it allows us to directly apply the three categories of UQ methods reviewed in Sec. \ref{sec:UQ_dynamic_system_models} (a) to quantify model uncertainty of the measurement equation \citep{ward2021continuous, astroza2019dual,song2019accounting,behmanesh2017probabilistic,burns2018identification,burns2014parameter,ramancha2022accounting}. The disadvantage is that the non-linearity of the new measurement function ${\bf{h}(\cdot)}$ could be much higher than that of ${\bf{g}(\cdot)}$ or ${\bf{f}(\cdot)}$. As a result, the model discrepancy of the measurement function (${\bf{h}(\cdot)}$) given in Eq. (\ref{eq:measurement_equation_UQ}) would be more difficult to be quantified than that of ${\bf{g}(\cdot)}$ or ${\bf{f}(\cdot)}$ in the state-space model given in Eq. (9) of our Part 1 paper. For instance, as illustrated in Fig. \ref{fig:bias_dynamic}, a very simple linear bias term in the state transition equation (${x_k=f(x_{k-1},u_k)+\omega_k}$) (see Fig. \ref{fig:bias_dynamic} (b)) could be translated into a highly nonlinear model discrepancy between the observation and the predicted output of the dynamic system model (i.e., model discrepancy of ${h(\cdot)}$, as illustrated in Fig. \ref{fig:bias_dynamic} (d)). In that case, it is more preferable to quantify model uncertainty of the state transition equation using the state-space model given in Eq. (9) of our Part 1 paper than that of the measurement equation using the formulation given in Eq. (\ref{eq:measurement_equation_UQ}). 

Next, we briefly summarize methods for the quantification of model uncertainty of the state transition equation.

\paragraph{(c) UQ of state transition equations}
We now assume that the measurement equation in a state-space model is adequately modeled, and we mainly focus on UQ of the state transition equation. This assumption holds for many problems since the measurement equation is usually simpler than the state transition equation. If this assumption does not hold (i.e., the measurement equation has a large model bias), a two-step process can be followed. Since the measurement equation is quasi-static in nature, it can be corrected first using methods for static models based on data collected in a controlled environment \citep{xi2019learning}. Following this first step, model uncertainty of the state transition equation can be quantified using the methods reviewed below. It is worth noting that, since the state transition equation models the transition of state variables over time (e.g., rate of change of state variables) and governs the dynamics of a dynamic system, \cite{subramanian2019error} referred to the bias of the state transition equation as \lq\lq model form error" and the resulting discrepancy between observation and prediction of the model output as \lq\lq model discrepancy". They also pointed out an important distinction between UQ of measurement equation and UQ of state transition equation (governing equation) that the recovery of the missing physics in the state transition equation allows for improving the prediction accuracy of the state-space model for extrapolation while it is difficult to achieve this purpose by just performing UQ of the measurement equation.

Numerous methods have been proposed in recent years to quantify model uncertainty of state transition equations \citep{wilkinson2011quantifying,zhang2019dynamic,subramanian2019error,hu2019model,viana2021estimating,jiang2022model,yucesan2020physics}. These methods can be classified as the \emph{category 2} methods reviewed in Sec. \ref{sec:UQ_dynamic_system_models} (a), since a model discrepancy function and a noise term are used to account for model uncertainty of the governing equation. After  model uncertainty is accounted for using a \emph{category 2} method, the state transition equation (governing equation) given in Eq. (9) of our Part 1 paper becomes
\begin{equation}
\label{eq:state_transition_UQ}
\begin{split}
&{{\bf{x}_{k}}} = {\bf{f}}({{\bf{x}}_{k-1}},\;{\boldsymbol{\uptheta }^*},\; {{\bf{u}}_{k-1}}) \cr 
& + \hat{\boldsymbol{\updelta}}({{\bf{x}}_{k-1}},{{\bf{u}}_{k-1}}; {{\boldsymbol{\upeta }}_\delta }) + {\boldsymbol{\upgamma}_k}, \cr
\end{split}
\end{equation}
where $\hat{\boldsymbol{\updelta}}({{\bf{x}}_{k-1}},{{\bf{u}}_{k-1}}; {{\boldsymbol{\upeta }}_\delta })$ is the model discrepancy function with unknown model parameters ${{\boldsymbol{\upeta }}_\delta }$, similar to Eq. (\ref{eq:measurement_equation_UQ}), ${{\boldsymbol{\upgamma}}_k}$ is a vector of Gaussian noise variables with either an unknown mean vector and covariance matrix or zero means and an unknown covariance matrix.

According to how the unknown parameters ${{\boldsymbol{\upeta }}_\delta}$ of the model discrepancy function are estimated, methods of this group can be further divided into two sub-groups.

\begin{itemize}
  \item The first sub-group sets ${{\boldsymbol{\upgamma }}_k}$ as ${{\boldsymbol{\upomega }}_k}$ which is the process noise of the original state transition equation given in Eq. (3) or (9) of our Part 1 paper. After that, model bias ${\boldsymbol{\updelta }}_k$ at each time instant $t_k$ is estimated along with state variables $\textbf{x}_k$ using one of the Bayesian filters given in Sec. 4.2.2 of Part 1 on state estimation and Bayesian filters. Based on the estimated ${\boldsymbol{\updelta }}_k$, a predictive model $\hat {\boldsymbol{\updelta }}({{\bf{x}}},\;{{\bf{u}}}; {{\boldsymbol{\upeta }}_\delta })$ is constructed to correct the state transition equation. To account for uncertainty in the estimated ${\boldsymbol{\updelta }}_k$ and uncertainty introduced by setting ${{\boldsymbol{\upgamma }}_k}$ as ${{\boldsymbol{\upomega }}_k}$, a probabilistic predictive model such as a Gaussian process regression model is usually constructed as $\hat {\boldsymbol{\updelta }}({{\bf{x}}},\;{{\bf{u}}}; {{\boldsymbol{\upeta }}_\delta })$. Examples of such methods include \cite{subramanian2019error,zhang2019dynamic,hu2019model}.
  \item The second sub-group treats ${{\boldsymbol{\upgamma }}_k}$ as a vector of Gaussian noise variables with either an unknown mean vector and covariance matrix or zero means and an unknown covariance matrix (i.e., the same treatment as Eq. (\ref{eq:measurement_equation_UQ})). The unknown distributional parameters of ${{\boldsymbol{\upgamma }}_k}$ are then estimated along with unknown parameters ${{\boldsymbol{\upeta }}_\delta }$ of $\hat {\boldsymbol{\updelta }}({{\bf{x}}},\;{{\bf{u}}}; {{\boldsymbol{\upeta }}_\delta })$. To enable for the end-to-end training of a data-driven model of $\hat {\boldsymbol{\updelta }}({{\bf{x}}},\;{{\bf{u}}}; {{\boldsymbol{\upeta }}_\delta })$, $\hat {\boldsymbol{\updelta }}({{\bf{x}}},\;{{\bf{u}}}; {{\boldsymbol{\upeta }}_\delta })$ needs to be integrated with the original state-space model (i.e., more specifically the original state transition equation) in the training process. Examples of this sub-group include \cite{hu2019model,yucesan2020physics,jiang2022model,wilkinson2011quantifying,viana2021estimating}. This sub-group of methods is very similar to Approach 4 (i.e., delta learning) of physics-informed ML in Sec. 3.4 of Part 1, which uses a data-driven ML model as $\hat \delta(\cdot)$ to recover the unmodeled physics.
\end{itemize}

Since the above two subgroups of methods can be classified as the \emph{category 2} methods discussed in Sec. \ref{sec:UQ_dynamic_system_models} (a), they inherit the advantage of the category 2 methods that the predictive capability of the model discrepancy function ${\hat {\boldsymbol{\updelta }}}({\bf{x}},\;{\bf{u}}; {{\boldsymbol{\upeta }}_\delta })$ can help improve the prediction accuracy of the state transition equation under previously unseen conditions. Constructing an accurate model of ${{\boldsymbol{\updelta }}}({\bf{x}},\;{\bf{u}}; {{\boldsymbol{\upeta }}_\delta })$, however, can be very challenging since the output of the state transition equation is time-varying and not directly measurable.

From the above review, we can conclude that most of the current UQ methods for dynamic system models implement either the \emph{category 1} methods on measurement equations (Sec. \ref{sec:UQ_dynamic_system_models} (b)) or the \emph{category 2} methods on state transition equations (Sec. \ref{sec:UQ_dynamic_system_models} (c)). Only a few methods apply the KOH framework (i.e., the \emph{category 3} methods) to dynamic system models based on the formulation given in Eq. (\ref{eq:measurement_equation_UQ}) (see Sec. \ref{sec:UQ_dynamic_system_models} (b)).

As illustrated in Fig. \ref{fig:bias_dynamic}, a small bias in the state transition equation could escalate as a highly nonlinear model discrepancy in the measurement equation, especially for a nonlinear dynamic system model. It implies that the reformulation of the state-space model in Eq. (\ref{eq:measurement_equation_UQ}) could significantly increase the difficulty in quantifying model uncertainty. Since the state transition equation governs the dynamics of a dynamic system, it is envisioned that quantifying model uncertainty of the state transition equation (Sec. \ref{sec:UQ_dynamic_system_models} (c)) could be more effective in improving the prediction accuracy of a state-space model than quantifying model uncertainty of the measurement equation (Sec. \ref{sec:UQ_dynamic_system_models} (b)). Given that the \emph{category 1} and \emph{category 2} methods for UQ of dynamic system models are getting mature, it is worth investing more research effort in the \emph{category 3} methods. We expect such an increased investment will yield newer, more mature \emph{category 3} methods that can help further improve the validity of digital models in digital twins, especially for the state transition equation. 

The quantification of model uncertainty of dynamic system models could improve the accuracy and robustness of MPC by improving the prediction accuracy of state-space models \citep{rohrs1982robustness,rohrs1985robustness,liu2002decentralized,li2016adaptive}, as has been discussed in Sec. 5.1 of our Part 1 paper on MPC, and enable model-based risk assessment for decision making under uncertainty. It plays a vital role in ensuring the effectiveness of digital twins in personalized control and optimization.



\subsection{Optimization for digital twins (OPT)}
\label{sec:optimization}

The role of optimization in digital twins can be classified into two categories: offline optimization and online optimization (as illustrated in Fig. 3 of our Part 1 paper). Offline optimization occurs prior to the deployment of a digital twin. Online optimization takes place when a digital twin has been deployed and is in operation. In the subsequent sections, we briefly discuss various optimization techniques used for digital twins.

\subsubsection{Optimization for sensor placement (offline)}
\label{sec:opt_sensor_placement}
Sensing is the forefront of the P2V connection and an indispensable element of a digital twin. Many different types of sensors, such as strain gauges, acoustic emission sensors, thermal cameras, optical cameras, and others, can be employed to collect data capturing different aspects of the physical system performance in-situ. Sensor data collected from a physical system serves as the inputs to the twinning enabling techniques reviewed in Sec. 4 of Part 1 paper, and are essential to establishing the P2V connection. 

No matter what type of sensor is used, an essential question that needs to be answered is where the sensors should be placed in the physical system. The locations where the sensors are placed could significantly affect the quality of the collected data and ultimately the inference of other information, which would affect the effectiveness of the P2V connection and, eventually, the performance of the digital twin. Therefore, it is particularly important to optimize the sensor placement at the design stage before online deployment of a digital twin. Moreover, it is worth mentioning that the number of sensors and sensor types can also be treated as design variables in sensor network design. That would add another level of complexity to sensor network design since sensor network performance would be conditional on the number of sensors and sensor types. Several studies have been conducted in recent years to address sensor network design at this higher level of complexity. For example, some studies consider the number of sensors as another design variable in addition to sensor locations. \cite{yang2020strategy} treated the number of sensors and sensor locations as design variables in a genetic algorithm-based sensor network design method. Similarly, \cite{an2022methodology,an2022optimal} optimized the number of sensors and sensor locations simultaneously using a non-dominated sorting genetic algorithm II in sensor network design for vibration-based damage detection. While optimizing the number of sensors and sensor types is also important for sensor network design, this section intentionally concentrates on sensor placement optimization, since it is fundamental to various sensor network design problems.

As mentioned in Sec. \ref{sec:UQ_digital_twin}, two types of uncertainty sources, namely aleatory and epistemic, are presented in digital twins. Epistemic uncertainty could be reduced through the data collected from sensors in the P2V connection. Aleatory uncertainty, however, is irreducible and is inherent in a digital twin. To ensure that the sensors collect the most informative data in the presence of natural variability (i.e., aleatory uncertainty), it is important to consider aleatory uncertainty in the optimization of sensor placements. To this end, a generalized model for sensor placement optimization under uncertainty can be formulated as follows
\begin{equation}
\label{eq:sensor}
\begin{split}
{{\bf{d}}^*} = \mathop {\arg \max }\limits_{{\bf{d}} \in \Omega } \{ L({\bf{d}},\;{\boldsymbol{\upalpha}})\} ,
\end{split}
\end{equation}
where ${\bf{d}}$ is a vector or matrix consisting of sensor locations such as the coordinates where the sensors are placed, $\Omega$ is the spatial design domain, ${\boldsymbol{\upalpha}}$ is a vector of random variables representing the aleatory uncertainty during the operation of a physical system and its digital twin, and $L({\bf{d}},\;{\boldsymbol{\upalpha}})$ is a cost function of ${\bf{d}}$ and ${\boldsymbol{\upalpha}}$. As mentioned above, accounting for aleatory uncertainty ${\boldsymbol{\upalpha}}$ in the sensor placement optimization is vital to ensuring that the designed sensor network can well perform its intended function when the digital twin is put into online operation.

Three key research questions usually need to be addressed in solving the above sensor network design optimization model:
\begin{enumerate}
    \item How to formulate the cost function $L({\bf{d}},\;{\boldsymbol{\upalpha}})$?
    \item How to efficiently and accurately evaluate the cost function in the presence of uncertainty?
    \item How to efficiently solve the optimization model given in Eq. (\ref{eq:sensor})?
\end{enumerate}

In what follows, we briefly review commonly used approaches to tackle the above three research questions.

\paragraph{(a) Cost function}
The cost function $L({\bf{d}},\;{\boldsymbol{\upalpha}})$ needs to be formulated in consideration of the P2V connection and the sensor type. 

For instance, for a network of wireless sensors, the cost function could be formulated as the resilience or vulnerability of the sensor network and needs to consider the routing algorithm for effective communication among different wireless sensors \citep{anand2005quantifying}. Since wireless sensors are usually self-powered, energy efficiency has also been an important consideration in the cost function for sensor network optimization \citep{sachan2012energy}. A few representative review papers about various cost functions and the corresponding optimization models of wireless sensor networks are available in \cite{kulkarni2010particle,adnan2013bio,asorey2017survey}.

For wired sensors, many performance metrics have been proposed in the past decades to optimize their placement. The commonly used cost functions can be roughly grouped into the following categories
\begin{itemize}
    \item {\bf{Information gain}}: This class of metrics/cost functions measures the amount of information contained in the data collected from a sensor network design for uncertainty reduction \citep{nath2017sensor,yang2021probabilistic,gomes2019multiobjective,hu2017calibration,meo2005optimal,kammer1991sensor}. Various metrics have been proposed in the information science domain to quantify information gain from data. The most widely used ones in sensor placement optimization include 
    \begin{enumerate}
        \item \emph{Fisher's information matrix}: It quantifies the information gain based on the assumption that the posterior distribution is a multivariate Gaussian distribution \citep{gomes2019multiobjective,hu2017calibration,meo2005optimal,kim2018development,heydari2020optimal}. Some examples of cost functions for sensor network design based on the Fisher's information matrix include A-optimality criterion (trace), D-optimality criterion (determinant), and E-optimality criterion (largest eigenvalue)  \citep{gomes2019multiobjective,kammer1991sensor, hu2017calibration,kim2018development,heydari2020optimal}. For instance, \cite{kim2018development} proposed a stochastic effective independence method for optimal sensor placement with A-optimality criteria. This method showed better performance in handling system uncertainty compared to an existing method. Another study along the same line is \cite{heydari2020optimal}, where the authors used D-optimality criterion to optimize sensor placement for source localization based on the received signal strength difference.
        \item \emph{Kullback–Leibler divergence}: It quantifies the amount of information gained from data using the relative entropy. When the Kullback–Leibler divergence is used in a sensor network design with the consideration of various uncertainty sources, the cost function is formulated as \citep{nath2017sensor,yang2021probabilistic}
        \begin{equation}
        \label{eq:EKL}
        \begin{split}
          & L({\bf{d}},\;{\boldsymbol{\upalpha}}) = \int_{{\Omega _{\boldsymbol{\uptheta}}}} {\int_{{\Omega _{\boldsymbol{\upalpha}}}} {{D_{KL}}(y({\boldsymbol{\upalpha}},\;{\bf{d}}),\;{\boldsymbol{\upalpha}})} }   \cr 
          & \times {p_{Y\vert{\boldsymbol{\uptheta}}}}(y({\boldsymbol{\upalpha}},\;{\bf{d}})\vert{\boldsymbol{\upalpha}},\;{\boldsymbol{\uptheta}}){f_{\boldsymbol{\upalpha}}}({\boldsymbol{\upalpha}}){f_{\boldsymbol{\uptheta }}}({\boldsymbol{\uptheta}}){\bf{d}{\boldsymbol{\upalpha}}\bf{d}\boldsymbol{\uptheta}}, \cr
        \end{split}
        \end{equation}
        where ${{D_{KL}}(y({\boldsymbol{\upalpha}},\;{\bf{d}}),\;{\boldsymbol{\upalpha}})}$ is the Kullback–Leibler divergence for given realization of ${\boldsymbol{\upalpha}}$ and synthetic observations $y({\boldsymbol{\upalpha}},\;{\bf{d}})$ which are generated using physics-based modeling (see Sec. 3.2 of Part 1), $\boldsymbol{\uptheta}$ is a vector of epistemic model parameters (see Sec. 4.2.4 of Part 1), such as model parameters of Paris's law for crack growth or capacity of a battery, ${f_{\boldsymbol{\upalpha}}}({\boldsymbol{\upalpha}})$ and ${f_{\boldsymbol{\uptheta }}}({\boldsymbol{\uptheta }})$ are respectively the joint probability density function of $\boldsymbol{\uptheta}$ and ${\boldsymbol{\upalpha}}$.
        
        \hspace{0.4cm} It is worth mentioning that the Kullback–Leibler divergence is one type of $f-$divergence. Other types of $f-$divergences can also be used to quantify the information gain. A comparison of different $f-$divergences for sensor network design optimization is given in \cite{yang2021probabilistic}.
    \end{enumerate}
    
    \item {\bf{Probability of detection}}: Cost functions falling within this group aim to minimize the type I and type II errors \citep{downey2018optimal,flynn2010bayesian,guratzsch2010structural,wang2015probabilistic}. The type I error is related to the scenario that a healthy (undamaged) state is incorrectly identified as damaged, i.e., a false alarm. The type II error is related to the probability that a damaged state is classified as healthy (undamaged), i.e., false negative, missed detection, or error of omission. For instance, \cite{downey2018optimal} optimized the placement of sensors used in the construction of accurate strain maps for large-scale structural components by minimizing the type I and II errors. \cite{flynn2010bayesian} proposed Bayes risk-based function for sensor placement optimization by associating decision costs with the type I and II errors. Similarly, the probability of detection has been employed as a metric for sensor placement optimization in \cite{guratzsch2010structural,wang2015probabilistic}, where aleatory uncertainty in the operation of physical systems was explicitly accounted for via probabilistic analysis.
    
    \item {\bf{Modal assurance criterion}}: Modal assurance criterion is a metric that is widely used in structural dynamics domain to quantify the similarity of mode shapes \citep{allemang2003modal}. It has also been applied to sensor placement optimization for SHM. For example, \cite{carne1994modal} proposed an approach to determine the optimal number and locations of sensors, where the modal assurance criterion was used to correlate a modal test with an FEA model. Following the work of \cite{carne1994modal}, \cite{yi2011optimal} minimized the off-diagonal elements of the modal assurance criterion matrix to optimize the sensor configuration for SHM. \cite{an2022methodology} considered model uncertainty in the root mean square error derived from the off-diagonal elements of a modal assurance criterion matrix in sensor network design for vibration-based damage detection.

    \item {\bf{Value of information (VoI)}}: VoI has emerged as a cost framework for sensor network design optimization in recent years and has gained much attention in a broad range of domains \citep{bisdikian2013quality,malings2016value,basagni2014maximizing,cantero2020optimal,chadha2021alternative}. This metric is particularly attractive because it directly quantifies the expected VoI of the data collected from a particular location and/or time by considering various costs associated with decision alternatives. The generalized form of the expected VoI for a sensor network design is defined as \citep{malings2016value, chadha2021alternative}
    \begin{equation}
    \label{eq:EVOI}
    {\rm{EVOI}}({\bf{d}}) = {\Psi _{\textrm{prior}}}({{\boldsymbol{\upchi}}_{\textrm{prior}}}) - {\Psi _{{\boldsymbol{\upalpha}},\;{\boldsymbol{\uptheta }}}}({{\boldsymbol{\upchi }}_{\bf{d}}},\;{\bf{d}}),
    \end{equation}
    where ${\Psi _{\textrm{prior}}}({{\boldsymbol{\upchi }}_{\textrm{prior}}})$ is the cost associated with the decision ${{\boldsymbol{\upchi }}_{\textrm{prior}}}$ by only considering the prior information of the epistemic uncertain parameters ${\boldsymbol{\uptheta }}$, ${\Psi _{{\boldsymbol{\upalpha}},\;{\boldsymbol{\uptheta }}}}({{\boldsymbol{\upchi }}_{\bf{d}}},\;{\bf{d}})$ is the expected cost associated with optimal decision ${{\boldsymbol{\upchi }}_{\bf{d}}}$ based on pre-posterior analysis of ${\boldsymbol{\uptheta }}$ for a sensor placement design ${\bf{d}}$ and with the consideration of other aleatory uncertain variables ${\boldsymbol{\upalpha}}$ in decision making.
\end{itemize}

Note that the above reviewed metrics are not exhaustive, but representative. Interested readers can find more comprehensive reviews on various metrics for sensor placement optimization in \cite{ostachowicz2019optimization,tan2020computational,gupta2010optimization}.  

\paragraph{(b) UQ methods}
As mentioned above, uncertainty sources need to be considered in the cost function since offline sensor placement optimization is performed before the online deployment of a digital twin. While the consideration of uncertainty sources could ensure the performance of the designed sensor network in collecting the most useful information after the deployment, it poses significant computational challenges to the evaluation of the cost functions. Efficient and accurate UQ methods have been developed to tackle the computational challenge and can be categorized into two classes.

\begin{itemize}
\item  {\bf{Analytical or numerical approximation}}: This class of methods approximates the cost function in the presence of uncertainty using analytical expressions \citep{long2013fast} or numerical approximations \citep{yang2021probabilistic,guratzsch2010structural,wang2015probabilistic}. For example, due to the lack of an analytical solution to the Kullback–Leibler divergence and the required high-dimensional integration to compute the expected value, solving Eq. (\ref{eq:EKL}) is generally computationally demanding. To address the computational challenge, analytical approximation of the Kullback–Leibler divergence has been pursued using Laplace approximations \citep{long2013fast}. Motivated by tackling the same computational challenge in sensor placement optimization, \cite{yang2021probabilistic} approximated the high-dimensional integration in Eq. (\ref{eq:EKL}) using univariate dimension reduction. When the probability of detection is employed as the cost function and needs to be estimated using physics-based probabilistic analysis, Monte carlo sampling-based approximations using finite element simulations have been developed in \cite{guratzsch2010structural} and \cite{wang2015probabilistic}.
\item  {\bf{Surrogate-based approximation}}: The first step in this class of methods is to construct a surrogate of 
\begin{itemize}
\item {\it{either}} the physics-based simulation model that is used to generate synthetic observations for sensor placement optimization \citep{huan2014gradient,eshghi2019design}, 
\item {\it{or}} the cost function with respect to uncertain variables \citep{nath2017sensor,an2022optimal}.
\end{itemize}
After the construction of the surrogate, Monte Carlo simulation is employed to evaluate the cost function. For instance, polynomial chaos expansion and Gaussian process regression surrogates have been built to replace the original physics-based models for the evaluation of the Kullback–Leibler divergence \citep{huan2014gradient} and probability of detection \citep{eshghi2019design}, respectively. For the direct surrogate modeling of cost functions, \cite{nath2017sensor} constructed a  Gaussian process regression model (surrogate) of the Kullback–Leibler divergence, making it possible to efficiently compute the expected Kullback–Leibler divergence in sensor placement for calibration of spatially varying model parameters. \cite{an2022optimal} built a Gaussian process regression model for the determinant of the Fisher information matrix and then 
computed the mean and standard deviation of the determinant using Monte Carlo simulation based on the surrogate model.
\end{itemize}

\paragraph{(c) Optimization methods}
Once an appropriate cost function $L({\bf{d}},\;{\boldsymbol{\upalpha}})$ is established and the uncertainty of the cost function is quantified, the last key research question is how to efficiently solve the optimization model formulated in Eq. (\ref{eq:sensor}). Current optimization methods for sensor placement optimization can in general be grouped into three categories as follows.

\begin{figure*}[!h]
  \centering
    {\includegraphics[scale=0.8]{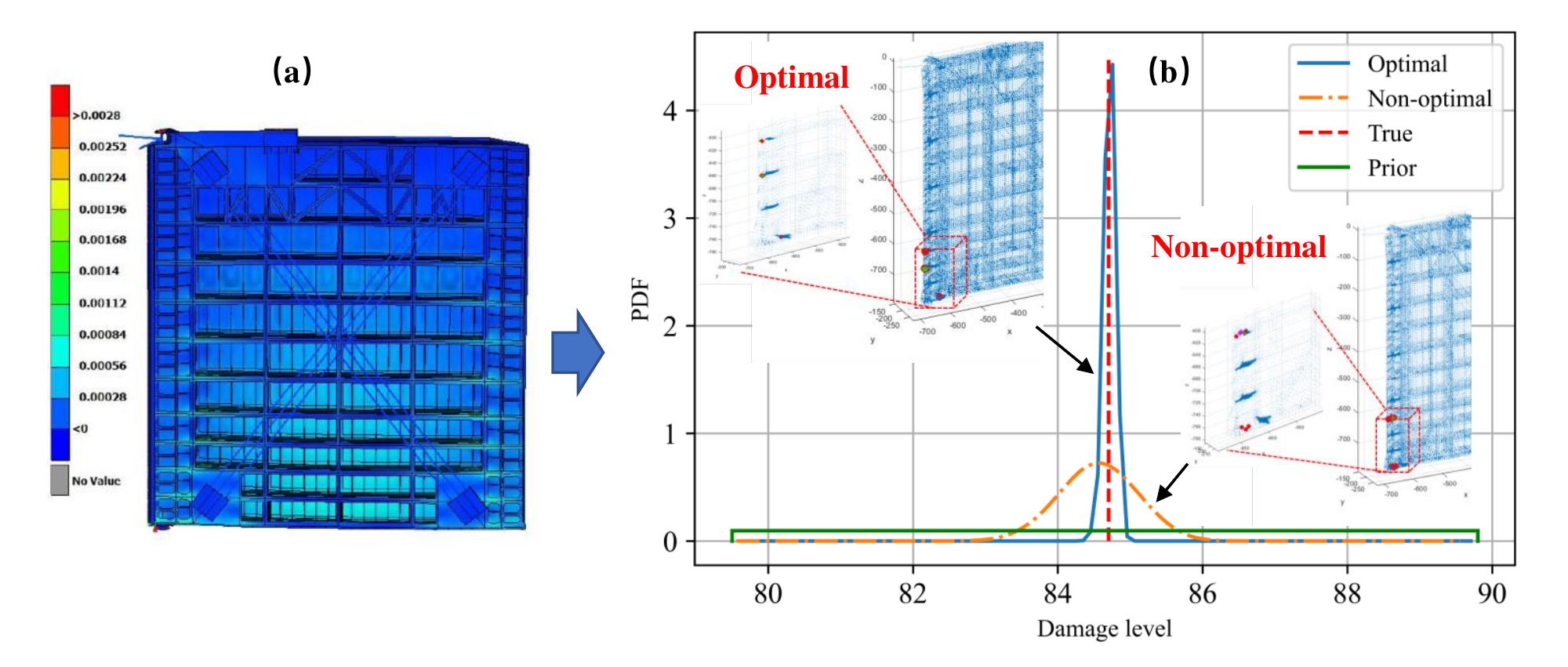}}
  \caption{Impact of sensor network design on probabilistic model updating (P2V connection). (a) Finite element simulation of a miter gate; (b) Comparison of posterior distributions of damage level obtained from the optimal sensor design and a non-optimal design. \citep{yang2021probabilistic}}
  \label{fig:posterior}
\end{figure*}

\begin{itemize}
\item {\bf{Evolutionary optimization methods}}: The optimization problem could be solved by an evolutionary optimization method directly, such as a genetic algorithm \citep{yao1993sensor,liu2008optimal,yi2011optimal,an2022methodology,an2022optimal,an2022methodology,an2022optimal,ehsani2010optimization,downey2018optimal,flynn2010bayesian}, simulated annealing \citep{tong2014optimal,chen1991optimal}, or particle swarm optimization \citep{zhang2014optimal,li2015optimal}, if it is computationally cheap to evaluate the cost function (e.g., probability of detection with analytical expressions, modal assurance criterion) and provided that the number of sensors is small. For instance, genetic algorithm has been employed for sensor placement optimization by using modal assurance criterion \citep{an2022methodology,an2022optimal} or probability of detection \citep*{ehsani2010optimization,downey2018optimal,flynn2010bayesian} as cost function since they can be evaluated very efficiently.
\item  {\bf{Greedy algorithm-based methods}}: In contrast, as the number of sensors increases or the cost function is increasingly computationally demanding to evaluate, such as the expected Kullback–Leibler divergence given in Eq. (\ref{eq:EKL}), it would be computationally intractable to directly solve Eq. (\ref{eq:sensor}) with evolutionary optimization methods, since evolutionary optimization methods usually need to evaluate the cost function thousands of times to find a near-optimal solution. To overcome this challenge, methods have been developed using greedy algorithms in conjunction with efficient UQ methods discussed in Sec. 2.2.1 (b) \citep{yang2021probabilistic,nath2017sensor,sela2018robust,malings2015sensor,blachowski2020convex}. In greedy algorithm-based methods, the sensor placement is optimized sequentially. Basically, we select the optimal sensor placement one-by-one conditioned on previous sensor placements. By doing so, it allows us to convert a high-dimensional optimization problem into multiple low-dimensional optimization problems that can be solved sequentially. For instance, in order to place 10 sensors on a 3-dimensional structure called miter gate (see Fig. \ref{fig:posterior}), a 30 dimensional optimization problem needs to be solved, if an evolutionary optimization method is employed directly. Instead of directly solving the 30 dimensional optimization problem, \cite{yang2021probabilistic} optimized the sensor placement one-by-one conditioned on previous sensor placements. In each iteration, only a three-dimensional optimization problem is solved using Bayesian optimization method \citep{yang2021probabilistic}.
\item  {\bf{Reinforcement learning-based methods}}: Even though the greedy algorithm-based methods make sensor placement optimization under uncertainty computationally more tractable, it may lead to sub-optimal solutions due to the nature of greedy algorithms. Reinforcement learning-based methods have recently been proposed to alleviate the limitation of greedy-based methods. In reinforcement learning-based methods, sensor placement optimization is formulated as a sequential decision-making problem, such as a Markov decision process model. This problem is solved using one of many reinforcement learning algorithms, such as dynamic programming, Q-learning, policy gradient reinforcement learning, and deep Q-learning \citep{alsheikh2015markov,wang2006adaptive,wang2019reinforcement,kaveh2022optimal,shen2021bayesian}. A unique properly of this problem is it considers the impact of a candidate sensor placement solution on both current decision making and the placements of other sensors and decision making at future time instances (i.e., look-ahead). For example, \cite{wang2019reinforcement} and \cite{shen2021bayesian} have developed reinforcement learning-based sensor placement optimization methods for spatiotemporal modeling and Bayesian model updating, respectively. Results of their papers show that reinforcement learning-based methods tend to be more effective in finding optimal solutions than greedy algorithm-based methods and generic algorithm-based methods \citep{wang2019reinforcement,shen2021bayesian}. A dedicated discussion on other applications of deep reinforcement learning in digital twins can be found in Sec. 6.3 of our Part 1 paper.

\end{itemize}

Optimizing a sensor network for a physical system allows the most informative data to be collected from the physical system. The collection of informative datasets will significantly improve the performance of the P2V connection for model updating as well as the efficacy of the overall digital twin in support of real-time decision making and control. Figure \ref{fig:posterior} shows an example of sensor placement optimization as part of a miter gate digital twin project sponsored by the U.S. Army Corps of Engineers \citep{vega2021novel}. A finite element structural analysis model was first developed to predict the structural response under different conditions, as shown in Fig. \ref{fig:posterior} (a). Based on the structural analysis model, sensor placement was optimized to collect data from the miter gate to estimate the level of structural damage located at the lower-left corner of the gate. The damage level (i.e., gap length) needed to be estimated by updating the finite element model given in Fig. \ref{fig:posterior} (a) using Bayesian filters described in Sec. 4.2.2 of our Part 1 paper. Figure \ref{fig:posterior} (b) compares the estimated posterior damage level obtained from the optimal sensor design and a non-optimal sensor design at a certain time instant. This figure shows that the posterior damage estimate obtained from the optimal sensor design is much more concentrated than that from the non-optimal design. Clearly, sensor network optimization led to a significant reduction in the uncertainty of the structural damage estimate compared to a non-optimal design. This example highlights the added value of sensor network design optimization in digital twins.

\subsubsection{Optimization for physical system modeling (offline)}
\label{sec:opt_physical_offline}

As mentioned in Sec. 4.2.1 of our Part 1 paper, digital state space is usually designed to be simple enough to make online model updating feasible and tractable. A digital state consists of 
\begin{itemize}
    \item a set of state variables $\textbf{x}$ which are the smallest set of variables that determine the state of the physical system, and
    \item a group of model parameters ${\boldsymbol{\tilde \uptheta }} = [ {\boldsymbol{\uplambda }},\; {\boldsymbol{\uptheta }}] $, in which ${\boldsymbol{\uptheta }}$ is a very small subset of the  ${\boldsymbol{\tilde \uptheta }}$ that will be updated online along with state variables $\textbf{x}$ using methods described in Sec. 4.2.4 of Part 1, and ${\boldsymbol{\uplambda }}$ represents the remaining model parameters that will not be updated online.
\end{itemize}

In order to bridge the gap between the initial digital model and the physical counterpart, the uncertain model parameters ${\boldsymbol{\tilde \uptheta }}$ need to be calibrated offline using experimental data before deploying the digital twin. Numerous approaches have been proposed to perform such an offline calibration of digital models, including Bayesian calibration-based methods and optimization-based methods. 

In Bayesian calibration-based methods, the optimal model parameters ${\boldsymbol{\tilde \uptheta }^*}$ can be estimated  by solving the following optimization problem
\begin{equation}
    \label{eq:Cal_opt}
    {{\boldsymbol{\tilde \uptheta }}^*} = \mathop {\arg \max }\limits_{{\boldsymbol{\tilde \uptheta }} \in \Omega } \{ f({\boldsymbol{\tilde \uptheta }}\vert{{\bf{y}}_e})\} ,
\end{equation}
where ${{\bf{y}}_e}\in \mathbb{R}^{n_e}$ is a vector of experimental observations, $f({\boldsymbol{\tilde \uptheta }}\vert{{\bf{y}}_e})$ is the posterior distribution of ${\boldsymbol{\tilde \uptheta }}$ for given experimental observation ${{\bf{y}}_e}$. Note that the true values of ${\boldsymbol{\tilde \uptheta }}$ are fixed but unknown to us, since the uncertain model parameters ${\boldsymbol{\tilde \uptheta }}$ are considered to be epistemic uncertainty only (see Sec. \ref{sec:UQ_dynamic_system_models}). 

\begin{figure}[!h]
  \centering
    {\includegraphics[scale=0.65]{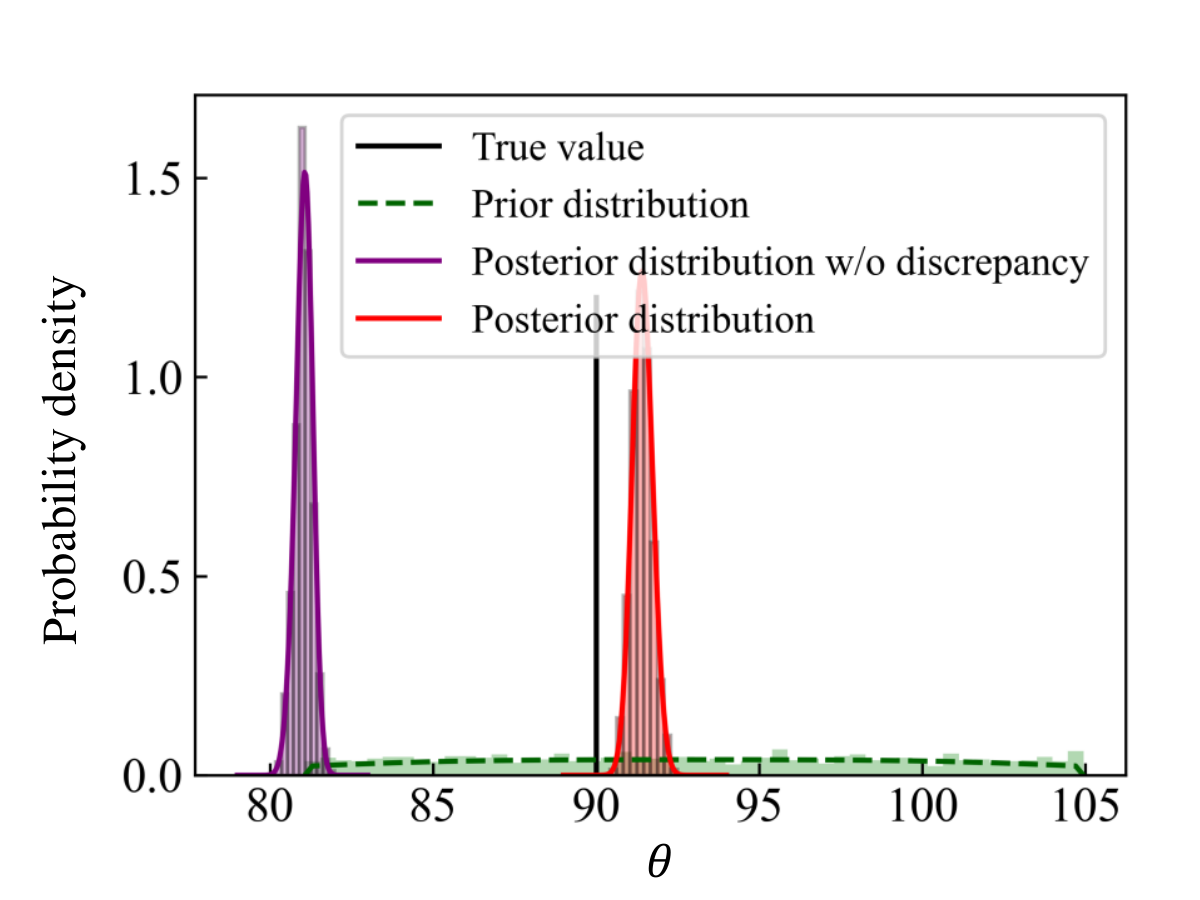}}
  \caption{Comparison of posterior distributions obtained with and without accounting for model discrepancy term \citep{jiang2022bayesian}.}
  \label{fig:post_comparison}
\end{figure}

$f({\boldsymbol{\tilde \uptheta }}\vert{{\bf{y}}_e})$ can be obtained using various Bayesian updating methods, such as the classical Markov chain Monte Carlo simulation, Gibbs sampling, and slice sampling. When model uncertainty of a digital model is considered, the uncertain model parameters ${\boldsymbol{\tilde \uptheta }}$ can be calibrated concurrently with a model discrepancy term using the KOH framework \citep{kennedy2001bayesian} (see Eq. (\ref{eq:KOH_equation}) in Sec. \ref{sec:UQ_dynamic_system_models}). It has been shown that accounting for model discrepancy during Bayesian calibration can not only improve the prediction accuracy of the digital model, but it can also lead to an improved accuracy in estimating the posterior distribution of the unknown model parameters ${\boldsymbol{\tilde \uptheta }}$. As indicated in Fig. \ref{fig:post_comparison}, the posterior distribution of a unknown model parameter $\theta$ considering model discrepancy is closer to the true value than its counterpart without accounting for model discrepancy. Additionally, it is worth mentioning that the Bayesian filters described in Sec. 4.2.2 of Part 1 can also be used to obtain $f({\boldsymbol{\tilde \uptheta }}\vert{{\bf{y}}_e})$ since a special case of online model updating is the offline calibration where the Bayesian filters are used to update only model parameters instead of both state variables and model parameters.

The estimation given in Eq. (\ref{eq:Cal_opt}) is also called maximum a posteriori estimate. In practice, a point estimate is used instead of the joint posterior distribution is to keep the number of uncertain model parameters as low as possible in the digital model, and thus make online model updating of digital twins feasible and tractable. While Bayesian calibration methods under the KOH framework have been shown to be accurate and robust in estimating uncertain model parameters, the implementation of those methods is relatively complicated and the required computational effort may be quite high, especially when ${\boldsymbol{\tilde \uptheta }}$ is high-dimensional. Therefore, optimization-based methods are often employed as an alternative that largely alleviates the computational burden in practice.

Optimization-based calibration methods estimate ${\boldsymbol{\tilde \uptheta }}$ by maximizing or minimizing a calibration metric as follows
\begin{equation}
\label{eq:Cal_opt2}
{{\boldsymbol{\tilde \uptheta }}^*} = \mathop {\arg \min }\limits_{{\boldsymbol{\tilde \uptheta }} \in \Omega } \;{\rm{or}}\;\mathop {\arg \max }\limits_{{\boldsymbol{\tilde \uptheta }} \in \Omega } \{ C({{\bf{y}}_e},\;g({\boldsymbol{\tilde \uptheta }},\;{\boldsymbol{\upalpha }}))\},
\end{equation}
where $C({{\bf{y}}_e},\;g({\boldsymbol{\tilde \uptheta }},\;{\boldsymbol{\upalpha }}))$ is a calibration metric, $g({\boldsymbol{\tilde \uptheta }},\;{\boldsymbol{\upalpha }})$ is the prediction of the digital model for given ${\boldsymbol{\tilde \uptheta }}$ and ${\boldsymbol{\upalpha }}$, and ${\boldsymbol{\upalpha }}$ stands for the aleatory uncertain variables in the digital model (the same as that in Sec. \ref{sec:opt_sensor_placement}).

Various calibration metrics have been proposed in the past decades. For instance, the least squares method as described in Eq. (7) in Sec. 4.2.4 of our Part 1 paper is an optimization-based method with the mean squared error as the calibration metric. The least squares method is the easiest to implement, and is probably the most widely used one in industry. But it is sensitive to outliers in experimental data. If a likelihood function is used as the calibration metric, it is called the maximum likelihood estimation method \citep{xiong2009better}. This method has shown similar performance as Bayesian methods under the KOH framework \citep{xiong2009better}. But it may not perform well if the amount of experimental data for calibration is small. Some other examples of calibration metrics include the moment matching metric which compares the difference between the statistical moments obtained by experiments and prediction \citep{bao2015monte}, similarity metric that measures the similarity between prediction and experiments \citep{cha2007comprehensive}, and the marginal probability and correlation residual metric considering both marginal probability and correlation coefficient residuals \citep{kim2020new}. 

As mentioned above, all methods have their own advantages and disadvantages. Among them, Bayesian methods, the maximum likelihood estimation method, and the least squares method are the three most widely used ones. The selection of an appropriate method is mainly dependent on the amount of available data and decision maker's acceptable level of complexity. Furthermore, the following topics also play a vital role in bringing the initial digital model closer to the physical system.
\begin{enumerate}
    \item \emph{Model validation}: Model validation is an essential step to validate the digital model after calibration of the model offline. It is \lq\lq the process of determining the degree to which a model or a simulation is an accurate representation of the real world from the perspective of the intended uses of the model or the simulation" \citep{NASA2008standard,mahadevan2022uncertainty}. In order to quantitatively quantify the agreement between the digital model prediction and experimental observations, various statistical metrics have been proposed, including Bayesian hypothesis testing \citep{jiang2009bayesian}, reliability-based metric \citep{rebba2008computational,ao2017dynamics}, area metric \citep{li2014new}, etc. An essential characteristic of various validation metrics is that they account for various uncertainty sources in the digital model used for calibration and in the experiments used to evaluate model validity. \cite{liu2011toward} and \cite{ling2013quantitative} analyzed the  pros and cons of different metrics through comparative studies. For instance, \cite{liu2011toward} pointed out that small perturbations in the pre-specified confidence level could significantly affect the rejection or non-rejection of a digital model using classical hypothesis testing.  \cite{ling2013quantitative} concluded that both a Bayes factor and reliability-based metric could be mathematically related to the $p$-value metric in classical hypothesis testing. An appropriate validation metric should be selected to validate the calibrated digital model according to the application by analyzing the pros and cons of different metrics. We direct interested readers to \cite{liu2011toward} and \cite{ling2013quantitative} for more detailed discussions on this important topic.
    \item \emph{Experimental design optimization}: Experimental design optimization is a process of optimizing experimental input settings to collect the most informative experimental data for model calibration and validation \citep{ao2017design,hu2017calibration,huan2013simulation,huan2014gradient}. Even though formulated in a different context, experimental design optimization is fundamentally the same as sensor placement optimization discussed in Sec. \ref{sec:opt_sensor_placement} and can be considered as a sub-topic of sensor placement optimization. It can help reduce the required number of experiments for the calibration of a digital model offline.
\end{enumerate}

After the offline calibration of the digital model, ${\boldsymbol{\tilde \uptheta^*}}$ obtained from Eq. (\ref{eq:Cal_opt}) or (\ref{eq:Cal_opt2}) will be used as the initial values of ${\boldsymbol{\tilde \uptheta }}$. A small subset of ${\boldsymbol{\tilde \uptheta }}$ denoted as ${\boldsymbol{\uptheta }}$ will be updated along with state variables $\textbf{x}$ using the methods discussed in  Sec. 4.2.4 of our Part 1 paper to account for the fact that some parameters (e.g., battery capacity) change very slowly over the life-cycle of a physical system.




\subsubsection{Optimization for predictive decision making (online)}
\label{sec:opt_decision_making}

\paragraph{(a) Real-time requirements of digital twins}

When discussing real-time requirements of digital twins, it is important to understand that the definition of ``real-time" varies depending on the application. In general, the definition of real-time is \emph{the minimum computational speed required to achieve seamless and uninterrupted optimization, prediction, and control of the system of interest}. Ultimately, the timescale of the system of interest is what defines the requirements for real-time computing. Take for example a digital twin built to model the degradation of a lithium-ion battery cell. A Li-ion cell is designed to last many thousands of cycles, which in standard applications, is on the time scale of years. In this case, real-time optimization and control related to a Li-ion cell's degradation needs to be computed on the time scale of days or weeks in order to enable timely control of its usage. On the other hand, high-rate systems like ultrasonic vehicles, hypersonic weapons, blast mitigation systems, and vehicle crashes operate on much shorter time scales, often $100 ms$ or shorter \citep{dodson2022high}. When modeling these systems with a digital twin, it is much more difficult to ensure that sensing, prediction, and control can take place on the desired time scales. Research in this area is actively investigating modeling techniques which can meet the demanding updating and predicting requirements. Sometimes, it is often intractable to set the requirement that the state estimation model operate on timescales shorter than the timescale of the system of interest. This is especially the case for very high-rate ($<$ 100 $\mu$s) and ultra high-rate systems ($<$ 1 $\mu$s). When this is the case, researchers will define an acceptable time delay, that if the state estimation model can achieve, would be useful to a larger predictive control framework. Examples of high-rate system modeling research include work by \cite{yan2021online}. In their paper, they investigated using a simplified physics-based model to track, update, and predict the state of highly dynamic systems. Their experiments on two different test setups showed the model was able to update and predict with an average computation time of 93 $\mu$s. Other work by \cite{barzegar2022ensemble} investigated using a deep-learning-based model architecture for high-rate system state prediction. Their proposed recurrent neural network, used for state estimation in high-rate structural health monitoring (HRSHM) applications, achieved accurate predictions with an average computational time of 25 $\mu$s.

\paragraph{(b) Real-time optimization of additive manufacturing processes}

Offline optimization as described in above sections is commonly applied to high-level functions in smart manufacturing that do not require real-time optimization. Examples of these high-level functions include product design, production planning, and maintenance scheduling (although online optimization, not in real-time, is required to schedule maintenance in some cases). These functions, as defined in ISA-95, have typical cycles from hours to months \citep{ISA95Part1}. For those activities, open-loop optimizations are applied and no feedback-based adjustments involved in executing the optimal decisions. However, process controls, especially for complex processes with high uncertainties and significant disturbances, require continuous adaption of control strategy and real-time optimization. 

Additive manufacturing is one of such complex layer-by-layer fabrication processes. For example, the metal powder bed fusion (LPBF) process involves spreading a thin layer of metal powder followed by exposure to high-intensity laser energy directed in scanned trajectories defined by digital models. The build process involves multiple physical phenomena: heat absorption, melt pool formation, solidification, and even re-melting and re-solidification \citep{PBF2014}. A great number of factors affect the quality of additively manufactured parts, including processing parameters such as laser power and scan velocity, environmental parameters such as chamber temperature and humidity, as well as the non-deterministic material powder characteristics. The complex and stochastic nature of the additive manufacturing process requires real-time  optimization for stable process and controllable part quality. 

\begin{figure*}[!ht]
  \centering
  \includegraphics[scale=0.35]{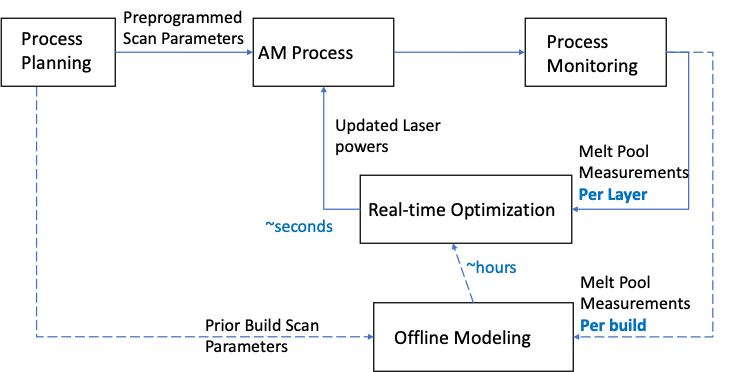}
  \caption{Real-time optimization for powder bed fusion process control}
  \label{fig:AM_Optimization}
\end{figure*}

Figure \ref{fig:AM_Optimization} shows a layerwise real-time optimization strategy for laser powder bed fusion process control. Process control commands and melt pool monitoring data collected from previous builds are used as training data for a melt pool size prediction model, which can be represented as:

\begin{equation}\label{eq:Melt_Pool Prediction Model}
{y_i} = f({t_i},{P_i},{v_i},\theta _i^{\Delta t},\theta _i^{\Delta d},J,\lambda ,{A_{avg}},{A_{max}},{A_{avr}})
\end{equation}
where $y_i$ represents the melt pool size at step $i$, $t_i$ represents the scan time, $P_i$ is the current laser power, $v_i$ is the current scan speed, $\theta_i^{\Delta t}$ is the temporal-accumulated prior scan effects, $\theta_i^{\Delta d}$ is the spatial-accumulated prior scan effect, $J$ is the total energy input on the previous layer, $\lambda$ represents the laser idle time from the end of the previous layer to start of current layer, and $A_{avg}$, $A_{max}$, and $A_{var}$ represent the statistical features of the melt pool size within the previous layer neighborhood of the current scanning position \citep{PBF_Modeling2020}. A neural network was trained to predict melt pool size accurately based on process parameters and earlier melt pool measurements at the same layer and from the previous layer.

The machine learning model trained from prior builds can be used for real-time layerwise scan parameter optimization \citep{PBFOptimization2020}. In Fig. \ref{fig:AM_Optimization}, the objective of the optimization is to regulate the melt pool size. To achieve this goal, the potential control variables for optimization can be laser power, scan speed, and laser scan path. However, modifying scan speed or laser scan path requires significant computing efforts. Therefore only laser power is selected as the single control variable for the optimization, which focuses on managing the melt pool size into a desired range.

\paragraph{(c) Real-time mission planning}
The continual data flow from a physical system allows its digital twin to closely map, monitor, and control real-world entities. Combined with P2V twinning enabling technologies, such as the Bayesian model updating methods described in Sec. 4.2 of Part 1, digital twins provide an adaptive mechanism to support data-driven decision automation in accordance with asset-specific conditions.

One representative application that demonstrates a digital twin's capability in adaptive decision support is mission planning. For example,~\cite{kapteyn2021probabilistic} considered the structural health of an UAV in a mission planning task, where the right wing of the UAV was assumed to have two defective regions on the upper surface. Based on a unifying probabilistic graphical model built upon Bayesian inference,~\cite{kapteyn2021probabilistic} showcased the application of a structural digital twin in the operations management of a UAV (see Fig.~13 of our Part 1 paper for a graphical illustration). The structural digital twin, after it was calibrated to a specific as-manufactured asset, assimilated incoming sensor data on the UAV to perform in-flight SHM and carry out health-aware mission planning, where optimal maneuvers (i.e., turn and bank angle) were derived using the structural digital twin of the UAV in consideration of its evolving structural health. ~\cite{sisson2022digital} pursued a similar idea to minimize the stress experienced by critical mechanical components of rotorcraft via optimizing the horizontal and vertical flight velocities for mission planning. Towards this goal, a digital twin approach was developed to diagnose component health and estimate the damage growth under different velocity conditions.

\begin{figure}[!h]
  \centering
    {\includegraphics[scale=0.65]{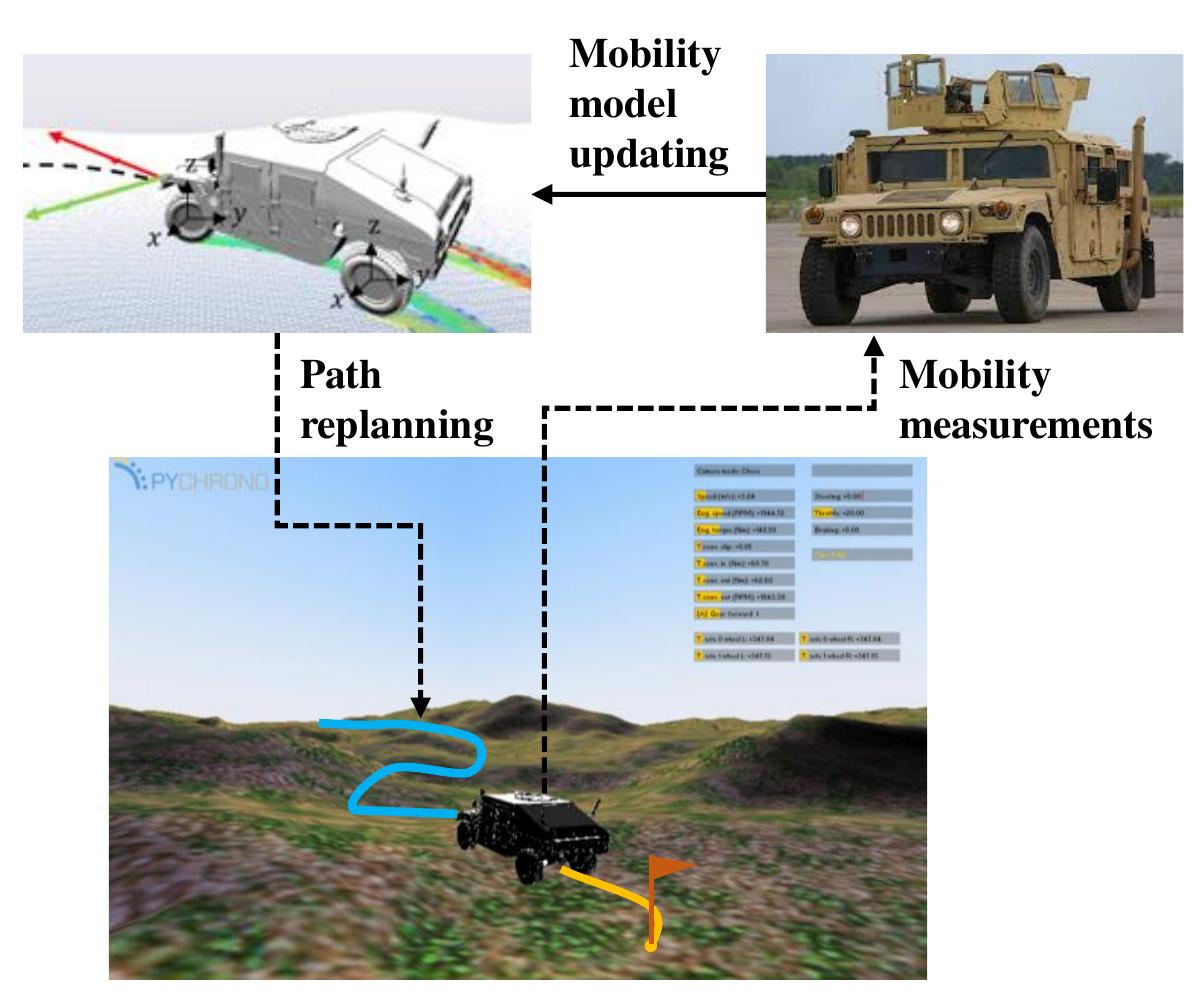}}
  \caption{Real-time mission planning of off-road autonomous ground vehicle (AGV) \citep{liu2021reliability,liu2021simulation}.}
  \label{fig:AGV_path}
\end{figure}

Another example is the real-time mission planning of off-road autonomous ground vehicles as illustrated in Fig. \ref{fig:AGV_path}. These vehicles operate in highly stochastic, harsh, and uncontrolled off-road environments. They have been used to replace humans in the battlefield for military applications \citep{jiang2021r2}, in the agricultural field to reduce labor requirements \citep{mousazadeh2013technical}, and in space for exploration of the Moon and Mars \citep{johnson2015discrete}. A major challenge in realizing off-road autonomy is the high uncertainty in vehicle mobility (e.g., energy consumption and maximum attainable speed) caused by complex terrain conditions, such as mud, sand, and grass \citep{jiang2022efficient}. Digital twin technology provides a promising solution to this challenging issue and plays a vital role in enabling real-time mission planning of off-road autonomous ground vehicles. 

As illustrated in Fig. \ref{fig:AGV_path}, a high-fidelity digital model of an off-road autonomous ground vehicle has been developed as part of the modeling and simulation effort of the U.S. Army Ground Vehicle Systems Center \citep{arc2022}. The digital model predicts vehicle mobility under different terrain conditions and allows for the generation of a probabilistic mobility map \citep{jiang2021r2}. The mobility map provides critical information about where the vehicle can go (i.e., GO) and where it cannot (i.e., NOGO). It is essential for the mission planning of the autonomous vehicle since it enables the vehicle to predict where the obstacles are. As the physical system of a vehicle travels along a path, as indicated in the orange color in Fig. \ref{fig:AGV_path}, vehicle mobility data and terrain information are collected through torque and speed sensors, LiDAR, and cameras. The collected mobility data can be used to update the digital mobility model using a Bayesian model updating method discussed in Sec. 4.2 of Part 1. Subsequently, the probabilistic mobility map can be updated. The real-time mission planning can then be performed based on the updated mobility map to ensure the success of a mission under uncertainty (as illustrated in Fig. \ref{fig:AGV_path}). 

Real-time path planning of off-road autonomous ground vehicles is a Nondeterministic Polynomial (NP)-hard optimization problem. Approaches have been developed for mission planning using sampling-based methods and search-based methods. Sampling-based methods randomly draw samples within a map and construct a space-filled tree to identify the optimal path. Some examples of sampling-based methods include Rapid-exploring Random Tree (RRT) \citep{kuffner2000rrt}, RRT* \citep{gammell2014informed}, R2-RRT* \citep{jiang2022efficient}, etc. Search-based methods identify the shortest path based on a pre-defined network or graph using searching algorithms, such as Dijkstra’s algorithm \citep{barbehenn1998note}, and its optimized version, A* algorithm \citep{duchovn2014path}. 

\paragraph{(d) Predictive maintenance scheduling}
\label{sec:predictive_maintenance_scheduling}
Another area that may significantly benefit from digital twin technology is maintenance scheduling. In a traditional setting like a manufacturing plant, maintenance is typically performed in a reactive approach (after the machine has failed) or in a proactive approach (fixed interval repairs regardless of machine status). However, the reactive approach can be dangerous, as it allows machines to fully fail, which can harm nearby personnel. Similarly, the proactive approach of constantly servicing the machine on a fixed interval is wasteful because the machine may be healthy and does not require repair for some time. A real solution to this problem is individualized digital twin models which can predict a machine's future degradation trajectory and online optimize the maintenance schedule. 

To effectively online optimize the maintenance of a unit, two key models are needed: (1) a prognostic model which can estimate the remaining useful lifetime of the unit, and (2) an optimization model which considers the RUL estimate from the prognostic model as well as the maintenance engineer's preferences before determining the optimal time to conduct maintenance. If the prognostic model is accurate, maintenance can be conducted "just-in-time," effectively maximizing the function of the unit while minimizing downtime and repair costs typically incurred using reactive or preventative maintenance approaches \citep{lee2013recent,lee2013predictive}.

The first key component in a digital twin for predictive maintenance scheduling is the prognostic model. Much research has already been done regarding prognostic methods for estimating RUL, and the major methods are discussed in detail in Sec. 5.2 of Part 1. The second key component is the optimization scheduling algorithm. Research in this area is focused on developing optimization methods to determine the best time to perform maintenance. In some cases, the main driving factor for optimizing the maintenance schedule is the long-term cost per unit of time \citep{grall2002continuous,bousdekis2019decision}. To solve these maintenance optimization problems, a cost function is formulated which captures the costs associated with maintenance, and is minimized using an applicable solver \citep{camci2009system,suresh2013hybrid}. Additionally, the cost functions can take into consideration the predicted remaining useful life of the unit at the time of maintenance. This is important because any remaining lifetime the unit had will discarded as soon as maintenance happens \citep{lei2018maintenance}.

In other scenarios, the cost of maintenance is not the main consideration, and other objectives need to be prioritized. Take for example a major highway. The highway serves many people, and needs to be maintained on a regular basis. The optimal time to close a highway for construction would be when the weather is best and when traffic is the least \citep{sabatino2015sustainability, allah2019network}. These objectives are harder to quantify and optimize together, so researchers have proposed creating utility functions, which map different objectives to the same range, so they can be more easily combined and optimized \citep{huber1974multi, winterfeldt1975multi}. This method, referred to as multi-attribute utility theory, is a popular method for optimizing non-traditional objectives \citep{froger2016maintenance,camci2015maintenance}.

Digital twin is a promising technology that has the potential to online optimize the maintenance times of units operating in the field. The automatic data flow between the prognostic and optimization models in the digital twin enables optimization solutions which are specific to a single unit in the field, and take into consideration its current and future health status. These ideas are further investigated in the case study, where we show how one might build a battery digital twin to optimize the retirement of a battery cell from its first life use. Ultimately, the strongest aspect of a digital twin is its ability to optimize over dimensions of interest. In the future, we see predictive maintenance scheduling on a unit-by-unit basis as a core application of digital twins.

\subsubsection{Summary of commonly used optimization methods in digital twins}
Table \ref{tab:optimization_summary} summarizes applications of different optimization methods in different dimensions of digital twins as discussed above. The advantages and disadvantages of different methods are also briefly summarized in this table. As illustrated in Table \ref{tab:optimization_summary}, evolutionary optimization methods, such as genetic algorithms, simulated annealing, etc., in general, can be applied to the dimensions of modeling (i.e., optimization of physical system modeling) and V2P connection (e.g., process control, mission planning, and maintenance scheduling) of a digital twin. The required number of function evaluations by the evolutionary optimization methods is usually very high. Therefore, in digital twin applications, evolutionary optimization methods are often used (1) when the objective function is very cheap to evaluate or (2) in conjunction with machine learning-based surrogate models to alleviate the required computational effort. In contrast, greedy algorithms are computationally much cheaper while it suffers from the downside of converging to local optima (or locally optimal solutions).

In recent years, reinforcement learning-based optimization methods have emerged as promising global optimization tools in many digital twin applications. This recent development was largely due to the significant increases in computational power and the many technological advances in deep learning (see Sec. 6.3 of our Part 1 paper). Nevertheless, reinforcement learning-based methods require high volumes of data for training and are complex to implement in practice. They are usually employed when an optimization problem is too complicated to be solved by conventional optimization methods.

Bayesian optimization based on Gaussian process regression is another widely used optimization method in digital twins \citep{snoek2012practical}, even though it is not elaborated in detail in the discussions above. But Bayesian optimization may suffer from the curse of dimensionality because of the scalability limitation of Gaussian process regression. To enable real-time mission planning in V2P connection, a group of mission/path planning algorithms (e.g., A*, RRT*) have been developed in the past decades. Even though they may lead to sub-optimal solutions, they were shown to be computationally efficient and highly effective in finding near-optimal paths in real-time for practical applications.

\begin{table*}[!ht]
\centering
\small
\caption{Summary of optimization methods in digital twins}
\begin{tabular}{p{0.7cm}p{2.2cm}|p{2.2cm}|p{1.8cm}|p{2.5cm}|p{1.9cm}|p{1.9cm}}
\hline\hline
\multicolumn{2}{l|}{\textbf{Method}} & \textbf{Evolutionary optimization methods} & \textbf{Greedy algorithms} & \textbf{Reinforcement learning-based methods} & \textbf{Bayesian optimization} & \textbf{A*, RRT, RRT*, Dijkstra algorithm}  \\ \hline
\multicolumn{2}{l|}{\textbf{Online/offline}} & Online/Offline & Offline    & Online/Offline & Online/Offline       & Online  \\ \hline
\multicolumn{1}{l|}{\multirow{5}{*}{\rotatebox[origin=c]{90}{\textbf{Application} \hspace{1.1cm} }}} & Sensor placement optimization   & \checkmark  & \checkmark     & \checkmark        & \checkmark      & \xmark \\ \cline{2-7} 
\multicolumn{1}{l|}{}                             & Physical system modeling & \checkmark                                            & \xmark    & \xmark       & \checkmark      & \xmark \\ \cline{2-7} 
\multicolumn{1}{l|}{}                             & Process control  & \checkmark                                            & \checkmark    & \checkmark       & \checkmark      & \xmark \\ \cline{2-7} 
\multicolumn{1}{l|}{}                             & Mission planning  & \checkmark                                            & \xmark    & \checkmark       & \xmark      & \checkmark \\
\cline{2-7} 
\multicolumn{1}{l|}{}                             & Maintenance  & \checkmark                                            & \checkmark    & \checkmark       & \xmark      & \xmark \\ \hline
\multicolumn{2}{l|}{\textbf{Pros}}                                    & High-dimensional optimization; Global optimization; & Relatively low number of function evaluations    & Global optimization with long-term rewards       & Global optimization      & Real-time global optimization \\ \hline
\multicolumn{2}{l|}{\textbf{Cons}}                                    & Large number of function evaluations                                            & Local optimal    & Data-intensive and high complexity       & Curse of dimensionality      & Suboptimal solution \\ 
\hline \hline
\end{tabular}
\label{tab:optimization_summary}
\end{table*}









\section{Case study: a battery digital twin}
\label{case_study}
Next, we use a battery digital twin case study to illustrate the implementation of a digital twin. Code and preprocessed data for generating all the results and figures presented in the case study are available on \cite{github_code}.

\subsection{Background}
\label{sec:case_study_introduction}
Lithium-ion (Li-ion) batteries power phones, laptops, and more recently electric vehicles. They can supply a great deal of power for many hundreds of cycles, but eventually, they degrade and lose capacity due to irreversible internal electrochemical changes \citep{lu2013review, liu2021effects}. It is therefore of great importance to accurately model and forecast future capacity degradation to estimate the lifetime cells operating in the field \citep{hu2020battery, zhang2011review}. Accurate RUL predictions of cells operating in the field can empower manufacturers and operators to make informed decisions regarding the best time for cell maintenance or replacement/retirement. Recently, there has been much discussion around second life applications for retired cells as the global supply of used batteries is rapidly increasing \citep{engel2019second, casals2019second}. However, not all used batteries are the same because their usage during their first life application shapes their present health status and remaining capacity \citep{birkl2017degradation}. For example, an automaker cannot simply retire cells from vehicles using a fleet-wide kilometer threshold because each vehicle would have been subjected to a unique driving pattern that may have been more or less degrading to the cells’ capacity. To make the most out of repurposed Li-ion cells, the time at which the cells are retired from their first life application needs to be optimized and determined online, in real-time, and on a per-cell or per-pack basis. One of the most promising solutions to the challenge of unit-specific real-time modeling is digital twin. 

Unfortunately, many of the digital twin models reported in the literature are not truly a digital twin model, as they only operate in three or four of the five total dimensions outlined in Fig. 3 in Part 1 of this review \citep{SMO1}. Many of the research papers we reviewed which claim to have created a digital twin have actually proposed a prognostic model for predicting the RUL of an engineered system. However, a prognostic model on its own is not fully a digital twin. In a prognostic model, sensor data from the physical system (PS) is used to update the digital state (P2V) of the digital system (DS). Then, the digital system is used to make a prediction about the future state of the physical system (V2P), i.e. the virtual prognostic model is used to predict the RUL of the physical system. The data flow then ends after RUL prediction of the physical system, thus not achieving all five of the data flow dimensions which characterize digital twin. This type of work neglects the most important dimension: optimization (OPT). To complete the data flow through all five dimensions, the RUL prediction from the prognostic model needs to be used as input into an optimization and control algorithm to manage some other system attributes.

In this case study, we develop and discuss a proof-of-concept battery digital twin which can be used to determine the optimal time to retire a Li-ion cell from its first life application in advance. The proposed five-dimensional digital twin model consists of two key pieces of software: 1) a particle filter battery prognostic model which can estimate the future capacity fade trajectory and predict cell RUL (PS, P2V, DS, V2P), and 2) a multi-attribute utility-based optimization model which takes into account user preference and the projected future capacity trajectory when determining the optimal time to retire a cell from its first life (OPT). The first piece of software, the particle filter battery prognostic model, is used to capture four of the five dimensions required for a digital twin: PS, P2V, DS, V2P. The last dimension, optimization (OPT) is captured by the second piece of software, the multi-attribute utility-based optimization model. The optimization algorithm interfaces with the physical system by triggering actions from operators, completing the data flow loop through all five dimensions of digital twin. The proposed digital twin model can operate online and is recursively updated each time a new battery capacity measurement becomes available. As a first attempt at a maintenance oriented digital twin for Li-ion batteries, it is our hope that the ideas discussed, and the code provided with this paper will help spark interest and innovation in this new area of research.

\subsection{Methods}
\label{sec:case_study_methods}
\begin{figure}[!h]
  \centering
    {\includegraphics[scale=0.55]{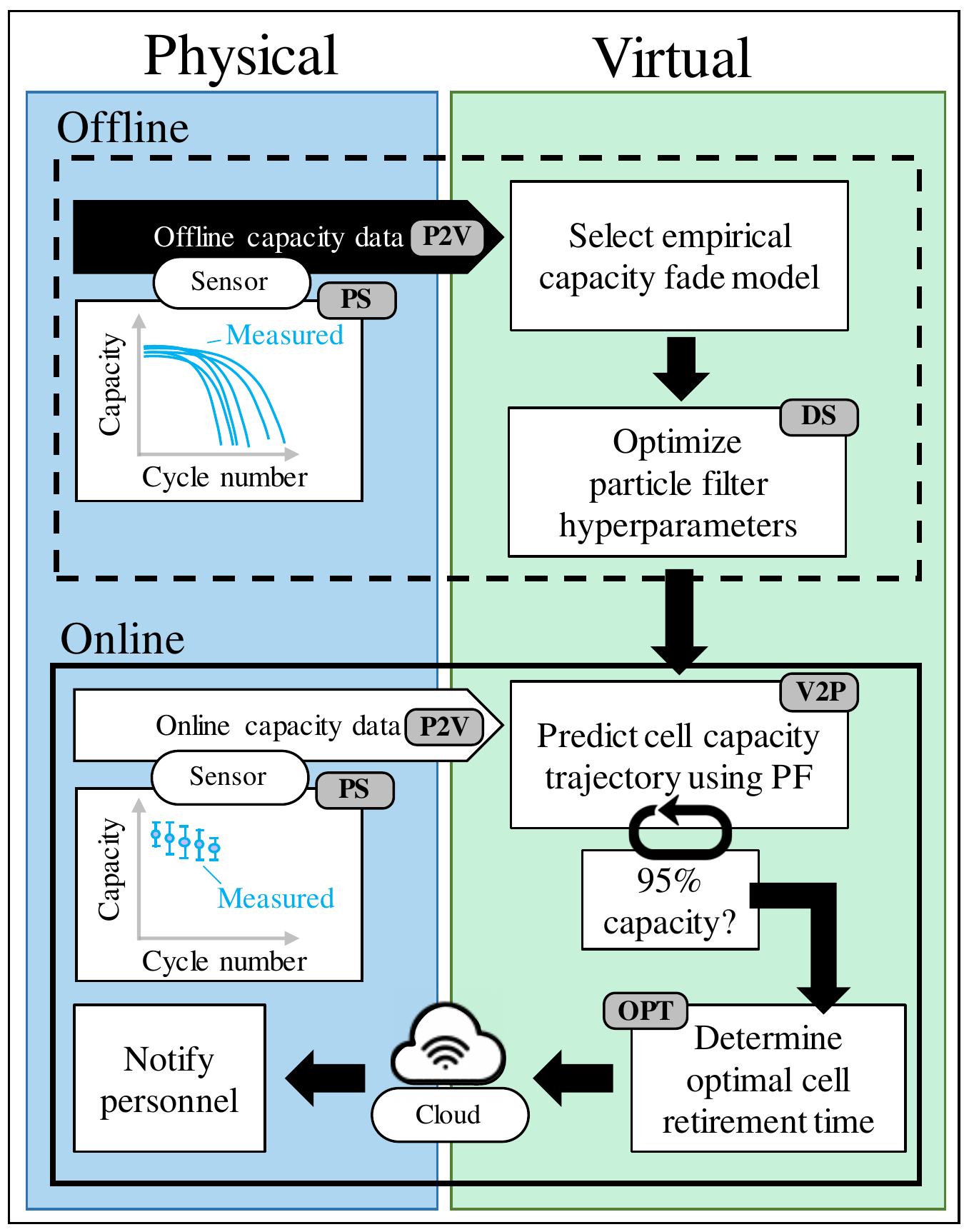}}
  \caption{Overview of the proposed battery digital twin framework.}
  \label{fig:battery digital twin overview}
\end{figure}

\begin{figure*}[!ht]
  \centering
    {\includegraphics[scale=0.7]{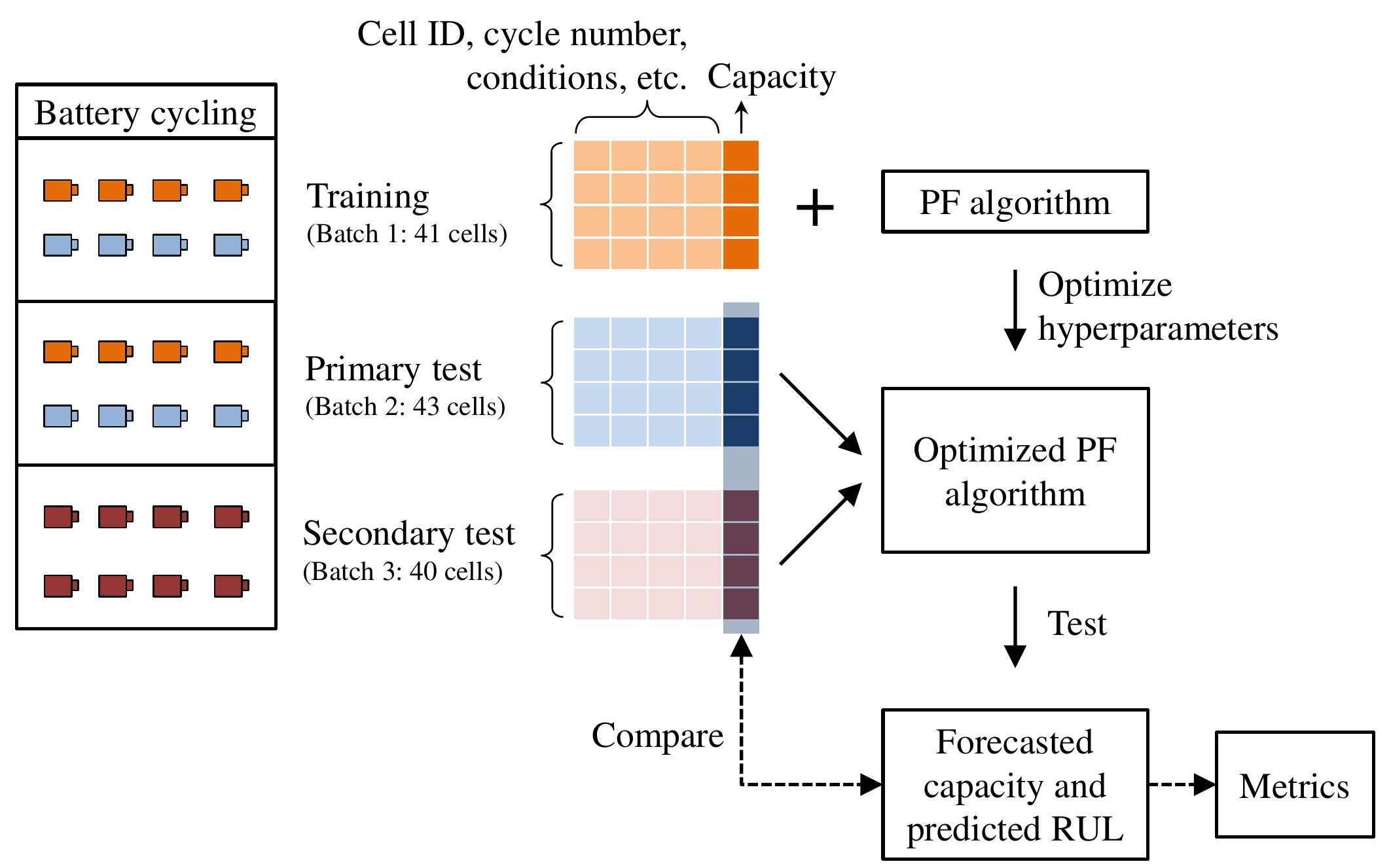}}
  \caption{Overview of the method used to offline optimize the particle filter (``PF" in the figure) and test it online.}
  \label{fig:offline_online PF pipeline}
\end{figure*}

An overview of the proposed battery digital twin framework for optimizing the retirement of Li-ion cells from their first life application is shown in Fig. \ref{fig:battery digital twin overview}. The digital twin model, covers all five of the key data transfer and modeling dimensions outlined in Fig. 3 of Part 1 of this review \citep{SMO1}. In the offline phase, previously collected run-to-failure battery data is used to optimize the initial parameters of the particle filter. Then during the online phase, the particle filter is used to predict the future capacity fade trajectory of a cell. When the online cell's capacity measurements reach $95\%$ their initial value, the first life retirement optimization code is triggered, and the predicted capacity trajectory from the particle filter is used to determine the optimal cycle to remove the cell from its first life application. This concept of first optimizing the particle filter offline and then using it on cells online is shown visually in Fig. \ref{fig:offline_online PF pipeline}. In the sections that follow, we provide more details about the individual algorithms used in the proposed digital twin framework. 

\subsubsection{Particle filter prognostic model}
Li-ion batteries are a complex electrochemical system which degrade and lose capacity based on how they are used. Use conditions like the charging/discharging rate, depth of discharge, time, and ambient temperature all significantly affect Li-ion cell degradation and capacity loss \citep{birkl2017degradation}. As a result of the many aging factors, the shape of a cell’s remaining capacity vs time plot can take on many forms (i.e. linear, exponential, power law, double exponential, all with respect to time). This can make modeling and forecasting of battery RUL difficult, especially when more than one capacity fade trend is present.

To accurately forecast remaining battery capacity and RUL, researchers have traditionally used empirical mathematical models to describe a cell's capacity loss as a function of time or cycle number. This approach, referred to as a model-based prognostics method (see Sec. 4.4 in Part 1), uses knowledge obtained from a small number of previous run-to-failure tests to inform researchers and practitioners about which mathematical model will most accurately capture the observed degradation trend. In the battery prognostics community, much work has been done to develop and test different empirical capacity degradation models for prognostics. Common mathematical functions used to model capacity loss as a function of time include linear \citep{honkura2011capacity}, exponential \citep{scott2005practical, miao2013remaining}, and power law functions \citep{attia2020revisiting, lui2021physics}, or combinations of them \citep{hu2014method, wang2013prognostics, hu2018remaining, he2011prognostics, diao2019accelerated}.

The most widely used implementation of an empirical capacity fade model is to use a recursive filtering technique to estimate the values of the model parameters and extrapolate the prediction to the end-of-life threshold at each time step, (see Secs. 4.2.2 and 4.2.4 in Part 1). The extrapolated prediction and its associated uncertainty can be used to generate a probabilistic RUL distribution (see Sec. \ref{sec:UQ and optimization}). Since many of the empirical capacity fade models are non-linear, Bayesian filtering methods like the extended Kalman filter and unscented Kalman filter are common choices \citep{plett2006sigma, plett2004extended}. However, the assumption of Gaussian noise in the Kalman filter is rather restrictive and sometimes not desirable. A popular alternative is to use a particle filter for battery RUL prediction because it is generally a more flexible approach that can easily switch between multiple models and estimate a non-parametric distribution, which is generally more exact \citep{walker2015comparison, he2011prognostics, miao2013remaining, saha2008prognostics}. 

In this work, we have chosen to use a power law model with two parameters, as it can accurately model the diverse capacity fade trends observed in the open-source dataset used in this study. The formulation for the power law model is as follows
\begin{equation}
\label{eq:power_law}
{Q_k} = 1 - {a}{k^b},
\end{equation}
where $Q_k$ is the normalized capacity of the cell at cycle $k$, and $a$ and $b$ are the model parameters which affect the shape of the estimated capacity fade trend \citep{attia2020revisiting, lui2021physics, diao2019accelerated}. To estimate the parameters $a$ and $b$, we chose to use a particle filter because it generally converges quicker and also provides non-parametric RUL distribution predictions. In a particle filter algorithm, the power law model is the \emph{measurement equation}, and the two \emph{state transition} equations for $a$ and $b$ are as follows
\begin{equation}
\label{eq:state_a}
{a_{k + 1}} = {a_k} + {w_{a,k+1}},
\end{equation}
\begin{equation}
\label{eq:state_b}
{b_{k + 1}} = {b_k} + {w_{b,k+1}},
\end{equation}
where $k$ denotes the cycle number and $w$ is the process noise terms, specific to $a$ and $b$. State estimation of model parameters and filtering are key components in a digital twin and have been discussed in detail in Secs. 4.2.2 and 4.2.4 in Part 1. Likewise, a detailed discussion regarding the mathematics for recursively updating a particle filter can be found in Appendix A of Part 1. Detailed pseudo code which accurately describes the steps one would take to implement a particle filter is presented in Fig. 28 of Part 1 and the working code used to implement the particle filter in this case study can be found on \cite{github_code}.

\subsubsection{Multi-attribute utility-based optimization model}
Multi-attribute utility theory, sometimes referred to as the acronym MAUT, is a method which aims to consider multiple optimization objectives of different scales or magnitudes by defining unique utility functions which map the quantities of interest to a common range for easier evaluation and optimization \citep{huber1974multi, winterfeldt1975multi}. Using this method, a unique utility function is defined for each value of interest (attributes) that is to be included in the optimization objective function. The utility function serves to scale the different attributes to a common range, typically [0,1]. If $n$ attributes are mutually preferential and independent, the general multi-attribute utility function is defined as follows \citep{engel2010multiattribute}
\begin{equation}
\label{eq:multi_attribute_utility}
\Lambda ({\upsilon _1},\;{\upsilon _2},\; \cdots ,\;{\upsilon _n}) = {1 \over n}\sum\limits_{i = 1}^n {{\varphi _i}({\upsilon _i})} ,
\end{equation}
where $\upsilon_i$ denotes the value of the $i$th attribute, ${\varphi _i}( \cdot )$ is the utility function over the values of the $i$th attribute, and $\Lambda ( \cdot )$ is the overall utility function. Then, the utility function is maximized by using any number of common optimization methods.

Multi-attribute utility theory has been widely applied in various maintenance scheduling applications because it is easy to incorporate non-traditional optimization objectives. For example, when servicing a large piece of equipment at a factory, it would be wise to incorporate knowledge of the company's sales when determining the optimal time to perform maintenance so that it has minimal impact on business. To achieve this, one can define a utility function which considers the company's sales, encoding their preference for maintenance times which will have a minimal impact on sales into the optimization problem. Similar work in the civil engineering field uses multi-attribute utility theory to schedule maintenance for bridges and other large structures \citep{sabatino2015sustainability, allah2019network}. When the number of utility functions is large, the design space increases drastically, and it becomes more important to select a proper solver, such as a genetic algorithm \citep{bukhsh2020multi, garmabaki2016maintenance}. In the energy field, multi-attribute utility theory has been applied to optimize the location of hybrid energy storage systems in the grid \citep{feng2018optimal}.

In this work, we take on the perspective of an EV manufacturer who has an agreement with a utility company to provide used EV batteries to them for grid-scale energy storage. We aim to optimize the time at which a Li-ion cell is retired from its first life application by considering two important attributes related to its first life use in an EV. The first attribute, the total discharge ampere-hours ($Ah$) before removal, encodes the preference that the cell should be used as much as possible in its first life. As the total Ah increases in value, so does the utility function. This utility function also encodes the preference to allow the customer to use the vehicle as much as possible and delay retirement as long as possible. The second attribute, the mean time between charges ($MTBC$), encodes the customers preference for a longer driving range on a single charge. This attribute correlates strongly with the health and remaining capacity of the cell, and conflicts with first attribute, total $Ah$. The mean time between charges will decrease as the cells ages, proportional to the capacity fade. By assuming that the two attributes are mutually preferential and independent, the two utility functions are combined using Eq. (\ref{eq:battery_multi_attribute_utility_2}) as follows
\begin{equation}\label{eq:battery_multi_attribute_utility_2}
\Lambda (Ah,\;MTBC) = \frac{1}{2}{\varphi _1}(Ah({x_c})) + \\ \frac{1}{2}{\varphi _2}(MTBC({x_c}))
\end{equation}
where $x_c$ is the decision variable (i.e., cycle number), ${\varphi _1}(Ah)$ and ${\varphi _2}(MTBC)$ are respectively the utility function of the total discharge ampere-hours (i.e., Ah) and the mean time between charges (i.e., MTBC) which are explained in details in Sec. \ref{sec:battery_optimization}, and $\Lambda (Ah,\;MTBC)$ is the overall utility function to be maximized.

\subsection{Results and discussion}
\subsubsection{Open-source battery dataset}
\label{sec:battery_dataset}
\begin{figure}[!h]
  \centering
    {\includegraphics[scale=0.70]{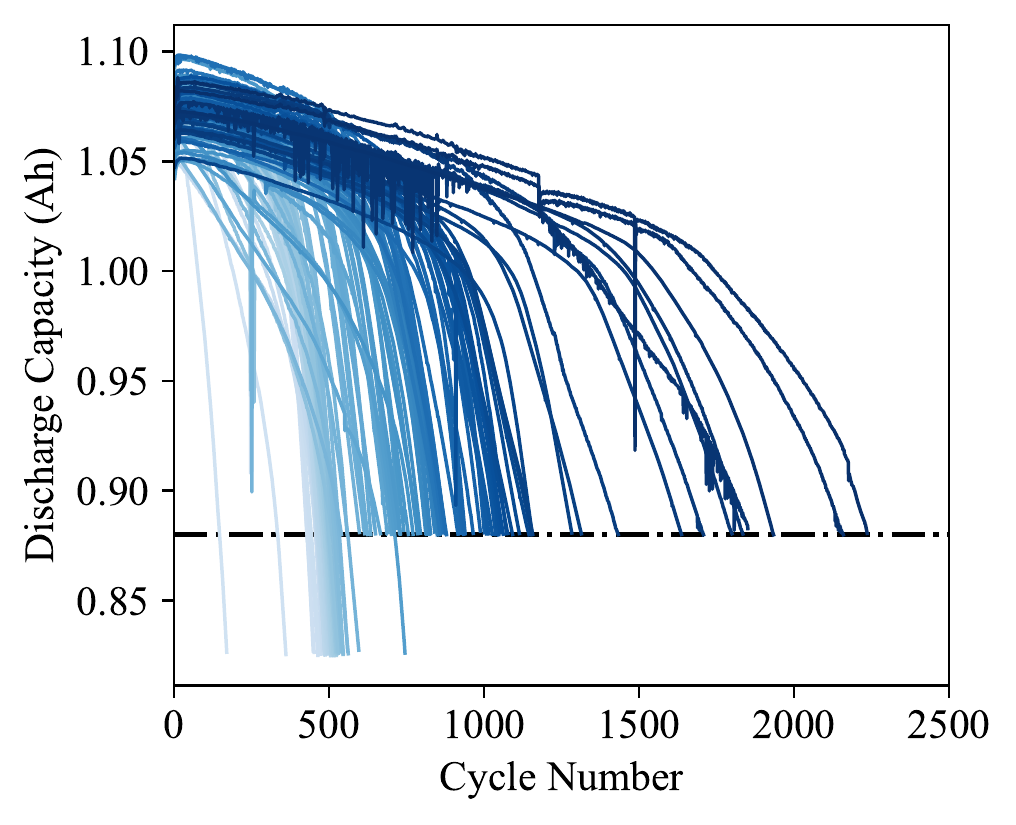}}
  \caption{MIT open-source battery dataset comprised of 124 LFP cells \citep{severson2019data}.}
  \label{fig:124_lfp_dataset}
\end{figure}

The open-source battery dataset used in this case study was released by the group in \cite{severson2019data}. The dataset consists of 124 lithium iron phosphate/graphite cells cycled with various two-step fast charging protocols. The discharge capacity fade curves for all the cells in the dataset are plotted in Fig. \ref{fig:124_lfp_dataset}. The variance in the capacity fade trajectories and lifetimes of the cells is due to the different fast charging protocols used. All cells are discharged at a constant current of 4C, where C denotes the amperage required to charge a battery from 0-100\% SOC in 1 hour. The large variance in the lifetime of the cells will provide a useful real-world assessment of the flexibility of the particle filter and optimization algorithms.

The dataset is divided into three groups, a \emph{training} dataset, and a \emph{primary} and \emph{secondary} test dataset. In this case study, we will use the training dataset to roughly determine suitable prognostic model hyperparmeters before testing the digital twin model online using the two test datasets. 

In both industry and academia, Li-ion cells are generally removed from their first life application when their remaining capacity approaches 80\% its initial value. Then, in a second life application, the cell would be further used until its remaining capacity is much lower, perhaps less than 60\%. We have therefore linearly extrapolated the last 30 capacity measurements of each cell well into the future, and set the end of second life limit at 50\% the initial capacity so that we have a more realistic problem to solve.

\subsubsection{Prognostic results}
\paragraph{Particle filter tuning}
An important step in building a prognostic model for online prediction is carefully selecting the initial hyperparameters. The particle filter model used in this study has five hyperparmeters that need to be initialized and tuned to maximize performance. The first parameter, the measurement noise, was set to $0.01$ to represent approximately 1\% error in the capacity measurements. The second and third parameters, the process noise terms for $a$ and $b$ from the empirical capacity fade model in Eqs. (\ref{eq:state_a}) and (\ref{eq:state_b}), were both set to $0.05$ because it produced acceptable results and reasonably quick convergence. The last two parameters, the initial values of $a$ and $b$, were set to the median values (${10^{ - 15.77}}$, $5.45$) as determined by fitting the empirical model to each of the cells in the training dataset.

In a real world scenario where the particle filter is to be deployed online, it would be best to perform a more rigorous search for the optimal hyperparmaeters. However, since our goal is mainly to shed light on the many digital twin enabling technologies and provide insights on how they can be used in a digital twin framework like the one proposed in this case study, we did not take the time to further optimize the hyperparmeters of the particle filter.

\begin{figure}[!h]
  \centering
    {\includegraphics[scale=0.75]{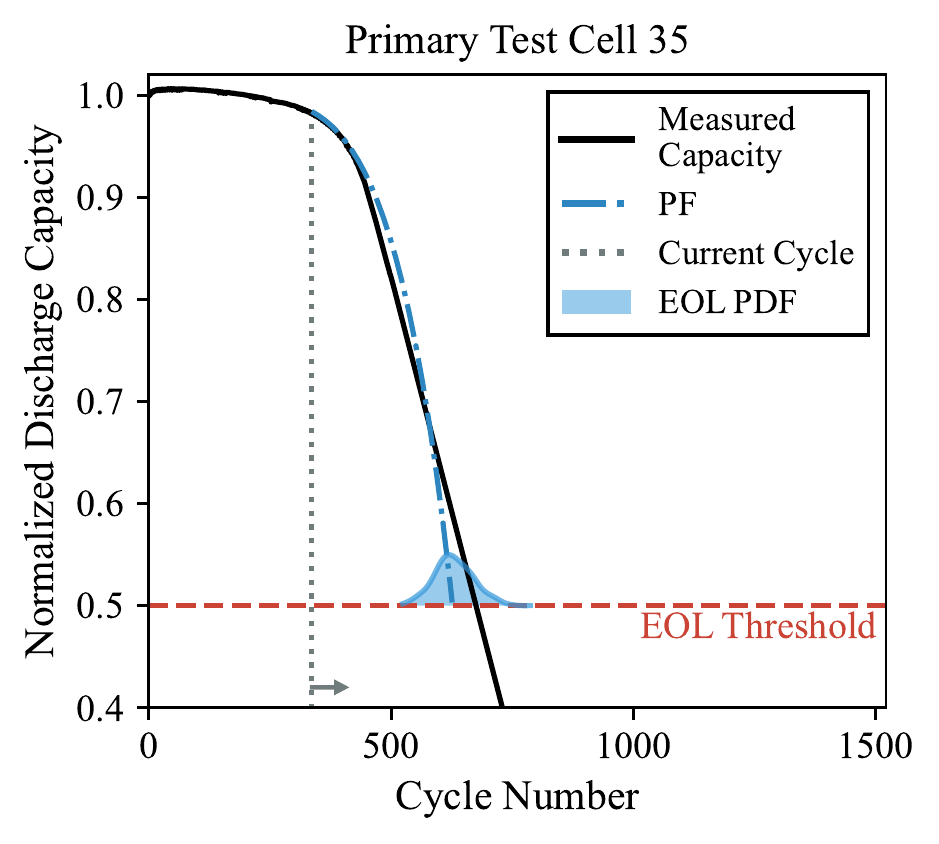}}
  \caption{Example particle filter capacity projection and empirical end-of-life distribution. EOL in the figure stands for \lq\lq end of life".}
  \label{fig:pf_eol_prediction}
\end{figure}

\paragraph{RUL prediction results}
First, we take a look at an example capacity projection from the particle filter. In Fig. \ref{fig:pf_eol_prediction}, we plot the measured capacity and the median projected capacity from the particle filter. Using the many particles, we can project many capacity curves into the future and determine an empirical end-of-life distribution, also shown in the figure.

\begin{figure}[!h]
  \centering
    {\includegraphics[scale=0.75]{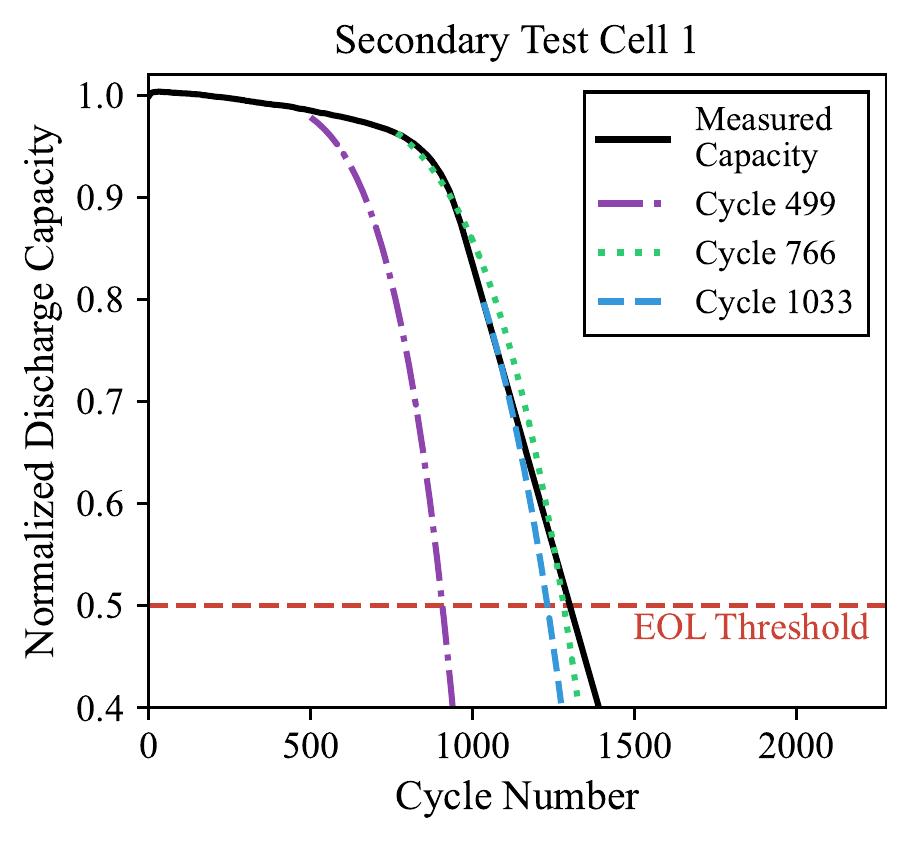}}
  \caption{Example particle filter capacity projections showing the underestimation of cell end of life. EOL in the figure stands for end of life.}
  \label{fig:many_pf_trajectory_predictions}
\end{figure}

\begin{figure*}[!ht]
  \centering
    {\includegraphics[scale=0.80]{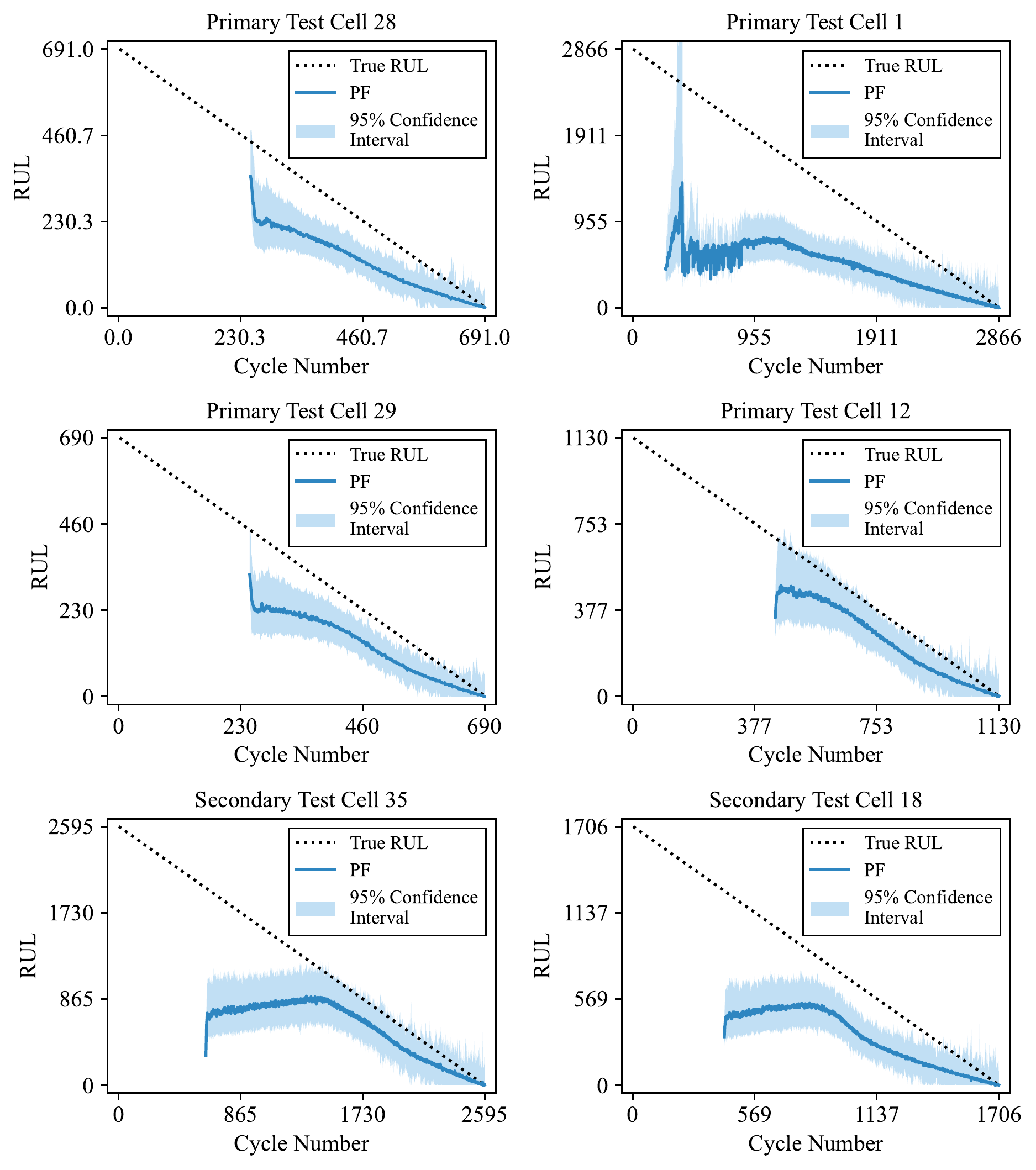}}
  \caption{Particle filter RUL predictions for six randomly selected cells from the two test datasets.}
  \label{fig:pf_panel_rul}
\end{figure*}

Figure \ref{fig:pf_panel_rul} shows some example RUL predictions from the particle filter. Right away, it can be seen that the particle filter mostly underestimates the RUL of the cells. The underestimation is further exacerbated for cells which have extremely long lifetimes ($> 2000$ cycles). This prediction pattern largely arises because of two factors. First, since the capacity fade trajectories of the cells were extrapolated using a linear model fit to the last 30 measurements, the late life capacity loss follows a linear trend (see Sec. \ref{sec:battery_dataset}). Therefore, we cannot expect the power law capacity fade model (Eq. (\ref{eq:power_law})) to accurately predict the long-term trend of the cells which have a linear degradation trajectory. This is visualized in Fig. \ref{fig:many_pf_trajectory_predictions} where we plot the measured capacity of a cell and the projections from the particle filter at different cycles.

Second, the initial values for the state parameters $a$ and $b$ were set to the median of the dataset, which encourages the initial lifetime prediction to be most accurate for cells of median lifetime. The median lifetime of training dataset is 750 cycles while the median lifetime of the two test datasets is 1125 cycles. The larger lifetime of the cells in the test datasets is an example of distribution shift (see Sec. 4.3 in Part 1), causing the prognostic model to generally underestimate test cell RUL. Additionally, the measurement and process noise parameters are likely not properly tuned. The current settings for the noise parameters does not allow the particle filter enough variability with each update of the states, causing it to slowly converge to the true RUL. There may be some additional performance to gain by rigorously optimizing the hyperparameters of the particle filter.

\subsubsection{Optimization results}
\label{sec:battery_optimization}
\paragraph{Defining the utility functions}
In order to optimize the time at which a cell is removed from its first life, we have to define the two utility functions which encode our preferences for how the cell is used in its first life. Drawing from previous literature, we aim to define two exponential utility functions. Used commonly in the finance industry, exponential utility functions are used to map real-world monetary gain to perceived value (utility). The general exponential utility function is defined as follows
\begin{equation}
{\varphi(\upsilon) = 1 - {e^{ - \upsilon/R}}},
\end{equation}
where $\varphi(\cdot)$ is the perceived utility, $\upsilon$ is the monetary gain, and $R$ is the risk tolerance. Additionally, the minimum monetary gain may be not always be zero, and the curve should be shifted to account for this. The exponential utility function can then be rewritten as follows
\begin{equation}
{\varphi(\upsilon) = \varsigma - \tau{e^{ - \upsilon/R}}},
\end{equation}
where $\varsigma $ and $\tau $ are scaling parameters which can be used to set the upper and lower limits, and are defined as 
\begin{equation}
{\varsigma = \frac{{{e^{ - L_u/R}}}}{{{e^{ - L_u/R}} - {e^{ - H_u/R}}}}},
\end{equation}
\begin{equation}
{\tau = \frac{1}{{{e^{ - L_u/R}} - {e^{ - H_u/R}}}}},
\end{equation}
where $R$ is the risk tolerance that defines the shape of the exponential curve, and $L_u$ and $H_u$ are the lower and upper limits to the utility function. 

\begin{figure}[!h]
  \centering
    {\includegraphics[scale=0.60]{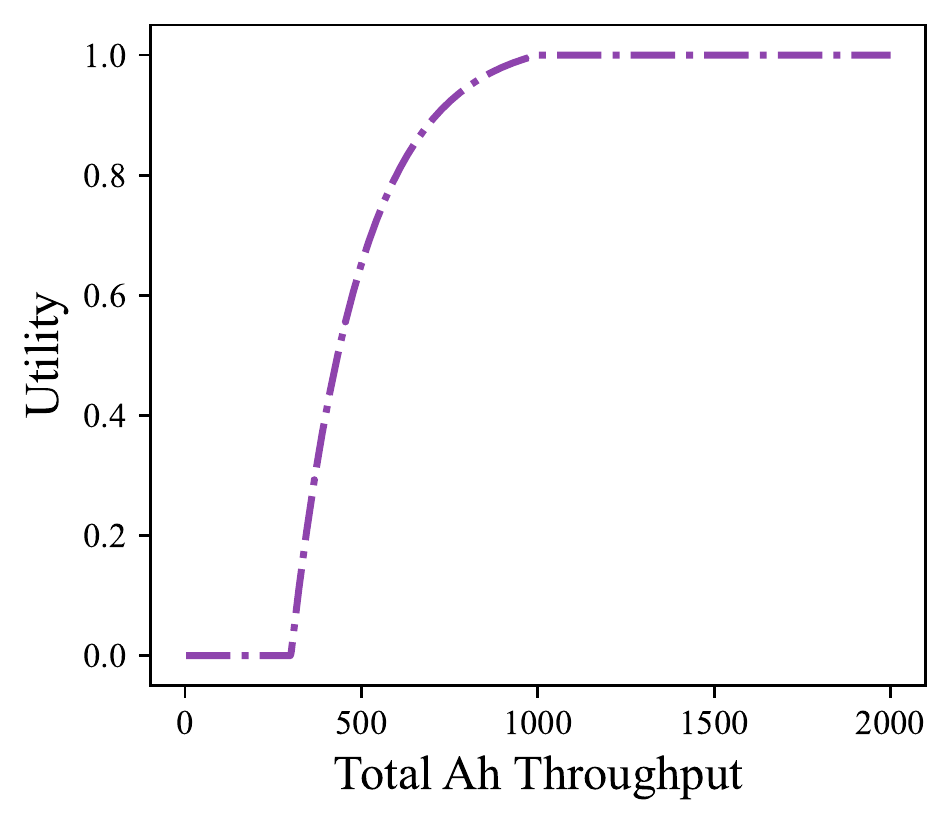}}
  \caption{Exponential utility function for total Ah throughput.}
  \label{fig:Ah_utility}
\end{figure}

We adopt a similar approach for crafting the utility functions used in the battery digital twin. The first utility function maps the total Ah throughput from the cell to the range $[0,1]$. This utility function is increasing, and takes the following form
\begin{equation}
    {{\varphi_1}(Ah(x_c)) = 1.0311 - 4.6212{e^{ - Ah(x_c)/200}}},
\end{equation}
where the parameters $\varsigma $ and $\tau $, defined by the lower and upper bounds $L_u$ and $H_u$ and the shape parameter $R$ are determined by examining the total Ah throughput for cells in the training dataset. From the electric vehicle manufacturer point of view, vehicles are typically rated to last a certain number of miles, years, or charge/discharge cycles. The bounds for this utility function are used to encode this requirement. For this case study, $L_u$, $H_u$, and $R$ were set to 300 Ah, 1000 Ah, and 200, respectively. These bounds are rounded values which are close to the 5th and 95th percentiles of the training dataset. The utility function for the total $Ah$ throughput is shown in Fig. \ref{fig:Ah_utility}. 

\begin{figure}[!h]
  \centering
    {\includegraphics[scale=0.60]{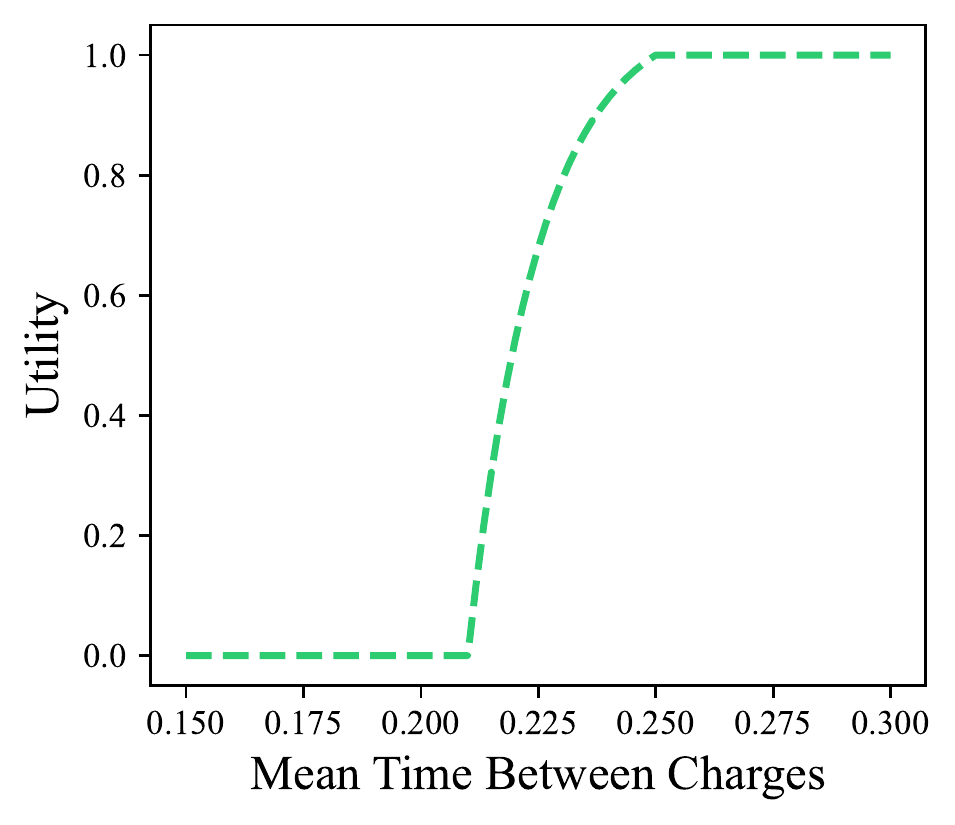}}
  \caption{Exponential utility function for the mean time between charges.}
  \label{fig:MTBC_utility}
\end{figure}

The second utility function maps the mean time between charges ($MTBC$) to the range $[0,1]$. This utility function is also increasing, but is conflicting with the first utility function because it is affected by cell aging and capacity loss. The function takes the following form
\begin{equation}\label{eq:battery_multi_attribute_utility}
\begin{split}
{\varphi _2}(MTBC({x_c})) = 1.0746 \\
- 1292405{e^{ - MTBC({x_c})/0.015}}
\end{split}
\end{equation}
where once again, the parameters $\varsigma $ and $\tau $, defined by the lower and upper bounds $L_u$ and $H_u$ and the shape parameter $R$ are determined by examining typical values for the cells in the training dataset. As the cell ages, the mean time between charges decreases, meaning that a customer driving the vehicle will have to recharge more frequently. This is not ideal, as customers prefer that the vehicle be as efficient as possible, having the largest possible range on a single charge. The parameters for this utility function are set in such a way to consider the diminishing utility a customer gets from a vehicle that needs to recharge too frequently. For this case study, $L_u$, $H_u$, and $R$ were set to 0.21 h, 0.25 h, and 0.015, respectively. The utility function for the mean time between charges is shown in Fig. \ref{fig:MTBC_utility}. 

To determine the optimal replacement time, the two utility functions are combined into a single meta-utility function using an equal weight average (see Eq. (\ref{eq:battery_multi_attribute_utility_2})). Then, the combined utility function can be maximized by minimizing its negative using any off-the-shelf solver. However, the simple utility functions used in this case study are easy to evaluate, and the range of possible cycles at which the cell can be retired from its first life is finite, so we can simply evaluate the combined utility function at every possible cycle, and plot the utility curve to get a better idea of what is going on.

\begin{figure}[!h]
  \centering
    {\includegraphics[scale=0.7]{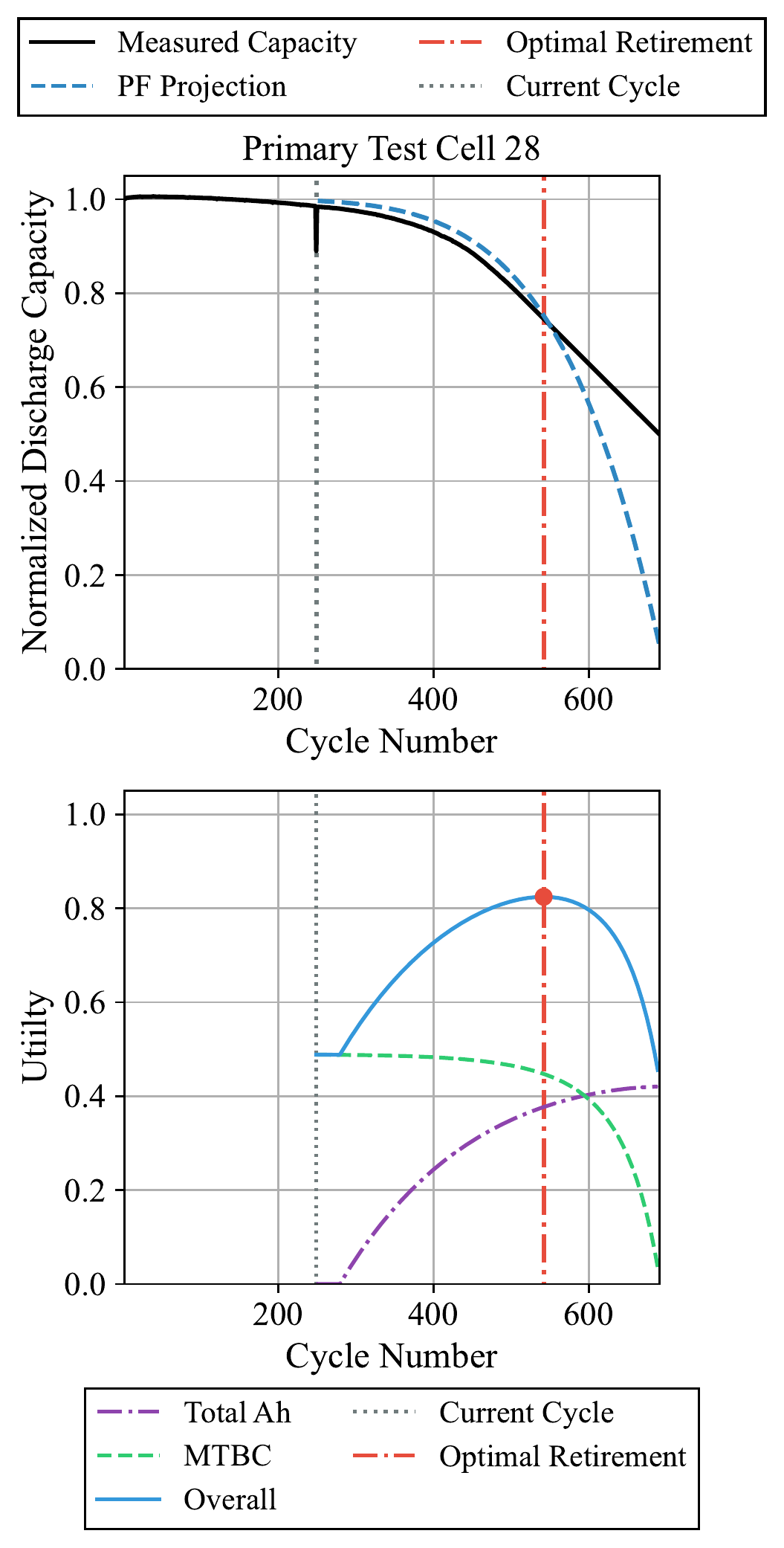}}
  \caption{Optimal cycle to retire primary test cell 28 from its first life application in an electric vehicle using the projected capacity from the particle filter.}
  \label{fig:opt_cell27}
\end{figure}

\begin{figure}[!h]
  \centering
    {\includegraphics[scale=0.7]{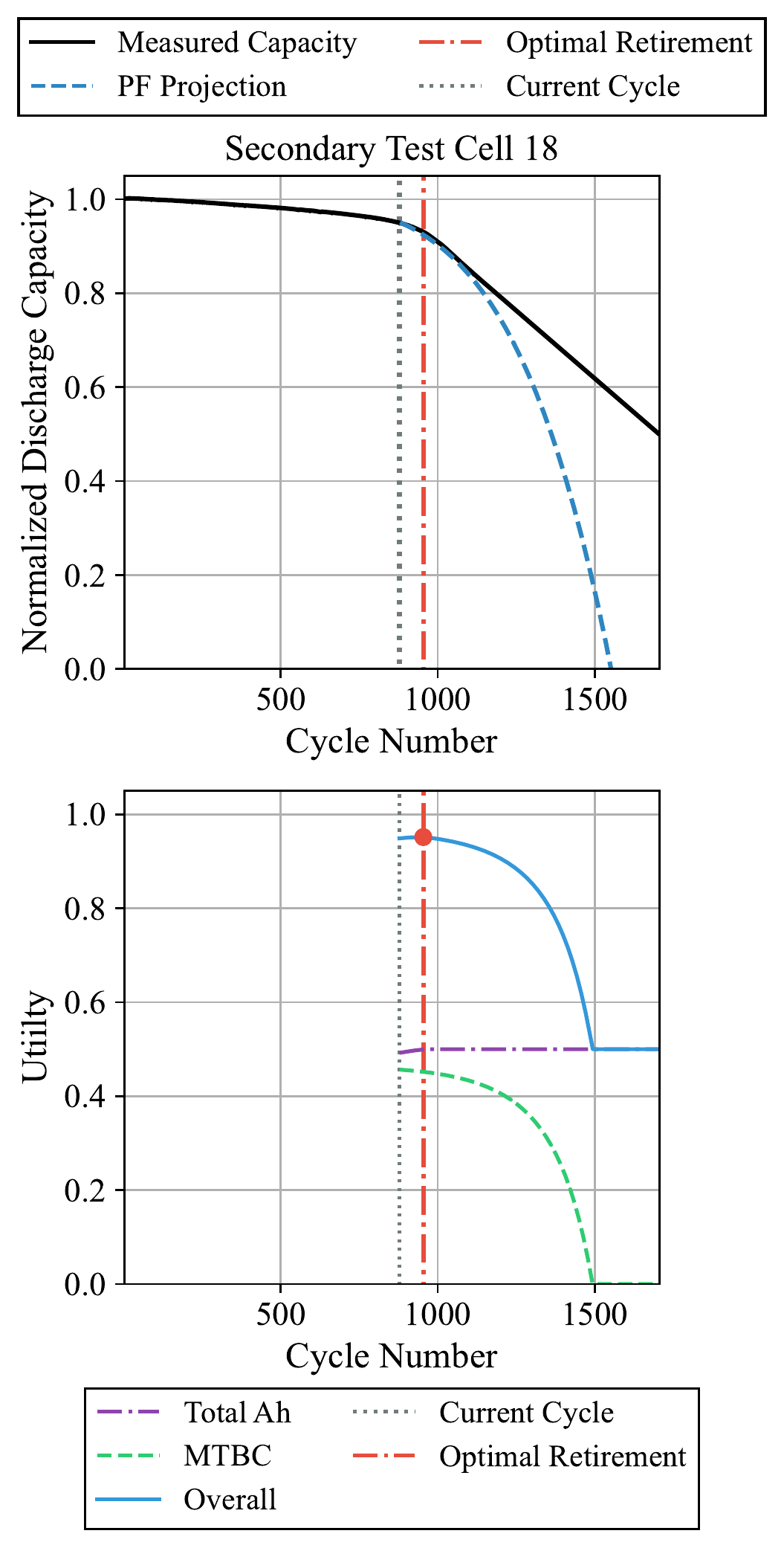}}
  \caption{Optimal cycle to retire secondary test cell 18 from its first life application in an electric vehicle using the projected capacity from the particle filter.}
  \label{fig:opt_cell59}
\end{figure}

\begin{figure}[!h]
  \centering
    {\includegraphics[scale=0.7]{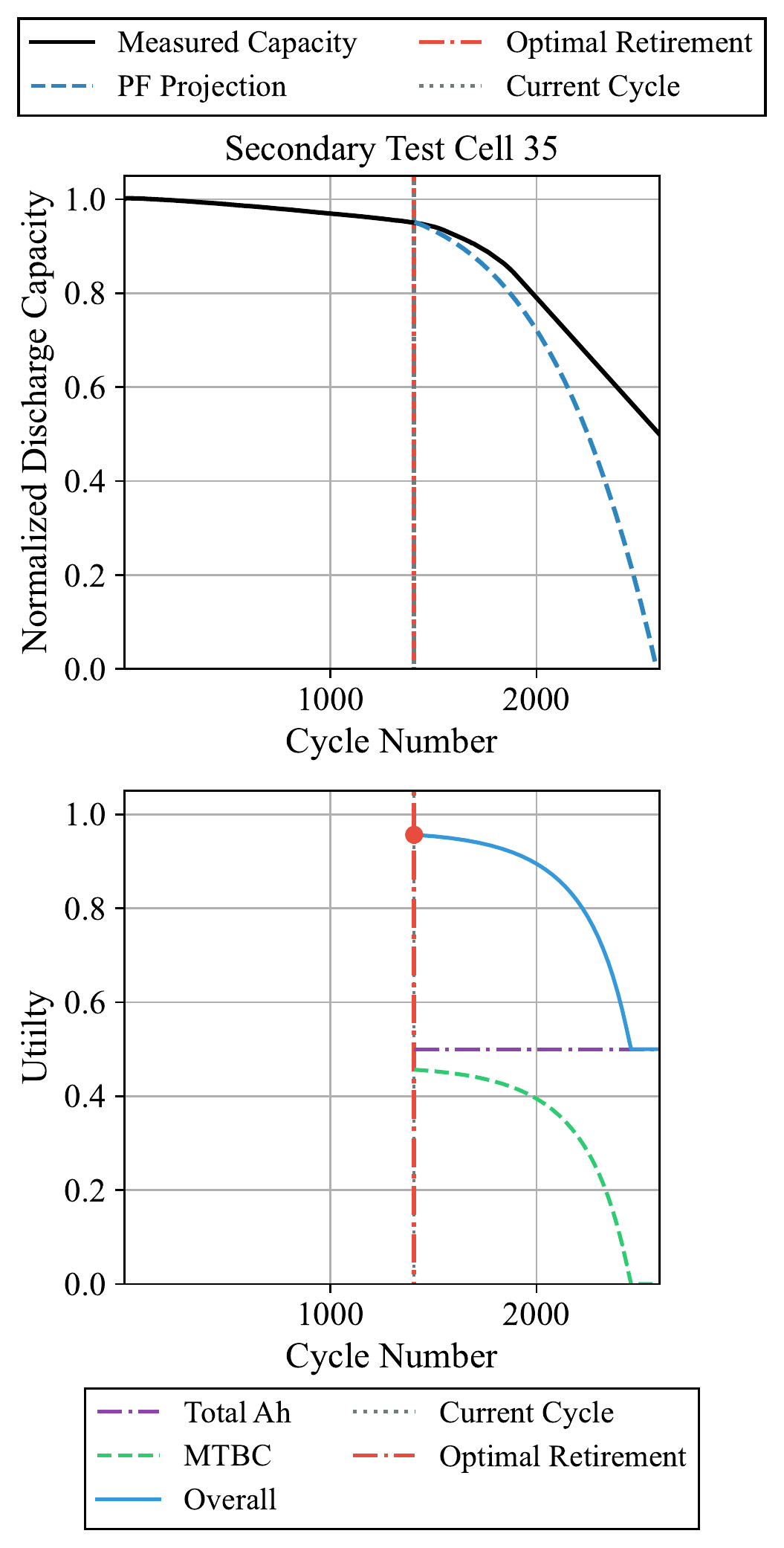}}
  \caption{Optimal cycle to retire secondary test cell 35 from its first life application in an electric vehicle using the projected capacity from the particle filter.}
  \label{fig:opt_cell76}
\end{figure}

\paragraph{Analyzing the optimization results}
Optimization results for three of the cells shown earlier (primary test cell 28 and secondary test cells 18 and 35) are in Figs. \ref{fig:opt_cell27}, \ref{fig:opt_cell59} and \ref{fig:opt_cell76}. These cells were chosen because they have different lifetimes (average, long, and longest, respectively), and provide a good contrast in how the optimization method might perform in the field.

Looking at all three figures together, the first thing we observe is that the cells with longer lifetimes, cells 18 (Fig. \ref{fig:opt_cell59}) and 35 (Fig. \ref{fig:opt_cell76}) from the secondary test datset, have optimal retirements which are much closer to the current cycle. This is because the utility function for the total Ah throughput was defined around the mean values for $Ah$ throughput of the training dataset, which were previously mentioned to have lower lifetimes than the two test datasets (median lifetimes for the training and test datasets are 750 and 1125 cycles respectively). By defining the utility function centered around the training cells with slightly lower lifetimes, it causes any cells with longer lifetimes, like cells 18 and 35, to have maximum $Ah$ throughput utility for all possible retirement cycles. Since the $Ah$ throughput utility is nearly constant over possible retirement cycles, this forces the optimal retirement decision to be made entirely based on the mean time between charges, which only decreases over the lifetime of the cell. Ultimately, this causes the optimal retirement to be as early as possible, to maximize the mean time between charges utility function.

For cells with lifetimes closer to the median of the training dataset, like cell 28 from the primary test dataset (Fig. \ref{fig:opt_cell27}), the utility functions work as intended, and the optimium retirement cycle strikes a perfect balance of the two utility functions. 

In general, we found that the optimal time to retire the cell is most closely related to the observed rate of capacity loss (essentially the slope of the capacity vs cycle number curve). Typically, the optimal replacement time is shortly after the cell's knee point, which is the bend in the capacity fade curve where more rapid capacity loss begins to occur. Intuitively, this makes sense, as the total $Ah$ utility can remain high, because the cell is still able to output a great deal of power even after its knee point, but the mean time between charges begins to rapidly increase, because the capacity is fading more quickly. As the capacity fades, the mean time between charges decreases, and the customer has to charge their vehicle more frequently, which is not desirable. In turn, the utility function for the mean time between charges decreases in value to reflect this.

Lastly, in testing, we observed that the optimal retirement time was not drastically affected by the particle filter's projected shape of the capacity fade trajectory. As long as the shape was fairly similar to that of the measured curve, the results were consistent. What we did notice, however, was that for cells with longer lifetimes, the particle filter would significantly underestimate the capacity trajectory, and the optimal replacement time would move closer to the current cycle to account for the perceived earlier EOL.

\subsection{Case study conclusion and ideas for future research}
Altogether, this case study demonstrated how one could create a battery digital twin for optimizing the retirement time of a Li-ion cell from its first life use. The success of the proposed digital twin framework relies on the integration of multiple individual pieces of software, which together, form an intelligent model capable of informing engineers and practitioners of optimal retirement times on a cell-by-cell basis. 

The first key piece of software, the particle filter prognostic model, was found to accurately predict the RUL of cells with varying lifetimes. However, the particle filter used in this study was limited in that it only considered a single capacity fade model, the power law model. In future research, it would be worthwhile to investigate a multi-model particle filter, which actively switches between different capacity fade models \citep{li2021remaining, ye2018double, boers2003interacting}.

The second key piece of software, the multi-attribute utility optimization model, was found to provide advanced notice of the optimal time to retire a cell from its first life application using the projected future capacity from the particle filter model. However, the optimization model struggled to provide meaningful optimal replacement predictions for cells which have much longer lifetimes. In future work, it would prove useful to investigate better methods of defining the many utility functions so that they can be applied to a wider range of cells. Additionally, this case study did not consider any preferences (attributes) relevant to the second life application of the Li-ion cell. The preferences of the second-life user were not considered for two main reasons. First, this was done to simplify the case study, as the goal was to demonstrate and explain in great detail the different software components that ultimately comprise the proposed digital twin model. Second, not considering the preferences of the second-life application simplified the optimization model, making it much easier to visualize and explain how it works. In reality, it is imperative to consider the preferences of all the involved parties when determining the optimal time to remove a Li-ion cell from its first life application. Both the electric vehicle manufacturer and second life grid-scale energy storage management company would have input on the attributes used in the optimization model, so that both parties' preferences are considered when determining the optimal time to switch from the first to second life application. However, we leave this to future work.

Lastly, the main strength of the particle filter is that it can output a non-parametric probabilistic prediction of a cell’s capacity trajectory and end of life. However, this case study used a deterministic method to determine the optimal time to retire a cell from its first life application. In practice, it is desirable to present practitioners with confidence levels and intervals, instead of point estimates. Therefore, in future work, it would prove insightful to investigate probabilistic methods of evaluating multi-attribute utility functions, so that an optimal replacement interval can be defined, instead of a single-point prediction. Since the utility function approach to optimization is relatively lightweight, it may prove tractable to use sampling approaches to yield probabilistic results. Similar ideas are further discussed in Sec. 5.2 of Part 1 and Sec. \ref{sec:predictive_maintenance_scheduling}, a probabilistic decision-making framework that considers model uncertainty (see Sec. \ref{sec:UQ_digital_twin}) empowers maintenance engineers and practitioners with more information to make the bests decisions possible.


\section{Demonstration and open source}
\label{sec5}

\subsection{Industry-scale demonstration of digital twin}
\label{sec:industry_demo}
Digital twins have become one of the key ingredients in improving the business outcomes across a wide range of industries. The global digital twin market is projected to grow from USD 3.1 billion in 2020 to USD 48.2 billion by 2026~\citep{mehra_2020}. In many industries, digital twins have emerged as an integral part of the digitization efforts that started almost two decades ago. While the first wave of digitization focused on reducing physical databases and interactions, the second wave happening now is when the digital twin concept really kicked-off. Specifically, the second wave of digitization focuses on creating a virtual model encompassing a physical asset or process throughout its life cycle so as to analyze, predict and optimize the asset and improve business outcomes. Most use cases of digital twins in industry fall under the categories of predictive maintenance, asset life cycle management, process planning \& optimization, product design, virtual prototyping and more \citep{sjarov2020digital}. The benefit of incorporating digital twins is multi-fold when solving various engineering and business problems, such as reducing costs and risk, improving efficiency, security, reliability, resilience, and supporting decision-making~\citep{vanderhorn2021digital}. Digital twins have been successfully applied to a broad range of industries, such as manufacturing, software, aerospace, agriculture, automobiles, healthcare, consumer goods, etc, and the usage of digital twins is still growing at a fast speed~\citep{durao2018digital,augustine2020industry}.

In the aerospace industry, digital twin effort and conceptualization was pioneered and directed by NASA and the U.S. Air Force in early 2000s (see, for example, ~\cite{glaessgen2012digital}). Since then, digital twin has been expanded to many areas, such as airframe, avionics, crack detection, feet level health management, etc. Digital twin has also been one of critical technologies that is listed as a part of future strategy for many leading companies, such as Boeing and Airbus~\citep{aydemir2020digital}. In GE electric, more than 500,000 digital twins have been deployed in various capacities (i.e., parts twins, products twins, process twins and system twins) across its production sites~\cite{}. In addition, GE aviation has reduced the amount of time for anomaly detections based on sensor data by 15-30 days. The reduction of anomaly detection time helps to achieve \$44 million savings in the life cycle management for turbine blade engines, and \$10 million savings in dynamic optimization to deliver optimal flight patterns and maintenance. 

In the commercial sector, Boeing has reported 40\% improvement in first-time quality of the parts and systems \citep{boeing40}. They have been creating and maintaining operational digital twins of components to track each individual component’s unique characteristics and detect degradation rates. Similarly, in the defense sector, Boeing is using digital twins to predict and find possible fatigue maintenance hot spots in the F15 Eagle to reduce operation and maintenance costs. More recently, Boeing is planning to build an airplane in the metaverse using a digital twin \citep{boeing_metaverse}.

In the automobile industry, digital twins have been playing an important role across all the stages of vehicle life cycle development, such as concept development, detailed design, and design verification \citep{sharma2018digital}. In addition, digital twins have been used to improve the manufacturing and production in the automobile manufacturing plants. For example, BMW has recently used a virtual factory built upon digital twins and have produced 30\% more efficient planning processes \citep{13_caulfield_2022}. In the healthcare sector, one of the main application areas lies in the development of technologies for monitoring, digital surgery, remote surgical assistance, drug development, medical treatment etc. using high-fidelity digital twin models of human bodies~\citep{skardal2016organoid, bruynseels2018digital,liu2019novel}. A digital twin can also be used in building a virtual model of a physical city, also known as “smart city”, to map the urban information system to into space and time. It has been used to solve a broad range of problems, such as water treatment, maintenance of facilities, and so on \citep{chen2018digital,lehner2020digital,lu2020developing}.

Additionally, digital companies like Siemens, Ansys, IBM, and others have been playing a pivotal role in providing the infrastructure and platform to support the adoptions of digital twin by both small and large-scale industries. For example, Siemens has been one of the leading companies developing digital twin technologies for products, production and performance  of various mechanics, electronics, software, or systems to minimize cost and optimize the requirements \citep{siemens_newletter_2020}. Some of Siemen's product like Simatic Real-time Locating Systems (RTLS), Teamcenter X software, and UltraSoC \citep{mehra_2020} are representative examples of Siemen's digital twin product. Ansys has demonstrated applications of digital twins for Kärcher \citep{adv_mag_2021}, ABB \citep{ boscaglia2019conjugate}, ENGIE \citep{de_2021} and others. IBM has applied digital twins in different sectors like life cycle management, healthcare, predictive maintenance, manufacturing and more \citep{ ibm_newsletter_2020, ibm_blog_2019, ibm_blog_2020, ibm_white_paper_2020}.

\subsection{Open-source tools and datasets related to digital twins}
\label{sec:open_source}

We have also identified publicly available, free-to-use tools and datasets related to digital twins through our literature review. Tables \ref{tab:tools} and \ref{tab:dataset} show, respectively, the publicly available tools and datasets. The roles of these tools and datasets in digital twins range from modeling, process mining, control, data streaming, and digital twin projects. These open-source tools and datasets could be helpful for production-level implementations of digital twins in both academia and industry. In what follows, we describe a few representative examples of open-source tools and datasets. Interested readers are again invited to consult Tables \ref{tab:tools} and \ref{tab:dataset} for an overview of the tools and datasets we have identified. 

\begin{table*}[!ht]
    \centering
    \caption{Publicly available tools related to digital twins (DS: digital system; P2V: physical to virtual; V2P: virtual to physical)}
    \begin{tabular}{ p{2.3cm}|p{2.1cm}|p{2.5cm}|p{2cm}|p{4cm} }
     \hline \hline
     \textbf{Name (link)} & \textbf{Organization} & \textbf{Application(s)} & \textbf{Role(s) in digital twins} & \textbf{Brief description (paper ref)}\\
     \hline
     TE Code (\cite{tecode}) & UW & Manufacturing process control & DS, P2V, V2P & A simulation program for controlling chemical processes, known as the Tennessee Eastman Plant-wide Industrial Process Control Problem (\cite{ricker1995nonlinear})\\
     \hline
     Chrono (\cite{chrono}; \cite{tasora2015chrono}) & UW-Madison, UNIPR & Multiphysics simulation & DS & A modelling and simulation infrastructure for multiphysics simulations, including multibody simulation, finite element analysis, and fluid-solid interaction\\
     \hline
     ProM (\cite{promtools}; \cite{van2005prom, verbeek2010prom}) & TU/e & Process Mining & DS, P2V, V2P & Business process mining software\\
    \hline
    PyMC3 (\cite{pymc3}; \cite{salvatier2016probabilistic}) & Collaborative & Multiple & P2V & A Python package for Bayesian modeling and probabilistic machine learning \\
    \hline
    QUESO (\cite{queso}; \cite{Prudencio2012}) & UT-Austin & Mechanical &  P2V & A collection of algorithms and C++ classes that can be used for offline, probabilistic model updating\\
    \hline
    mFUSE (\cite{mFUSE2022};) & LANL/UCSD Engineering Institute & Mechanical, Civil, Structural &  P2V & A set of MATLAB functions organized into modules according to the three primary stages of SHM\\
  \hline
    Digital Twin Open-Source Repository (\cite{digitaltwinconsortium}) & Digital Twin Consortium & Manufacturing, mechanical, healthcare, etc. & DS, P2V, V2P & Open-source code implementations, collaborative documents for guidance and training, open-source models, or other assets related to digital twins\\
    \hline
    Dito (\cite{dito}) & Eclipse Foundation & IoT middleware & P2V, V2P & IoT middleware, focusing on data modelling and IoT device connectivity \\
    \hline
    Unreal Engine (\cite{unrealengine}) & Epic Games, Inc. & Transportation, mixed reality, etc. & DS, P2V & Real-time 3D modeling software \\
     \hline \hline
    \end{tabular}
    \label{tab:tools}
\end{table*}

\paragraph{Open-source software}
\begin{itemize}
\item \textbf{Chrono:} Chrono is an open-source multiphysics modeling and simulation infrastructure capable of multibody dynamics simulation, finite element analysis, and fluid-solid interaction. Its core is a C++ object-oriented library called the Chrono::Engine, which can be embedded in third-party applications, as part of a large digital twin project. As noted on the PROJECTCHRONO web page, its multibody dynamics capability allows for vehicle dynamics simulations where wheeled vehicles may operate on soft, deformable soils. This effort started in 1998, and the current version, Chrono 7.0.1, was released in November 2021. It is rare to see such a long-lasting open-source effort, which has substantially pushed the boundary of open-source middleware and fostered an ecosystem of multiphysics software tools. 
\item \textbf{Digital twin repositories:} These open-source repositories are being managed by a non-profit organization called Digital Twin Consortium as their \lq\lq open collaboration initiative”. The repositories are hosted on GitHub ({\url{ https://github.com/digitaltwinconsortium}}) and currently include several industry-relevant projects focused on manufacturing digitatlization (e.g., ManufacturingDTDLOntologies and EcolCafe-Industrie-4.0). The open-source nature of these code implementations, digital models, and training and guidance documents will drive development, interoperability, and adoption of digital twins in industry. 
\item \textbf{Commercial software tools:} Commercially available software tools enabling the realization of digital twins include (1) CAD software for geometric modeling, such as CATIA, Autodesk AutoCAD, Siemens NX, PTC Creo, SolidWorks, to name a few, (2) 3D manufacturing simulation software for production simulation and layout design, such as DELMIA and Tecnomatix, and (3) modeling and data analytics software for operations optimization and predictive maintenance, such as MATLAB (the Simulink modeling environment and Predictive Maintenance Toolbox), and (4) cloud-based IIoT platforms for predictive maintenance, including Siemens Mindsphere, Azure IoT Edge, Google Cloud, and Azure IoT Edge. Although not serving as digital twins on their own, these software tools have created success stories in realizing the digital twin concept in industry settings by working collaboratively with other enabling software tools. What is now needed are frameworks and software platforms that cohesively integrate modeling and simulation efforts of different forms and at various levels of detail to create complete digital twins in various engineering applications (\cite{malik2021framework, malik2021digital}). In this direction, some efforts have been undertaken in the commercial design software domain. For example, ESTECO modeFRONTIER is a commercially available design software tool that integrates and automates multiple CAD/CAE tools to create digital twins for engineering design. Another example is ....
\end{itemize}

\paragraph{Open-source data}
\begin{itemize}
\item \textbf{Bosch Production Line Performance Dataset:} this dataset was introduced in 2016 as part of a data challenge whose goal was to reduce quality control escapes. Participants in the data challenge were to develop a model that could predict which parts will fail quality control by using extensive measurements taken at different locations along the assembly line. Large datasets like the one provided by Bosch represent the first steps toward realizing digital twin models in the production setting. 
\item \textbf{NASA Prognostics Data Repository:} NASA’s Ames Research Center hosts 18 well-documented datasets, collectively called the Prognostics Data Repository, on their Prognostics Center of Excellence website. These datasets are mostly run-to-failure time series data collected from an engineered system or component starting at a healthy state and ending at a failed state. They were contributed by universities (e.g., ETH Zurich, the FEMTO-ST Institute, UC Berkeley, the University of Cincinnati, and Arizona State University), companies (e.g., PARC and Sentient Corporation), and government agencies (e.g., NASA Ames). This data repository intends to increase data availability for prognostic algorithm development and validation. Among the 18 datasets is (1) the first publicly available battery aging dataset \cite{saha2007battery} that has led to the production of the first few battery prognostics papers \cite{saha2008prognostics, goebel2008prognostics, saha2009comparison} and (2) the first two publicly available engine degradation datasets \cite{saxena2008phm08, saxena2008turbofan} that has led to the creation of numerous data-driven prognostics algorithms in the literature \cite{wang2008similarity, heimes2008recurrent,hu2012ensemble,moghaddass2014integrated,zhang2016multiobjective,zhao2017remaining}. This data repository is actively being expanded, and its impact on the field of predictive maintenance, which has been profound, will be increasingly significant.

\item \textbf{Data Challenges in PHM and SHM:} In order to foster and encourage innovations in the PHM and SHM communities, data challenges/competitions have been organized in different PHM and SHM conferences. The released open-source datasets after the competitions have been extensively used in the literature as benchmark to test the performance of different diagnostics and prognostics algorithms (i.e., P2V connection). For instance, the datasets from the PHM Data Challenge of the PHM08 conference have been posted at NASA prognostics data repository, and have become one of the first publicly available datasets for failure prognostics. Since the data challenge in the PHM08 conference, many similar data challenges for fault diagnostics and failure prognostics have been organized by different PHM conferences including the annual PHM conference \citep{PHM2022}, the European PHM conference \citep{PHME2022}, and the  Asia-Pacific PHM Conference \citep{PHMAP2021}. In the SHM community, the first data compitition was organized in 2020 by the Asia-Pacific Network of Centers for Research in Smart Structures Technology, Harbin Institute of Technology, the University of Illinois at Urbana-Champaign \citep{ICSHM2020}. In this competition, algorithms are compared in terms of accuracy for detecting fatigue crack based on images, anomaly of bridge using acceleration data, and damage of cables based on monitored cable tension data. Even though the addressed fundamental fault diagnostics and failure prognostics problems are similar, the data competition in the SHM community focuses more on fault diagnostics, anomaly detection, and prognostics of large civil infrastructures. The popularity of these data challenges shows the importance of common datasets for the development of new algorithms in establishing the P2V connection in digital twins. 
\item \textbf{Battery Archive:} This web-based platform, launched in 2020, is the first public platform that allows the battery community to visualize, analyze, and compare battery aging data across different organizations, studies, chemistries, and aging conditions. The site development is led by Sandia National Laboratories (SNL), a U.S. Department of Energy (DOE) sponsored Federally Funded Research and Development Center. The Battery Archive platform is built based on an open-source platform called the Battery Life Cycle Framework \cite{de2021battery} and uses an open-source extract-transform-load engine called Redash. It has two unique properties: (1) all the battery datasets shared on the platform share a standard format, making it easy to compare performance across studies, battery chemistries, and aging conditions, and (2) the platform has an open-source backend, opening the possibilities of interoperating with existing software tools for more advanced analytics. Compared to the NASA Prognostics Data Repository, Battery Archive goes one step further by standardizing the data format and setting up open-source tools for data visualization and basic analysis. This further step is expected to have a long-last impact on future research studies surrounding battery degradation modeling. Further, we think Battery Archive represents an exemplary data-sharing effort that could be scaled up to benefit other digital twin applications. 
\item \textbf{Sandia Thermal Challenge Problem:} This thermal challenge problem was launched by SNL in 2006 and has been used extensively as a benchmarking dataset to evaluate the performance of methods and approaches for experimental validation of computer simulation models and quantification of model uncertainty \citep{hills2008thermal}. It is the first publicly available dataset to evaluate the performance of different UQ methods for model calibration and validation. The mathematical model of one-dimensional, linear heat conduction in a solid slab and the associated experimental data are provided by \cite{dowding2008formulation}. The problem is concered about (1) evaluating whether the prediction accuracy of a calibrated/validated model meets a regulatory requirement and (2) quantifying the uncertainty in the accuracy assessment.  In the past decades, solutions to this challenge problem (model validation approaches and results) have been developed using both Bayesian (e.g., the KOH framework discussed in Sec. \ref{sec:UQ_dynamic_system_models}) and non-Bayesian methods. The problem is formulated in a way that it is independent from specific applications and representative of the complexities of realistic physical systems.

\end{itemize}

\paragraph{More collaborative, open-source efforts are needed.}
The industry-scale adoption of digital twins will be accelerated by increasing efforts in sharing tools (e.g., applications, libraries, and plug-ins), data, and best practices (e.g., tutorials, implementation guides, and training materials) between industry, academia, and government. Open-source efforts by academic researchers are often directed to create rigorous, application-agnostic solutions for specific technical components of digital twins (e.g., probabilistic model calibration, MPC). In contrast, open-source efforts by industrial practitioners tend to be focused on integrating these technical pieces into easy-to-use, well-documented software platforms thoughtfully customized to individual applications. These two groups of open-source efforts, representing two distinct ranges of technology readiness levels (TRL), are complementary. The cross-sharing of knowledge among these communities of interest will grow an open-source knowledge pool and promote the development of high quality commercial-grade software for digital twins. In this review, we want to call for the academic, industrial, and government communities to work together and share their software tools and data. Such collaborative, open-source efforts will go a long way in shaping the future of the digital twin field.

\begin{table*}[!ht]
    \centering
    \caption{Publicly available datasets related to digital twin}
    \begin{tabular}{p{2.2cm}|p{2.2cm}|p{2.5cm}|p{2cm}|p{4.5cm}}
     \hline \hline
     \textbf{Name (link)} & \textbf{Organization} & \textbf{Application(s)} & \textbf{Role(s) in digital twins} & \textbf{Brief description (paper ref)}\\
     \hline
     NASA Prognostics Data Repository (\cite{pcoedatasets}) & NASA & Predictive maintenance & P2V, V2P, OPT & A collection of 18 prognostic datasets for prognostic algorithm development \\
     \hline
     Battery Archive (\cite{batteryarchive}) & DOE SNL & Battery degradation modeling & P2V, OPT & A web-based platform for visualizing, analyzing, and comparing battery aging data across different studies (\cite{de2021battery})  \\
     \hline
     Thermal Challenge Problem (\cite{hills2008thermal}) & DOE SNL & Model validation and UQ & P2V, OPT & A mathematical model and the solution of a one-dimensional, linear heat conduction in a solid slab. Experimental data related to the mathematical model are also provided \citep{dowding2008formulation}.  \\
     \hline
     Kadi4Mat & KIT & Materials modeling & P2V, OPT & A web-based application facilitating easy and intuitive access, exchange, and integration of mostly battery materials data. https://demo-kadi4mat.iam-cms.kit.edu/ \\
    \hline
    Bosch Production Line Performance Dataset & Kaggle & Production monitoring for quality control & P2V, V2P, OPT & https://www.kaggle.com/
    competitions/bosch-production-line-performance/data \\
    \hline
    3D Data Hack Dublin (\cite{datahackdublin}) & Dublin City Council etc. & 3D modeling for urban planning, etc. & DS, P3V, V2P & 3D datasets for creating a digital twin of the Docklands area in Dublin, Ireland \cite{white2021digital} \\
     \hline \hline
    \end{tabular}
    \label{tab:dataset}
\end{table*}

\section{Perspectives on UQ and optimization for digital twins}
\label{sec8}

\subsection{Digital twins for structural life cycle management}
The concept of digital twin for structural life cycle management is primarily defined by its usage objectives, and the properties required to perform those objectives. These digital representations or manifestations may be used for several such goals including (but not limited to) visualization/curation, structural state awareness quantification, and prediction of future structural performance (e.g., when critical limit states are expected to occur). Consequently, digital twins take on a variety of forms ranging from point clouds graphical representations to machine learners to physics-based simulators such as finite element models. Taken together, this collection of capabilities empowers the possibility of structural life cycle emulation. Initial designs (drawings, CAD, etc.) can be configured into baseline digital realizations associated with nominal expected performance properties (e.g., material properties, connectivity); these realizations may be updated by surveys (e.g., a UAV multi-mode lidar, thermal, visual data taken in the as-built state), and/or deployed in-situ data streams that come from a variety of IoT sources. Diagnostic analytic capability can be added by a variety of supervised or unsupervised learning techniques and physical knowledge of degradation or system state evolution. Stochastic load/demand and environmental models can be added to evolve the structure through a variety of operational scenarios, with uncertainty quantified by any of a variety of techniques, to predict future performance variables all the way to the limit states, thus completing the life cycle. This “cradle-to-grave” capability can be further integrated with utility/cost models so that the consequences of all future states (or decisions made because of these states) may be inferred, leading to the ability to optimize the structure’s life cycle management, including preventative maintenance, repair, etc. The digital twin would thus become the primary tool of the structural owner to manage the structure, leading to “smart structure”-informed decision-making that could potentially mitigate total structural asset management risk. This vision becomes more and more realizable as large-scale computing, data and predictive analytics, and data management continually empower its execution.

\subsection{Digital twins for sustainability}
As discussed in Sec. \ref{Sec1} of our Part 1 paper \citep{SMO1}, a physical system's life cycle can be divided into four distinct phases: development, manufacturing, service, and disposal. Most of the digital twin papers that we reviewed cover one or multiple of the first three life cycle phases. We did not find many studies that look at the disposal phase, where a physical system or product which has reached its end of life in the first life cycle could be reused, remanufactured, or recycled. Let us now look at two representative examples where digital twins have the potential to make an enormous and long-lasting impact at the disposal phase: (1) \emph{remanufacturing} in the equipment manufacturing sector and (2) \emph{battery repurposing} in the energy sector. 

Remanufacturing is a process where used products/subassemblies are returned to an ``as-new" condition \citep{ijomah2004remanufacturing} or even a ``better-than-new" condition \citep{zhao2010varying} with significantly lower energy and materials consumption than original manufacturing \citep{li2021reliability}. Remanufacturing has been gaining popularity in recent years, as it plays an important role in driving the global transition to a circular economy. Examples of physical systems that can be remanufactured when approaching the end of design life include a piece of industrial equipment on a production floor \citep{zacharaki2021reclaim} and a high-value component of an agricultural machine in the field \citep{li2021reliability}. More research is needed to exploit the digital twin concept to promote wide-scale adoption of remanufacturing in the manufacturing industry. An interesting research topic along this line is the development of ML methods for non-destructive damage assessment and RUL prediction of cores (used products/subassemblies returned for remanufacturing) on a remanufacturing floor. Although fault diagnostics and failure prognostics have found successes in industrial, structural, and energy applications, as discussed in Sec. 4.4 of our Part 1 paper, such success stories are rare in remanufacturing mainly due to difficulties in accessing historical data from returned products/subassemblies and in collecting faulty/run-to-failure data representative of returned products/subassemblies. Physics-informed ML approaches discussed in Sec. 3.4 of Part 1 can certainly be used to tackle these data challenges. A recent attempt was made to quantify and forecast accumulated fatigue damage in recycled materials \citep{hsu2021machine}. This attempt can be classified as the ML-assisted approach (Approach 6) shown in Fig. 10 of Part 1. It is anticipated that remanufacturing will continue to grow in the manufacturing industry. We call for industry-relevant studies that investigate how to harness the power of digital twins and their building blocks, such as ML, DL, and IIoT, to build rapid diagnostic, prognostic, and decision support tools. Making these tools available to remanufacturers will lead to opportunities to reduce the cost of remanufacturing and increase the core reuse rate, eventually making remanufacturing more cost-competitive and sustainable. 

The other example is repurposing used electric vehicle batteries for stationary storage applications \citep{zhu2021end}. To date, battery repurposing has been mostly for small-scale commercial applications, such as portable battery packs for camping and backpacking, waste disposal bins that use batteries to store solar energy, and off-grid lighting, as reported in \cite{zhu2021end}, and small-scale recreational applications including recreational vehicles and golf cars. Building on the digital twin concept, two notable efforts are being pursued to support battery repurposing for larger-scale commercial applications. 

\begin{itemize}
\item \textbf{Rapid diagnostics:} Similar to remanufacturing, battery repurposing typically requires rapid testing/grading methods to estimate the state of health of used electric vehicle batteries and predict their RUL. Estimating the state of health at the end of the first life and predicting the RUL over the second life is essential to determining the economic and technical viability of repurposing any used battery pack/module for a specific second-life application. Now we can look back at the proposed five-dimensional digital twin model in Fig. 3 of the Part 1 paper and draw connections. The sensor data (P2V or physical-to-virtual) would be voltage and current measurements from a used battery pack/module (physical system or PS) during a rapid physical test. The digital state would be the instantaneous state of health or RUL. This digital state can be the output of an ML model (digital system or DS) built on a training dataset (P2V, physical-to-virtual) to estimate either the probabilities of predefined health grades (classification) or the probability distribution of the state of health or RUL (regression). The resulting estimate of the digital state (state of health or RUL) will enable rapid, near real-time assessment of the economic value (e.g., expected revenue from a simulated second-life scenario vs. associated costs) and technical feasibility (e.g., expected remaining lifetime vs. desirable lifetime, and safety risks) of repurposing the battery pack/module. The economic and technical assessment will provide the battery repurposing company with actionable information on optimally selecting (1) used battery modules with similar state of health levels to create a repurposed pack and (2) the second-life application that a repurposed pack has the best fit for (V2P or virtual-to-physical). Rapid diagnostics for battery repurposing is an exciting application of digital twins. It is probably the right time to invest serious and dedicated research efforts in building easy-to-use and accurate tools for rapid battery diagnostics. Such efforts could start with collecting and sharing performance and degradation data from second-life applications, which we had difficulties finding when populating the open-source datasets in Sec. \ref{sec:open_source}. 
\item \textbf{Battery Passport:} An emerging and attractive alternative to rapid physical testing in a battery repurposing plant is the concept of Battery Passport, initiated by a public-private collaboration platform called the Global Battery Alliance \citep{alliance2020global} in November 2020 and discussed extensively in the battery modeling literature \citep{bai2020energy,zhu2021end,ayerbe2021digitalization}. The Battery Passport can be viewed as a digital twin of a physical battery that comprises all relevant information about the battery from cradle (resource extraction and original manufacturing) to gate (repurposing or recycling). It gives each battery a unique identity consisting of information stored in the cloud or on a digital chip, for example, as part of a battery management system that manages the power and health of the battery. An achievable outcome of the Battery Passport is three-fold: (1) collecting state of health estimates and use conditions of a battery over its first life, (2) forecasting the battery's state of health by comparing the collected first-life information with the expected aging patterns for this battery chemistry, and (3) predicting the cycle life over the second life based on the state of health forecasting results. An obvious benefit would be the possibility of eliminating physical testing in a battery repurposing facility, substantially lowering the overall cost of repurposing. We believe that the value of digital twins to battery repurposing is far greater than what has been realized. As the Battery Passport continues to be adopted globally and across the entire battery value chain, we expect to see increasing adoption of digital twins to drive industry-scale battery repurposing that is both profitable and environmentally responsible.
\end{itemize}

\subsection{UQ of digital twins}

As discussed in earlier sections, the methodologies within UQ can be grouped within several broad categories: input uncertainty quantification, model calibration (parameter and discrepancy estimation), model verification (numerical error quantification), model validation, and uncertainty aggregation towards prediction uncertainty quantification. Several methods are available for each category, and an end-to-end UQ approach can be implemented for static models. On the other hand, a digital twin, which is an evolving model being updated when new data becomes available, presents new challenges, especially in model validation and decision making. The model updated at one time instant can only be validated with future data; thus validation is a lagging indicator of digital twin model quality. However, decisions have to be made with the currently updated model, whose validation is pending based on future data. 

Validation of a digital twin is not the same as validation of a general purpose prediction model. In the case of the latter, validation experiments can be performed to test the model’s performance. This is not possible in the case of a digital twin, since the digital twin is tailored to a specific specimen or realization of a physical system, not to a population of specimens or realizations. Therefore validation data also has to be obtained from future use of a particular specimen; a specimen used in laboratory validation experiment is not the particular specimen for which the digital twin has been constructed. This is why validation is a lagging indicator of a digital twin’s quality.

Since online decision-making for a future time instant is based on a current model which can only be validated in the future, there is additional uncertainty due to lack of validation at the current time instant. How to quantify this uncertainty is a new challenge. One option is to quantify the validation performance and the corresponding model uncertainty in previous time instants and extrapolate to future time instants; however, this raises another question regarding how to quantify the confidence in the extrapolation. In general, the uncertainty of the digital twin is expected to decrease with time, but this is only true when the conditions of the system operation remain similar. When aging (degrading) systems, such as aircraft and civil infrastructure, are used beyond their certified design life or operated outside the envelope of design conditions, newer methods are needed to quantify the uncertainty in the system behavior and its prediction by the digital twin. 

\section{Conclusion}
\label{sec9}
In Part 1, we reviewed key digital twin modeling methods which connect the virtual and physical worlds. However, in practice, measurements of the physical world (P2V) and virtual model outputs (V2P) are never certain. In this paper, we reviewed different methods to quantify the uncertainty of machine learning models and the many digital twin modeling methods discussed in Part 1. Additionally, we discussed various optimization techniques used to improve accuracy, reduce uncertainty, and improve the usefulness of a digital twin model.

In addition to a comprehensive review of digital twin technology, we implemented and presented a battery digital twin model used for determining the optimal time to retire a battery cell from its first life application. The proposed battery digital twin serves as an example of how degradation modeling and optimization can be combined to produce a digital twin model which provides practitioners with actionable information regarding a cell's status in its overall life cycle.

The idea of a true digital twin which can accurately mirror all aspects of a physical asset is rapidly becoming a possibility thanks to the many state-of-the-art modeling techniques discussed in this two part review. In the future, digital twin modeling stands to revolutionize the way manufacturers and end users deliver, maintain, and retire high-value assets. Industry 4.0 is upon us, and big data for digital twin modeling is leading the way.

\backmatter

\section*{Acknowledgements}
Adam Thelen and Chao Hu would like to thank the financial support from the U.S. National Science Foundation under Grant No. ECCS-2015710. Xiaoge Zhang is supported by a grant from the Research Committee of The Hong Kong Polytechnic University under project code 1-BE6V. Sankaran Mahadevan acknowledges the support of the National Institute of Science and Technology. Michael D. Todd and Zhen Hu received financial support from the U.S. Army Corps of Engineers through the U.S. Army Engineer Research and Development Center Research Cooperative Agreement W912HZ-17-2-0024.

\section*{Competing Interests}
The authors have no relevant financial or non-financial conflicts of interest to disclose.

\section*{Replication of results}
The Python code and preprocessed dataset used for the battery case study are available for download on \cite{github_code}.

\section*{Authors' contributions}
All authors read and approved the final manuscript. Hu, C. and Hu, Z. devised the original concept of the review paper. Hu, Z, Thelen, A., and Zhang, X. were responsible for the literature review. Thelen, A. was responsible for geometric modeling. Hu, C., Thelen, A., and Hu, Z. were responsible for physics-based modeling. Hu, Z. was responsible for data-driven modeling. Hu, C. and Hu, Z. were responsible for physics-informed ML. Zhang, X. was responsible for system modeling. Hu, C., and Hu, Z., were responsible for probabilistic model updating. Zhang, X. was responsible for ML model updating. Hu, C. and Hu., Z were responsible for fault diagnostics, failure prognostics, and predictive maintenance. Lu, Y. was responsible for MPC. Fink, O. was responsible for federated learning and domain adaptation. Zhang, X., Hu, Z., and Fink, O. were responsible for deep reinforcement learning. Hu, C. was responsible for UQ of ML models. Hu, Z. was responsible for UQ of dynamic system models, optimization for sensor placement, and optimization for physical system modeling. Lu, Y. was responsible for the optimization of additive manufacturing processes. Zhang, X. and Hu, Z. were responsible for real-time mission planning. Thelen, A. and Hu, C. were responsible for the case study and predictive maintenance scheduling. Hu, C. was responsible for open-source software and data. Ghosh, S. was responsible for the industry demonstration. Hu, C., Todd, M., and Mahadevan, S. were responsible for perspectives. All authors participated in manuscript writing, review, editing, and comment.



\bibliography{sn-bibliography}

\end{document}